\documentclass[preprint, 12pt]{elsarticle}
\usepackage{amsmath, booktabs, graphicx, makecell, stmaryrd, hyperref, bm, amsfonts, threeparttable, mathrsfs, subfigure, geometry, siunitx, multirow}

\DeclareMathOperator*{\argmin}{arg min}

\begin{document}

\clearpage
\title{
    Spectral Analysis of Hard-Constraint PINNs: The Spatial Modulation Mechanism of Boundary Functions
}

\author[1]{Yuchen XIE\fnref{equal}}
\ead{xieych28@mail2.sysu.edu.cn}

\author[1]{Honghang CHI\fnref{equal}}
\ead{chihh@mail2.sysu.edu.cn}

\author[1]{Haopeng QUAN}
\ead{quanhp@mail2.sysu.edu.cn}

\author[1,2]{Yahui WANG}
\ead{wangyh296@mail.sysu.edu.cn}

\author[1]{Wei WANG}
\ead{wangw223@mail.sysu.edu.cn}

\author[1,2]{Yu MA\corref{cor1}}
\ead{mayu9@mail.sysu.edu.cn}

\address[1]{Sino--French Institute of Nuclear Engineering and Technology, Sun Yat-sen University, \\Zhuhai, 519082, PR China}
\address[2]{CNPRI-SYSU Joint Research Center of Coolant Chemistry for Nuclear Reactor, \\Zhuhai, 519082, P.R. China}

\cortext[cor1]{Corresponding author}
\fntext[equal]{These authors contributed equally to this work and should be considered co-first authors.}

\begin{abstract}
    Physics-Informed Neural Networks with hard constraints (HC-PINNs) are increasingly favored for their ability to strictly enforce boundary conditions via a trial function ansatz $\tilde{u} = A + B \cdot N$, yet the theoretical mechanisms governing their training dynamics have remained unexplored.
    Unlike soft-constrained formulations where boundary terms act as additive penalties, this work reveals that the boundary function $B$ introduces a multiplicative spatial modulation that fundamentally alters the learning landscape.
    A rigorous Neural Tangent Kernel (NTK) framework for HC-PINNs is established, deriving the explicit kernel composition law.
    This relationship demonstrates that the boundary function $B(\vec{x})$ functions as a spectral filter, reshaping the eigenspectrum of the neural network's native kernel.
    Through spectral analysis, the effective rank of the residual kernel is identified as a deterministic predictor of training convergence, superior to classical condition numbers.
    It is shown that widely used boundary functions can inadvertently induce spectral collapse, leading to optimization stagnation despite exact boundary satisfaction.
    Validated across multi-dimensional benchmarks, this framework transforms the design of boundary functions from a heuristic choice into a principled spectral optimization problem, providing a solid theoretical foundation for geometric hard constraints in scientific machine learning.
\end{abstract}


\newgeometry{top=5cm}

\maketitle

\restoregeometry
\clearpage

\clearpage
\section{Introduction}
\label{sec_intro}

Physics-Informed Neural Networks (PINNs) \cite{Raissi_2019_Physicsinformedneuralnetworksdeeplearningframeworksolvingforwardinverseproblemsinvolvingnonlinearpartialdifferentialequations} have emerged as a powerful approach for solving partial differential equations (PDEs) \cite{Dwivedi_2021_Distributedlearningmachinessolvingforwardinverseproblemspartialdifferentialequations,Dwivedi_2020_Physicsinformedextremelearningmachinepielmarapidmethodnumericalsolutionpartialdifferentialequations,Xiang_2022_SelfadaptivelossbalancedPhysicsinformedneuralnetworks,Ye_2022_Deepneuralnetworkmethodssolvingforwardinverseproblemstimefractionaldiffusionequationsconformablederivative} by incorporating physical laws directly into the training of neural networks. 
By encoding PDE residuals and boundary conditions (BCs) into the loss function, PINNs offer a mesh-free approach that has demonstrated remarkable success across diverse scientific computing domains, including fluid dynamics \cite{Sun_2020_Surrogatemodelingfluidflowsbasedphysicsconstraineddeeplearningsimulationdata,Cai_2024_PhysicsinformedneuralnetworkssolvingincompressibleNavierStokesequationswindengineering,Mao_2020_Physicsinformedneuralnetworkshighspeedflows,Lou_2021_PhysicsinformedneuralnetworkssolvingforwardinverseflowproblemsBoltzmannBGKformulation}, heat transfer \cite{Gao_2021_PhyGeoNetPhysicsinformedgeometryadaptiveconvolutionalneuralnetworkssolvingparameterizedsteadystatePDEsirregulardomain,Mishra_2021_Physicsinformedneuralnetworkssimulatingradiativetransfera,Mostajeran_2022_DeepBHCPDeepneuralnetworkalgorithmsolvingbackwardheatconductionproblems}, neutron transport \cite{Wang_2022_Surrogatemodelingneutrondiffusionproblemsbasedconservativephysicsinformedneuralnetworksboundaryconditionsenforcement,Xie_2021_Neuralnetworkbaseddeeplearningmethodmultidimensionalneutrondiffusionproblemsnoveltreatmentboundary,Elhareef_2023_PhysicsInformedNeuralNetworkMethodApplicationNuclearReactorCalculationsPilotStudy}, and inverse problems \cite{Baldan_2023_Physicsinformedneuralnetworksinverseelectromagneticproblems,Depina_2022_Applicationphysicsinformedneuralnetworksinverseproblemsunsaturatedgroundwaterflow,Lu_2021_PhysicsInformedNeuralNetworksHardConstraintsInverseDesigna,Sahin_2024_Solvingforwardinverseproblemscontactmechanicsusingphysicsinformedneuralnetworks,Yuan_2022_APINNAuxiliaryphysicsinformedneuralnetworksforwardinverseproblemsnonlinearintegrodifferentialequationsa}. 
The flexibility and generality of PINNs have made them an increasingly popular tool for both forward and inverse PDE problems.

Despite these successes, PINNs face significant challenges when applied to complex problems involving high-frequency components, multiple scales, or sharp gradients \cite{Zhu_2019_Physicsconstraineddeeplearninghighdimensionalsurrogatemodelinguncertaintyquantificationlabeleddata,Fuks_2020_Limitationsphysicsinformedmachinelearningnonlineartwophasetransportporousmedia,Raissi_2018_DeephiddenphysicsmodelsDeeplearningnonlinearpartialdifferentialequations,WangWhenWhyPINNs2022}. 
In such scenarios, training often becomes highly unstable, convergence is exceedingly slow, and the achieved accuracy remains unsatisfactory. 
To address these limitations, research efforts have proceeded along two complementary directions. 
On one hand, various physics-specific strategies have been proposed, including adaptive network architectures \cite{Lu_2021_DeepXDEDeepLearningLibrarySolvingDifferentialEquations}, adaptive loss function designs \cite{McClenny_2023_Selfadaptivephysicsinformedneuralnetworksa}, and adaptive sampling strategies \cite{Wu_2023_comprehensivestudynonadaptiveresidualbasedadaptivesamplingphysicsinformedneuralnetworks}. 
These approaches tailor the PINN framework to exploit specific physical characteristics of the target problems. 
On the other hand, recent work has begun to investigate the training dynamics of PINNs from a theoretical perspective, aiming to reveal fundamental issues that underlie the training difficulties and guide more principled improvements \cite{WangWhenWhyPINNs2022,Saadat_2022_NeuraltangentkernelanalysisPINNadvectiondiffusionequation,Tataranni_2024_PhysicsInformedNeuralNetworksNeuralTangentKernelpreliminaryresultsparametricOptimalControlProblems}.

Among theoretical frameworks for understanding neural network training dynamics, the Neural Tangent Kernel (NTK) theory \cite{Jacot_2018_NeuraltangentkernelConvergencegeneralizationneuralnetworksa} has emerged as a particularly powerful tool. 
Originally developed to analyze the training of deep neural networks in the over-parameterized regime, NTK theory characterizes the evolution of network predictions during gradient-based optimization through a kernel matrix. 
This theory has been extended to analyze PINN training dynamics \cite{WangWhenWhyPINNs2022}, providing insights into convergence behavior and the interplay between PDE residuals and BCs. 
However, existing theoretical work has primarily focused on soft-constrained PINNs, where both PDE residuals and BC violations appear in the loss function. 
This soft-constraint formulation leads to training dynamics equations that simultaneously govern the evolution of both PDE and boundary residuals. 
Consequently, the corresponding NTK matrix contains components from PDE residuals, BC residuals, and their mutual coupling terms, rendering theoretical analysis exceptionally complex.

More critically, the inherent characteristics of the soft-constraint formulation impose fundamental limitations on the depth of NTK theory applications. 
The NTK matrix in soft-constrained PINNs necessarily incorporates loss weight factors that balance PDE and boundary terms. 
This creates a dilemma: if weights are fixed to unity, training is often dominated by boundary fitting, yielding poor results; conversely, if weights are dynamically adjusted based on NTK analysis, then the role of NTK theory becomes confined to weight balancing, precluding its use for guiding other crucial aspects of training such as network architecture design, optimizer selection, and learning rate scheduling. 
As a result, existing studies have predominantly utilized NTK merely as a tool for balancing PDE and BC residuals \cite{WangWhenWhyPINNs2022}, failing to fully exploit the potential of NTK theory in comprehensively guiding PINN training.

Hard-Constrained PINNs (HC-PINNs) offer a promising alternative that addresses these limitations. 
By constructing trial functions that automatically satisfy BCs through their functional form \cite{Lagaris_1998_Artificialneuralnetworkssolvingordinarypartialdifferentialequations,Lagaris_2000_Neuralnetworkmethodsboundaryvalueproblemsirregularboundaries,McFall_2009_Artificialneuralnetworkmethodsolutionboundaryvalueproblemsexactsatisfactionarbitraryboundaryconditions,Xie_2024_Physicsspecializedneuralnetworkhardconstraintssolvingmultimaterialdiffusionproblems,Xie_2023_Automaticboundaryfittingframeworkboundarydependentphysicsinformedneuralnetworksolvingpartialdifferentialequationcomplexboundaryconditions}, HC-PINNs eliminate BC violations entirely. 
However, the impact of this hard-constraint design extends far beyond simple error elimination.
It is argued that the boundary function $B(\vec{x})$ acts as a fundamental \textbf{spatial modulator} of the neural tangent kernel, reshaping the optimization landscape itself.
Unlike soft constraints which merely add penalty terms, hard constraints multiplicatively interact with the neural network's gradients, fundamentally altering the spectral properties of the underlying kernel.
This perspective shifts the focus from "how to balance weights" to "how to design the kernel's spectrum," unlocking a new dimension of theoretical guidance for PINN training.

This paper systematically establishes an NTK theoretical framework for HC-PINNs, identifying the spatial modulation mechanism and its spectral consequences.
The main contributions are as follows:
\begin{enumerate}
    \item Discovery of the Spatial Modulation Mechanism: The analytical relationship $\bm{K}_t = \bm{B} \bm{K}_n \bm{B}$ for the trial function NTK is derived, revealing that the boundary function $B$ acts as a multiplicative gate that locally scales the neural network's tangent kernel. This mechanism encodes boundary information directly into the kernel's structure, enforcing constraints while inevitably altering its spectral properties.
    \item Spectral Analysis of Hard Constraints: A spectral analysis framework for the residual NTK $\bm{K}_r$ is established, identifying the effective rank ($r_{\text{eff}}$) as the primary spectral indicator of training difficulty. The analysis shows that while hard constraints eliminate boundary errors, they can induce ill-conditioning if not properly designed. It is demonstrated that $r_{\text{eff}}$ provides a more reliable predictor of convergence than the condition number, bridging the gap between theoretical spectral properties and practical training dynamics.
    \item Theoretical Foundation for Boundary Function Design: The influence of the geometric properties of $B$ (e.g., curvature, dynamic range) on the spectrum of $\bm{K}_r$ is systematically investigated. This establishes a theoretical basis for automated boundary function design, moving beyond heuristic choices to principled selection criteria based on spectral optimization.
    \item Validation across Dimensions: The framework is validated through comprehensive experiments on 1D, 2D, and 3D diffusion problems. The results confirm that the identified spectral properties robustly predict training performance across varying dimensions and boundary function families, even when using modern optimizers (Adam/L-BFGS) that deviate from the strict lazy training regime.
\end{enumerate}

The remainder of this paper is organized as follows. Section~\ref{sec:methodology} presents the theoretical framework, including the HC-PINN formulation, definitions of NTK matrices for neural networks, trial functions, and PDE residuals, along with theoretical analysis of their spectral properties and training dynamics. Section~\ref{sec:numerical_results} provides comprehensive numerical experiments validating the theoretical predictions and analyzing the influence of various factors on NTK properties and training behavior. Section~\ref{sec:conclusion} concludes the paper with a summary of findings and discussion of future directions.

\clearpage

\clearpage
\section{Methodology}
\label{sec:methodology}

\subsection{HC-PINN}
\label{sec:hc_pinn}

Consider the following well-posed PDE with Dirichlet boundary condition:
\begin{equation}
    \label{eq_pde}
    \mathcal{L}[u(\vec r)] = f(\vec r), \quad \vec r \in \Omega,
\end{equation}
\begin{equation}
    \label{eq_bc}
    u(\vec r) = g(\vec r), \quad \vec r \in \partial \Omega,
\end{equation}
where $\mathcal{L}[\cdot]$ is a differential operator, $f(\vec r)$ and $g(\vec r)$ are known functions, $\Omega$ is the computational domain, and $\partial \Omega$ is the boundary of $\Omega$, $u$ is the unknown solution function, $\vec r$ is the coordinate vector.

In this work, the time variable $t$ is treated as an additional coordinate component, so that $\vec r$ includes both spatial and temporal coordinates.

HC-PINN approximates the solution of the PDE by a function $\tilde{u}$ containing a neural network \cite{Liu_2022_unifiedhardconstraintframeworksolvinggeometricallycomplexpdes}, which is called the trial function and is constructed as follows:
\begin{equation}
    \label{eq_trial}
    \tilde{u}(\vec r, \vec \theta) = A(\vec r) + B(\vec r) N(\vec r, \vec \theta),
\end{equation}
where $N(\vec r, \vec \theta)$ is a neural network with tunning parameters $\vec \theta$,
$\vec \theta = [\theta_1, \theta_2, \ldots, \theta_P]$ is composed of all weights and biases of the neural network, and $P$ is the total number of parameters.

The boundary functions $A$ and $B$ are constructed to satisfy the following conditions:
\begin{align}
    \label{eq_a_bc}
    A(\vec r) & = g(\vec r), \quad \vec r \in \partial \Omega,                        \\
    \label{eq_b_bc}
    B(\vec r) & = 0, \quad \vec r \in \partial \Omega,                                \\
    \label{eq_b_interior}
    B(\vec r) & > 0, \quad \vec r \in \Omega_I = \Omega \setminus \partial \Omega.
\end{align}
where $\Omega_I$ is the interior of the domain $\Omega$.

Conditions \eqref{eq_a_bc} and \eqref{eq_b_bc} ensure that the trial function $\tilde{u}$ in the form of \eqref{eq_trial} always satisfies the Dirichlet BC \eqref{eq_bc}.

Condition \eqref{eq_b_interior} ensures that the trial function does not degenerate to $A$ in the interior of the domain, thus retaining the expressiveness of the neural network, enabling the trial function to approximate the unknown solution arbitrarily well.

The loss function of HC-PINN is defined based on the residual of the PDE \eqref{eq_pde} only:
\begin{equation}
    \label{eq_hc_pinn_loss}
        \mathcal{J}(\vec \theta) = \frac{1}{N_r} \sum_{i=1}^{N_r} \left| \mathcal{L}[\tilde{u}(\vec r_i, \vec \theta)] - f(\vec r_i) \right|^2,
    \end{equation}
where $\{\vec r_i\}_{i=1}^{N_r}$ are the collocation points in the domain $\Omega$, and $N_r$ is the number of collocation points.

The collocation points are also referred to as the training set.

The training of HC-PINN involves minimizing the loss function \eqref{eq_hc_pinn_loss} with respect to the neural network parameters $\vec \theta$. In this work, a hybrid optimization strategy combining the Adam optimizer \cite{kingmaAdamMethodStochastic2014} and the L-BFGS algorithm \cite{liuLimitedMemoryBFGS1989} is employed.
The Adam optimizer is first used for a number of epochs to provide a good initial guess, followed by L-BFGS for fine-tuning.

Stochastic gradient descent (SGD) optimizers \cite{amariBackpropagationStochasticGradient1993} commonly used in NTK analysis are not widely employed in this work, as they converge too slowly during HC-PINN training; see Section~\ref{sec_optimizers} for details.

\subsection{Overview of NTK theoretical framework for HC-PINNs}
\label{sec:framework}

\begin{figure}[!htb]
    \centering
    \includegraphics[width=0.95\textwidth]{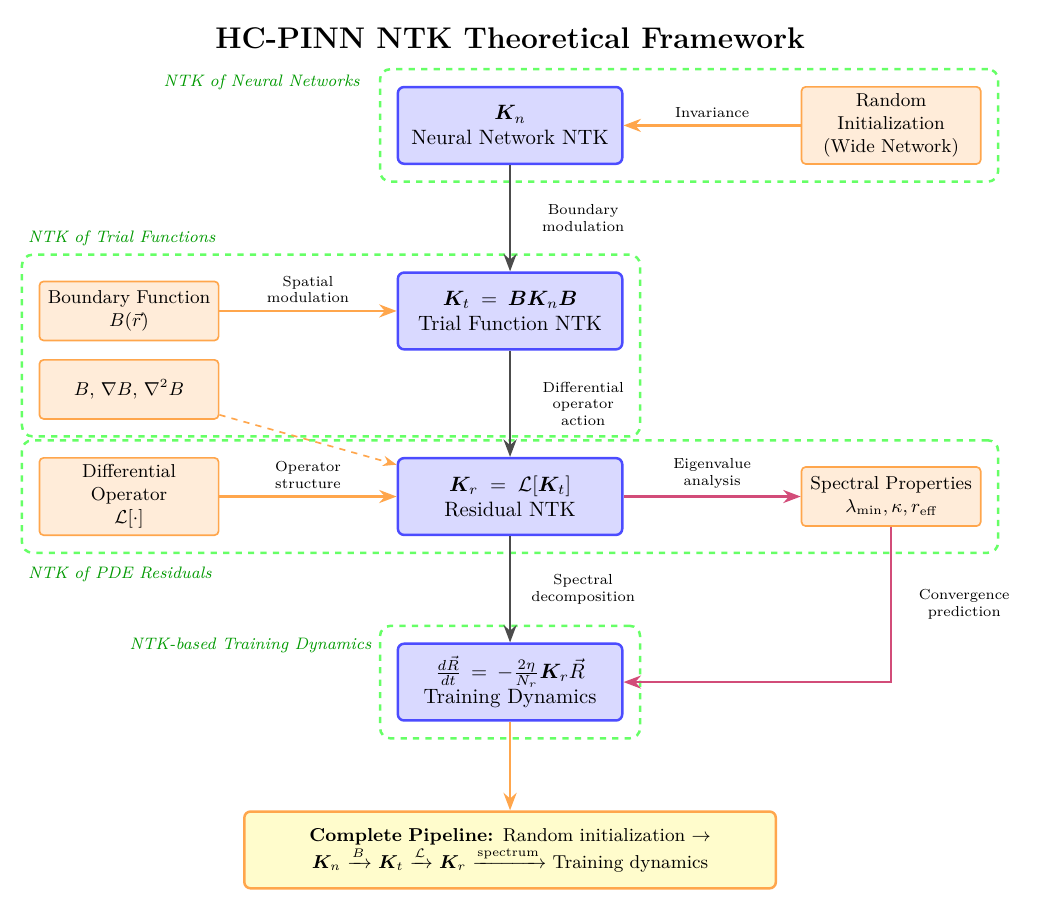}
    \caption{Theoretical framework for HC-PINN NTK analysis.}
    \label{fig:framework}
\end{figure}

This section outlines the theoretical framework for analyzing HC-PINN training dynamics via NTK theory, as illustrated in Fig.~\ref{fig:framework}. The framework establishes a systematic pipeline connecting initialization, trial function construction, residual computation, and convergence analysis through four key components.

First, Section~\ref{sec:ntk_nn} defines the neural network NTK matrix $\bm{K}_n$ and verifies its invariance under random initialization. This property ensures that the analysis is robust and independent of specific random seeds in the wide network regime.

Second, Section~\ref{sec_ntk_trial} derives the trial function NTK $\bm{K}_t$ and identifies the spatial modulation mechanism $\bm{K}_t = \bm{B} \bm{K}_n \bm{B}$. This relationship reveals how the boundary function $B$ in the ansatz $\tilde{u} = A + BN$ shapes the kernel's spectral properties and approximation capacity.

Third, Section~\ref{sec_ntk_residual} formulates the residual NTK $\bm{K}_r$ for general linear differential operators. This matrix governs the optimization landscape, demonstrating how the differential operator interacts with the boundary function and its derivatives to influence training efficiency.

Finally, Section~\ref{sec_ntk_training_dynamics} derives the residual evolution equation $\frac{d\vec{R}}{dt} = -\frac{2\eta}{N_r} \bm{K}_r \vec{R}$. By analyzing the spectral properties of $\bm{K}_r$---particularly the effective rank $r_{\text{eff}}$---convergence behavior can be predicted and theoretical guidance for boundary function selection is provided.

\subsection{NTK of Neural Networks}
\label{sec:ntk_nn}

The NTK of a neural network evaluated on a set of collocation points $\{\vec r_i\}_{i=1}^{N_r}$ is defined as a matrix $\bm{K}_n \in \mathbb{R}^{N_r \times N_r}$, with elements given by \cite{ntk}:
\begin{equation}
    \label{eq_ntk_nn}
    \bm{K_n}(\vec r_i, \vec r_j) = \left\langle \vec J(\vec r_i, \vec \theta), \vec J(\vec r_j, \vec \theta) \right\rangle,
\end{equation}
where $\vec J(\vec r, \vec \theta)$ is the Jacobian vector of the neural network with respect to its parameters $\vec \theta$ and $\langle \cdot, \cdot \rangle$ denotes the inner product operation.

The inner product $\langle \vec{a}, \vec{b} \rangle$ between two vectors $\vec{a} = [a_1, a_2, \ldots, a_n]$ and $\vec{b} = [b_1, b_2, \ldots, b_n]$ is defined as the sum of the products of their corresponding elements, as follows:
\begin{equation}
    \langle \vec{a}, \vec{b} \rangle = \sum_{i} a_i b_i
\end{equation}
where $a_i$ and $b_i$ are the components of vectors $\vec{a}$ and $\vec{b}$, respectively.

The Jacobian vector $\vec J(\vec r, \vec \theta)$ is defined as:
\begin{equation}
    \vec J(\vec r, \vec \theta) = [\frac{\partial N(\vec r, \vec \theta)}{\partial \theta_1}, \frac{\partial N(\vec r, \vec \theta)}{\partial \theta_2}, \ldots, \frac{\partial N(\vec r, \vec \theta)}{\partial \theta_P}]
\end{equation}
where $\theta_1, \theta_2, \ldots, \theta_P$ are the individual parameters of the neural network.

An equivalent form of equation \eqref{eq_ntk_nn} is given by:
\begin{equation}
    \label{eq_ntk_nn_sum}
    \bm{K_n}(\vec r_i, \vec r_j) = \sum_{k=1}^{P} \frac{\partial N(\vec r_i, \vec \theta)}{\partial \theta_k} \frac{\partial N(\vec r_j, \vec \theta)}{\partial \theta_k}
\end{equation}

According to the theory of NTK \cite{ntk}, for sufficiently wide neural networks, the NTK at initialization exhibits remarkable stability across different random initializations.
This means that networks with the same architecture but initialized with different random seeds produce highly similar initial NTK matrices.
Consequently, for a given network architecture, the initial NTK computed from any randomly initialized network can be used as a representative of that architecture's kernel characteristics.
This property is particularly useful for analyzing the training dynamics and generalization behavior of neural networks without repeatedly computing the NTK for different initializations.

A detailed numerical analysis of the invariance of the initial $\bm{K_n}$ under Kaiming uniform initialization method \cite{HeDeepResidualLearning2016} can be found in \ref{sec_invariance_K_n}.

\subsection{NTK of trial functions}
\label{sec_ntk_trial}

This section derives the NTK matrix $\bm{K}_t$ for HC-PINN trial functions and analyzes how the boundary function $B$ modulates the neural network's spectral properties.

\subsubsection{Derivation and Spatial Modulation Mechanism}

The NTK of a trial function $\tilde{u}(\vec r, \vec \theta)$ on collocation points $\{\vec r_i\}_{i=1}^{N_r}$ is defined as:
\begin{equation}
    \label{eq_ntk_trial}
    \bm{K_t}(\vec r_i, \vec r_j) = \left\langle \nabla_{\vec \theta} \tilde{u}(\vec r_i, \vec \theta), \nabla_{\vec \theta} \tilde{u}(\vec r_j, \vec \theta) \right\rangle.
\end{equation}
Substituting the HC-PINN ansatz $\tilde{u}(\vec r) = A(\vec r) + B(\vec r)N(\vec r, \vec \theta)$, and noting that $A$ and $B$ are independent of $\vec \theta$, the gradient is $\nabla_{\vec \theta} \tilde{u} = B(\vec r) \nabla_{\vec \theta} N$. This yields the relationship:
\begin{equation}
    \label{eq_ntk_trial_matrix}
    \bm{K_t} = \bm{B} \bm{K_n} \bm{B},
\end{equation}
where $\bm{B} = \text{diag}(B(\vec r_1), \dots, B(\vec r_{N_r}))$.

Equation \eqref{eq_ntk_trial_matrix} reveals that the boundary function $B$ acts as a spatial modulator on the neural network kernel $\bm{K}_n$. This mechanism has several key implications:
\begin{itemize}
    \item Boundary Vanishing: For points on the boundary $\partial \Omega$, $B(\vec r)=0$, forcing $\bm{K}_t$ entries to zero. This reflects the hard constraint nature: the network's training dynamics are completely suppressed at the boundary, ensuring the boundary condition is strictly maintained by $A(\vec r)$.
    \item Scaling Property: If the boundary function is scaled by a constant $c$ (i.e., $\tilde{B} = cB$), the eigenvalues of $\bm{K}_t$ scale by $c^2$, while the condition number and effective rank remain invariant. This implies that the \textit{shape} of $B$, rather than its magnitude, determines the optimization landscape's complexity.    
    \item Spectral Filtering: The multiplication by $\bm{B}$ can be interpreted as applying a window function in the spatial domain, which corresponds to a filtering operation in the spectral domain. By suppressing the kernel magnitude near the boundaries, $B$ effectively filters out certain eigenmodes of $\bm{K}_n$, reshaping the eigenspectrum. This explains why different choices of $B$ (e.g., with different decay rates or curvatures) lead to drastically different spectral properties (effective rank, condition number) of the resulting kernel, as will be shown in Section \ref{sec_ntk_trial_spectrum}.    
    \item Initialization Invariance: Since $\bm{B}$ is deterministic, the initialization invariance of $\bm{K}_n$ (discussed in Section \ref{sec:ntk_nn}) directly extends to $\bm{K}_t$.
\end{itemize}

\subsubsection{Spectral Characterization Metrics}
\label{sec:K_t_spectral}

To quantitatively analyze the spectral properties of $\bm{K}_t$ (and later $\bm{K}_r$), three key metrics are introduced. These indicators will be used in subsequent sections to predict training convergence and stability.

\begin{enumerate}
    \item Condition Number ($\kappa$): Defined as $\kappa = \lambda_{\max} / \lambda_{\min}$. A lower $\kappa$ indicates a more uniform loss landscape and better numerical stability.
    \item Effective Rank ($r_{\text{eff}}$): A measure of the effective dimensionality of the feature space, defined as:
          \begin{equation}
              r_{\text{eff}} = \frac{(\sum_{i} \lambda_i)^2}{\sum_{i} \lambda_i^2} = \frac{\text{tr}(\bm{K})^2}{\|\bm{K}\|_F^2}.
          \end{equation}
          A higher $r_{\text{eff}}$ implies a more uniform distribution of eigenvalues, suggesting that the network can learn features across a broader frequency spectrum efficiently.
    \item Trace ($\text{tr}(\bm{K})$): The sum of eigenvalues, representing the total "energy" or magnitude of the kernel.
\end{enumerate}

\subsection{NTK of PDE residuals}
\label{sec_ntk_residual}

In this section, the definition of the NTK matrix $\bm{K}_r$ for the PDE residuals in HC-PINN is presented, and its expanded expression is derived.

Similar to the NTK of trial functions presented in Section \ref{sec_ntk_trial}, the NTK of PDE residuals evaluated on a set of collocation points $\{\vec r_i\}_{i=1}^{N_r}$ is defined as a matrix $\bm{K}_r \in \mathbb{R}^{N_r \times N_r}$, with elements given by:
\begin{equation}
    \label{eq_ntk_residual}
    \bm{K}_r(\vec r_i, \vec r_j) = \left\langle \vec J_r(\vec r_i, \vec \theta), \vec J_r(\vec r_j, \vec \theta) \right\rangle,
\end{equation}
where $\vec J_r(\vec r, \vec \theta)$ is the Jacobian vector of the PDE residual with respect to its parameters $\vec \theta$.
The Jacobian vector $\vec J_r(\vec r, \vec \theta)$ is defined as:
\begin{equation}
    \label{eq_jacobian_residual}
    \vec J_r(\vec r, \vec \theta) = \left[\frac{\partial \mathcal{R}(\vec r, \vec \theta)}{\partial \theta_1}, \frac{\partial \mathcal{R}(\vec r, \vec \theta)}{\partial \theta_2}, \ldots, \frac{\partial \mathcal{R}(\vec r, \vec \theta)}{\partial \theta_P}\right]
\end{equation}
where $\mathcal{R}(\vec r, \vec \theta) = \mathcal{L}[\tilde{u}(\vec r, \vec \theta)] - f(\vec r)$ denotes the PDE residual.
An equivalent form of Eq.~\eqref{eq_ntk_residual} is given by:
\begin{equation}
    \label{eq_ntk_residual_sum}
    \bm{K}_r(\vec r_i, \vec r_j) = \sum_{k=1}^{P} \frac{\partial \mathcal{R}(\vec r_i, \vec \theta)}{\partial \theta_k} \frac{\partial \mathcal{R}(\vec r_j, \vec \theta)}{\partial \theta_k}.
\end{equation}

To systematically analyze the residual NTK matrix, a general linear differential operator is first considered that can be expressed as a linear combination of derivatives of different orders:
\begin{equation}
    \label{eq_general_operator}
    \mathcal{L}u = c_0(\vec r) u + \vec c_1(\vec r) \cdot \nabla u + c_2(\vec r) \nabla^2 u,
\end{equation}
where $c_0(\vec r)$ is a scalar coefficient for the zeroth-order term, $\vec c_1(\vec r)$ is a vector coefficient for the first-order term (gradient), and $c_2(\vec r)$ is a scalar coefficient for the second-order term (Laplacian).
This general form encompasses a wide range of PDEs, including Poisson equations, convection-diffusion equations, and reaction-diffusion equations.

Substituting the HC-PINN trial function \eqref{eq_trial} into the linear differential operator \eqref{eq_general_operator}:
\begin{equation}
    \begin{aligned}
        \mathcal{L}\tilde{u} =
         & c_0(A + BN)                                                               \\
         & + \vec c_1 \cdot (\nabla A + N\nabla B + B\nabla N)                       \\
         & + c_2(\nabla^2 A + N\nabla^2 B + 2\nabla B \cdot \nabla N + B\nabla^2 N).
    \end{aligned}
\end{equation}
Collecting terms involving the neural network $N$ and its derivatives yields:
\begin{equation}
    \begin{aligned}
        \mathcal{L}\tilde{u} = & \; \underbrace{c_0 A + \vec c_1 \cdot \nabla A + c_2 \nabla^2 A}_{\text{non-trainable terms}}                                     \\
                               & + \underbrace{(c_0 B + \vec c_1 \cdot \nabla B + c_2 \nabla^2 B)}_{\alpha(\vec r)} N                                              \\
                               & + \underbrace{(\vec c_1 B + 2c_2 \nabla B)}_{\vec \beta(\vec r)} \cdot \nabla N + \underbrace{c_2 B}_{\gamma(\vec r)} \nabla^2 N.
    \end{aligned}
\end{equation}
Since the boundary functions $A$ and $B$ do not contain trainable parameters, the partial derivative of the residual $\mathcal{R} = \mathcal{L}\tilde{u} - f$ with respect to parameter $\theta_k$ is:
\begin{equation}
    \label{eq_residual_derivative_general}
    \frac{\partial \mathcal{R}(\vec r, \vec \theta)}{\partial \theta_k} = \alpha(\vec r) \frac{\partial N(\vec r, \vec \theta)}{\partial \theta_k} + \vec \beta(\vec r) \cdot \frac{\partial [\nabla N(\vec r, \vec \theta)]}{\partial \theta_k} + \gamma(\vec r) \frac{\partial [\nabla^2 N(\vec r, \vec \theta)]}{\partial \theta_k},
\end{equation}
where the coefficient functions are:
\begin{equation}
    \label{eq_res_coefficients_general}
    \begin{cases}
        \alpha(\vec r) & = c_0 B + \vec c_1 \cdot \nabla B + c_2 \nabla^2 B,  \\
    \vec \beta(\vec r) & = \vec c_1 B + 2c_2 \nabla B,        \\
    \gamma(\vec r)     & = c_2 B. 
\end{cases}
\end{equation}
Substituting Eq.~\eqref{eq_residual_derivative_general} into Eq.~\eqref{eq_ntk_residual_sum}, the general form of the residual NTK matrix is:
\begin{equation}
    \label{eq_Kr_general_form}
    \begin{aligned}
        \bm{K}_r(\vec r_i, \vec r_j) = & \; \alpha_i \alpha_j \bm{K}_n(\vec r_i, \vec r_j) + \alpha_i (\vec \beta_j \cdot \bm{K}_{N,\nabla N}(\vec r_i, \vec r_j)) + (\vec \beta_i \cdot \bm{K}_{\nabla N, N}(\vec r_i, \vec r_j)) \alpha_j                     \\
                                       & + (\vec \beta_i)^T \bm{K}_{\nabla N}(\vec r_i, \vec r_j) \vec \beta_j + \alpha_i \gamma_j \bm{K}_{N,\nabla^2 N}(\vec r_i, \vec r_j) + \gamma_i \alpha_j \bm{K}_{\nabla^2 N, N}(\vec r_i, \vec r_j)                     \\
                                       & + (\vec \beta_i)^T \bm{K}_{\nabla N,\nabla^2 N}(\vec r_i, \vec r_j) \gamma_j + \gamma_i (\vec \beta_j)^T \bm{K}_{\nabla^2 N,\nabla N}(\vec r_i, \vec r_j) + \gamma_i \gamma_j \bm{K}_{\nabla^2 N}(\vec r_i, \vec r_j),
    \end{aligned}
\end{equation}
where the shorthand notation is used: $\alpha_i = \alpha(\vec r_i)$, $\vec \beta_i = \vec \beta(\vec r_i)$, $\gamma_i = \gamma(\vec r_i)$.
The NTK components are defined as follows:
\begin{equation}
    \begin{cases}
        \bm{K}_N(i,j)                      & = \sum_k \dfrac{\partial N_i}{\partial \theta_k} \dfrac{\partial N_j}{\partial \theta_k},                                   \\
        \bm{K}_{\nabla N}(i,j)             & = \sum_k \dfrac{\partial [\nabla N_i]}{\partial \theta_k} \otimes \dfrac{\partial [\nabla N_j]}{\partial \theta_k},         \\
        \bm{K}_{N,\nabla N}(i,j)           & = \sum_k \dfrac{\partial N_i}{\partial \theta_k} \dfrac{\partial [\nabla N_j]}{\partial \theta_k},                          \\
        \bm{K}_{N,\nabla^2 N}(i,j)         & = \sum_k \dfrac{\partial N_i}{\partial \theta_k} \dfrac{\partial [\nabla^2 N_j]}{\partial \theta_k},                        \\
        \bm{K}_{\nabla N,\nabla^2 N}(i,j)  & = \sum_k \dfrac{\partial [\nabla N_i]}{\partial \theta_k} \cdot \dfrac{\partial [\nabla^2 N_j]}{\partial \theta_k},         \\
        \bm{K}_{\nabla^2 N}(i,j)           & = \sum_k \dfrac{\partial [\nabla^2 N_i]}{\partial \theta_k} \dfrac{\partial [\nabla^2 N_j]}{\partial \theta_k}.
\end{cases}
\end{equation}
where $\otimes$ denotes the outer product of two vectors, and $N_i = N(\vec r_i, \vec \theta)$ denotes the neural network evaluated at $\vec r_i$
Due to the symmetry of the inner product, the following transpose relations hold:
\begin{equation}
    \bm{K}_{\nabla N, N} = \bm{K}_{N,\nabla N}^T, \quad \bm{K}_{\nabla^2 N, N} = \bm{K}_{N,\nabla^2 N}^T, \quad \bm{K}_{\nabla^2 N,\nabla N} = \bm{K}_{\nabla N,\nabla^2 N}^T.
\end{equation}

The general expression \eqref{eq_Kr_general_form} reveals several fundamental characteristics of $\bm{K}_r$:
\begin{enumerate}
    \item Multi-component structure: $\bm{K}_r$ is a weighted combination of multiple NTK components corresponding to different derivative orders. Unlike $\bm{K}_t = \bm{B} \bm{K}_n \bm{B}$ in Eq.~\eqref{eq_ntk_trial_matrix}, there is no simple algebraic relationship between $\bm{K}_r$ and $\bm{K}_t$ or $\bm{K}_n$.

    \item Spatially-varying coefficients: The weights $\alpha(\vec r)$, $\vec \beta(\vec r)$, and $\gamma(\vec r)$ depend explicitly on the boundary function $B$ and its spatial derivatives $\nabla B$ and $\nabla^2 B$, causing spatial modulation of the NTK components.

    \item Operator-dependent structure: The specific form of $\bm{K}_r$ varies significantly with the differential operator through the coefficients $c_0$, $\vec c_1$, and $c_2$. Different types of PDEs lead to different relative contributions of the zeroth, first, and second-order NTK components.

    \item Coupling between derivative orders: The cross terms such as $\bm{K}_{N,\nabla N}$ and $\bm{K}_{\nabla N,\nabla^2 N}$ indicate coupling between different derivative orders of the neural network, which does not appear in $\bm{K}_t$.
\end{enumerate}

Due to the high complexity of the definition of $\bm{K}_r$ in Eq.~\eqref{eq_Kr_general_form} and the diversity of differential operators, it is generally not feasible to mathematically analyze the influence of the boundary function $B$ on the properties of $\bm{K}_r$. Therefore, this work focuses on numerical analysis; comprehensive numerical experiments are presented in Section~\ref{sec_ntk_res_spectrum}.

\subsection{NTK-based Training Dynamics Analysis}
\label{sec_ntk_training_dynamics}

In this section, the training dynamics of HC-PINN are analyzed through the lens of NTK theory. By establishing the relationship between the residual NTK matrix $\bm{K}_r$ and the parameter $\vec \theta$ evolution during training, some analytical predictions for convergence behavior are derived and the influence of boundary function $B$ on training efficiency is investigated.

Unless otherwise specified, the NTK matrix mentioned in this section refers to the NTK matrix at the time before training, i.e. $\bm{K}^{(0)}$.

\subsubsection{Residual evolution equation}
\label{subsec_residual_evolution}

For HC-PINN, the training loss function is defined as the mean squared residual over a set of collocation points $\{\vec r_i\}_{i=1}^{N_r}$, as shown in Eq.~\eqref{eq_hc_pinn_loss}, and can be expressed by PDE residual $\mathcal{R}$:
\begin{equation}
    \label{eq_loss_function}
    \mathcal{J}(\vec \theta) = \frac{1}{N_r} \sum_{i=1}^{N_r} \mathcal{R}^2(\vec r_i, \vec \theta).
\end{equation}
The training objective is to find optimal parameters $\vec \theta^*$ that minimize the loss:
\begin{equation}
    \label{eq_training_objective}
    \vec \theta^* = \argmin_{\vec \theta} \mathcal{J}(\vec \theta).
\end{equation}
To solve this optimization problem, the gradient descent method is employed, which updates the parameters in the direction opposite to the gradient of the loss function. 

In practice, gradient descent corresponds to using a stochastic gradient descent optimizer without momentum to train the neural network.
In continuous time, this corresponds to the following ordinary differential equation (ODE):
\begin{equation}
    \label{eq_gradient_descent_basic}
    \frac{\text{d} \vec \theta}{\text{d} t} = -\eta \nabla_{\vec \theta} \mathcal{J}(\vec \theta),
\end{equation}
where $\eta > 0$ is the learning rate controlling the step size of parameter updates.
To derive the explicit form of the evolution equation \eqref{eq_gradient_descent_basic}, the gradient of the loss function is expanded by applying the chain rule:
\begin{equation}
    \label{eq_gradient_loss}
    \nabla_{\vec \theta} \mathcal{J}(\vec \theta) = \frac{1}{N_r} \sum_{i=1}^{N_r} \nabla_{\vec \theta} \left[\mathcal{R}^2(\vec r_i, \vec \theta)\right] = \frac{2}{N_r} \sum_{i=1}^{N_r} \mathcal{R}(\vec r_i, \vec \theta) \nabla_{\vec \theta} \mathcal{R}(\vec r_i, \vec \theta).
\end{equation}
Substituting the gradient expression \eqref{eq_gradient_loss} into Eq.~\eqref{eq_gradient_descent_basic}, the parameter evolution equation is obtained as:
\begin{equation}
    \label{eq_param_evolution}
    \frac {\text{d} \vec \theta}{\text{d} t} = -\frac{2\eta}{N_r} \sum_{i=1}^{N_r} \mathcal{R}(\vec r_i, \vec \theta) \nabla_{\vec \theta} \mathcal{R}(\vec r_i, \vec \theta).
\end{equation}

To understand how the training progresses, the evolution equation for the residual is derived. 
Taking the time derivative of the residual:
\begin{equation}
    \frac{\text{d} \mathcal{R}(\vec r_i, \vec \theta)}{\text{d} t} = \sum_{k=1}^{P} \frac{\partial \mathcal{R}(\vec r_i, \vec \theta)}{\partial \theta_k} \frac{\text{d} \theta_k}{\text{d} t} = \vec J_r(\vec r_i, \vec \theta)^T \frac{\text{d} \vec \theta}{\text{d} t},
\end{equation}
where $\vec J_r(\vec r_i, \vec \theta)$ is the Jacobian vector defined in Eq.~\eqref{eq_jacobian_residual}.
Substituting Eq.~\eqref{eq_param_evolution} into the above equation:
\begin{equation}
    \frac{\text{d} \mathcal{R}(\vec r_i, \vec \theta)}{\text{d} t} = -\frac{2\eta}{N_r} \vec J_r(\vec r_i, \vec \theta)^T \sum_{j=1}^{N_r} \mathcal{R}(\vec r_j, \vec \theta) \vec J_r(\vec r_j, \vec \theta).
\end{equation}
In matrix form, the residual evolution can be written as:
\begin{equation}
    \label{eq_residual_evolution}
        \frac{\text{d} \vec R}{\text{d} t} = -\frac{2\eta}{N_r} \bm{K}_r \vec R
\end{equation}
where $\bm{K}_r$ is the residual NTK matrix defined in Eq.~\eqref{eq_ntk_residual},
$\vec R$ is the residual vector at all collocation points:
\begin{equation}
    \vec R = \left[\mathcal{R}(\vec r_1, \vec \theta), \mathcal{R}(\vec r_2, \vec \theta), \ldots, \mathcal{R}(\vec r_{N_r}, \vec \theta)\right]^T.
\end{equation}

This fundamental equation reveals that in the NTK regime (gradient-based optimization with infinitesimal learning rate), the residual evolution \eqref{eq_residual_evolution} is governed by a linear differential equation with the NTK matrix $\bm{K}_r$ acting as the evolution operator.

\subsubsection{Spectral decomposition and convergence analysis}
\label{subsec_spectral_analysis}

To analyze the convergence behavior, the spectral decomposition of the residual NTK matrix is performed. 
Since $\bm{K}_r$ is real symmetric and positive semi-definite, it admits an eigenvalue decomposition:
\begin{equation}
    \label{eq_Kr_spectral}
    \bm{K}_r = \sum_{k=1}^{p} \lambda_k \vec v_k \vec v_k^T,
\end{equation}
where $\lambda_1 \geq \lambda_2 \geq \cdots \geq \lambda_p \geq 0$ are the eigenvalues in descending order, $\vec v_k$ are the corresponding orthonormal eigenvectors, and $p \leq N_r$ is the rank of $\bm{K}_r$.
Expanding the initial residual $\vec R(0)$ in the eigenbasis:
\begin{equation}
    \vec R(0) = \sum_{k=1}^{p} c_k \vec v_k,
\end{equation}
where $c_k = \vec v_k^T \vec R(0)$ are the expansion coefficients.
The analytical solution of Eq.~\eqref{eq_residual_evolution} is given by:
\begin{equation}
    \label{eq_residual_solution}
        \vec R(t) = \sum_{k=1}^{p} c_k \exp\left(-\dfrac{2\eta}{N_r} \lambda_k t\right) \vec v_k
\end{equation}
Decomposition \eqref{eq_residual_solution} reveals several key theoretical insights about the convergence:
\begin{enumerate}
    \item Mode-wise exponential decay: Each eigenmode decays exponentially at a rate proportional to its eigenvalue $\lambda_k$. Modes with larger eigenvalues converge faster.
    \item Convergence time scale: The overall convergence time $t_{\text{conv}}$ is dominated by the slowest mode, which corresponds to the smallest non-zero eigenvalue $\lambda_{\min}$:
    \begin{equation}
        \label{eq_convergence_time}
        t_{\text{conv}} \sim \frac{N_r}{2\eta \lambda_{\min}},
    \end{equation}
    \item Loss decay: The training loss evolves as:
    \begin{equation}
        \mathcal{J}(t) = \frac{1}{N_r} \|\vec R(t)\|^2 = \frac{1}{N_r} \sum_{k=1}^{p} c_k^2 \exp\left(-\frac{4\eta}{N_r} \lambda_k t\right).
    \end{equation}
    For sufficiently large $t$, the loss is dominated by the slowest mode:
    \begin{equation}
        \mathcal{J}(t) \approx \frac{c_{\min}^2}{N_r} \exp\left(-\frac{4\eta}{N_r} \lambda_{\min} t\right), \quad t \gg t_{\text{conv}},
    \end{equation}
    where $c_{\min}$ denotes the expansion coefficient corresponding to the minimum eigenvalue $\lambda_{\min}$,
    exhibiting exponential decay with rate $4\eta \lambda_{\min} / N_r$.
\end{enumerate}

\subsubsection{Impact of spectral properties on training}
\label{subsec_spectral_impact}

The spectral properties of $\bm{K}_r$ directly determine the training efficiency. Three key spectral characteristics are identified:
\begin{enumerate}
    \item Minimum eigenvalue $\lambda_{\min}$.
    The minimum non-zero eigenvalue controls the convergence rate of the slowest mode. According to Eq.~\eqref{eq_convergence_time}, a larger $\lambda_{\min}$ leads to faster convergence, as the convergence time scale is inversely proportional to this eigenvalue. When $\lambda_{\min} \to 0$, the problem becomes ill-conditioned, resulting in extremely slow convergence that can make training impractical.
    \item Condition number $\kappa$.
    The condition number measures the uniformity of convergence across different modes.
    A small condition number indicates that all modes converge at similar rates, leading to stable and efficient training. Conversely, a large condition number means that fast modes converge quickly while slow modes lag behind, resulting in unbalanced training dynamics that potentially require more iterations to achieve the desired accuracy. 
    
    \item Effective rank $r_{\text{eff}}$.
    The effective rank quantifies the number of significant modes participating in training. 
    A large effective rank indicates that many modes contribute significantly, suggesting rich expressive capacity of the neural network for representing the solution. On the other hand, a small effective rank suggests that only a few dominant modes are active, indicating limited degrees of freedom in the solution space.
\end{enumerate}

\subsubsection{Influence of boundary function on training dynamics}
\label{subsec_boundary_influence}

As analyzed in Section~\ref{sec_ntk_residual}, the boundary function $B$ profoundly influences the residual NTK matrix $\bm{K}_r$ through its derivatives $\nabla B$ and $\nabla^2 B$, which appear in the coefficient functions $\alpha(\vec r)$, $\vec \beta(\vec r)$, and $\gamma(\vec r)$ (see Eq.~\eqref{eq_Kr_general_form} and Eq.~\eqref{eq_res_coefficients_general}). 
These coefficients create spatially-varying weights that modulate different NTK components, thereby affecting the eigenvalue spectrum of $\bm{K}_r$.

From the training dynamics perspective established in this section, the choice of boundary function $B$ directly impacts the spectral properties of $\bm{K}_r$, including the minimum eigenvalue $\lambda_{\min}$, condition number $\kappa$, and effective rank $r_{\text{eff}}$, which in turn determine the convergence rate, training stability, and expressive capacity according to the analysis in Section~\ref{subsec_spectral_analysis}.

However, as discussed in Section~\ref{sec_ntk_residual}, the relationship between $B$ and the spectral properties of $\bm{K}_r$ is highly complex and PDE-dependent. Unlike the trial function NTK $\bm{K}_t$ which has the simple form $\bm{K}_t = \bm{B} \bm{K}_n \bm{B}$, there is no simple algebraic transformation relating $\bm{K}_r$ to $\bm{K}_t$ or $\bm{K}_n$ due to the coupling of multiple derivative orders through the differential operator $\mathcal{L}$. Consequently, the optimal choice of boundary function $B$ cannot be determined purely from theoretical analysis and must be investigated numerically for specific applications.

A comprehensive numerical study of how different boundary function parametrizations affect the training dynamics and convergence behavior is presented in Section~\ref{sec_simple_train}, where the theoretical predictions from this section are partially validated and the practical impact of $B$ on HC-PINN performance is quantified.

\subsubsection{Applicability and limitations of the NTK-based analysis}
\label{subsec_ntk_limitations}

It is important to emphasize that the training dynamics analysis presented in this section is rigorously valid under specific theoretical assumptions that are typically associated with the NTK regime:
\begin{enumerate}
    \item Infinite width limit: The NTK theory assumes neural networks in the infinite width limit where the network parameters remain close to their initialization throughout training.
    \item Gradient descent with small learning rate: The continuous-time evolution Eqs.~\eqref{eq_param_evolution} and \eqref{eq_residual_evolution} are derived under the assumption of gradient descent (i.e. stochastic gradient descent without momentum) with infinitesimally small learning rate $\eta \to 0$.
    \item No momentum or adaptive learning rates: The linear dynamics in Eq.~\eqref{eq_residual_evolution} does not account for momentum terms or adaptive learning rate adjustments present in modern optimizers, such as Adam or L-BFGS.
\end{enumerate}

As demonstrated in the numerical experiments (Section~\ref{sec_simple_train}), vanilla gradient descent with small learning rates is often impractical due to slow convergence, making Adam and L-BFGS the preferred choices in practice:
\begin{itemize}
    \item Adam optimizer: Uses adaptive learning rates for each parameter and incorporates momentum through exponentially weighted moving averages of gradients and squared gradients.
    \item L-BFGS optimizer: A quasi-Newton method that approximates the inverse Hessian matrix using limited memory, enabling second-order optimization with superlinear convergence rates.
\end{itemize}

These advanced optimizers deviate from the idealized gradient descent setting in several ways:
\begin{enumerate}
    \item The learning rate varies across parameters and iterations
    \item Momentum terms introduce history-dependent dynamics that are not captured by the instantaneous gradient
    \item Second-order information (in L-BFGS) can accelerate convergence beyond the predictions of first-order NTK analysis
\end{enumerate}

Consequently, the quantitative predictions from the NTK-based analysis (such as exact convergence times or eigenvalue-based rates) do not hold exactly for Adam or L-BFGS training. 

Nevertheless, the NTK framework still provides valuable qualitative insights even when the strict assumptions are violated:
\begin{itemize}
    \item The spectral properties of $\bm{K}_r$ (eigenvalue distribution, condition number) remain indicative of problem difficulty and potential convergence issues
    \item The identification of slow modes (small eigenvalues) helps diagnose problematic regions or ill-conditioned aspects of the problem
    \item The analysis of how $B$ influences $\bm{K}_r$ guides the design of boundary functions
    \item Comparative analysis of $\bm{K}_r$ for different choices of boundary function $B$ can inform hyperparameter selection
\end{itemize}
While the NTK-based training dynamics analysis should be viewed as an approximate theoretical guide rather than an exact predictive tool for practical optimization algorithms, it remains a useful framework for understanding and improving HC-PINN training. The extent to which these theoretical insights translate to practical performance with Adam and L-BFGS optimizers is investigated through comprehensive numerical experiments in Section~\ref{sec_simple_train}.

\subsubsection{Summary of NTK-based training dynamics}

In this section, the NTK-based training dynamics framework for HC-PINN was established by deriving the residual evolution equation from first principles. Starting from the optimization objective of minimizing the loss function, gradient descent is applied to obtain the parameter evolution equation, and then the fundamental residual evolution equation (Eq.~\eqref{eq_residual_evolution}) is derived, which reveals that in the NTK regime, the training dynamics is governed by a linear ODE with $\bm{K}_r$ as the evolution operator. Through eigenvalue decomposition of $\bm{K}_r$, analytical solutions showing mode-wise exponential decay are obtained and three key spectral characteristics that directly determine training efficiency are identified: the minimum eigenvalue $\lambda_{\min}$ controls the convergence time scale, the condition number $\kappa$ measures the uniformity of convergence across modes, and the effective rank $r_{\text{eff}}$ quantifies the expressive capacity participating in training. It is also clarified how the boundary function $B$ affects training dynamics by modulating these spectral properties of $\bm{K}_r$ through its derivatives, though the complex, PDE-dependent relationship between $B$ and $\bm{K}_r$ prevents analytical determination of optimal choices. Finally, it is emphasized that the NTK-based analysis is rigorously valid under idealized assumptions (infinite width limit, small learning rate gradient descent without momentum), which are typically violated by practical optimizers like Adam and L-BFGS. Therefore, while the framework provides valuable qualitative guidance for understanding convergence behavior and designing boundary functions, numerical experiments remain essential for validation and practical optimization, as detailed in Section~\ref{sec_simple_train}.

\clearpage

\clearpage
\section{Numerical results and analysis}
\label{sec:numerical_results}

\subsection{Experimental Setup}
\label{sec:experimental_setup}

To systematically investigate the influence of boundary function $B$ formulation on the spectral properties of NTK and training dynamics, a standardized experimental framework is established.
Unless otherwise specified in subsequent sections, the following default settings are adopted for all numerical experiments.

A fully connected neural network (MLP) is employed as the backbone for all experiments, including the analysis of trial NTK $\bm{K}_t$, residual NTK $\bm{K}_r$, and training dynamics:
\begin{itemize}
    \item Architecture 2 hidden layers with 500 neurons per layer.
    \item Activation Function Hyperbolic tangent ($\tanh$).
    \item Initialization Kaiming uniform initialization.
\end{itemize}

For the training dynamics experiments (Section \ref{sec_simple_train}), a hybrid optimization strategy is utilized to ensure both robust convergence and high precision:
\begin{enumerate}
    \item Phase 1 (Adam) The Adam optimizer is used for the initial \num{10000} iterations with a learning rate of $10^{-3}$ to rapidly navigate the loss landscape.
    \item Phase 2 (L-BFGS) The L-BFGS optimizer is employed for fine-tuning until convergence or a maximum of \num{500} iterations, with a tolerance of machine epsilon.
\end{enumerate}
This hybrid approach combines the robustness of Adam in the initial phase with the high precision of L-BFGS near the minimum.
Final performance is evaluated using the relative $L^2$ error on a uniform grid of test points ($1000$ for 1D/2D, $125,000$ for 3D).

\subsection{Analysis of the spectral properties of \texorpdfstring{$\bm{K_t}$}{Kt}}
\label{sec_ntk_trial_spectrum}

\begin{figure}[!htb]
    \centering
    \includegraphics[width=0.95\textwidth]{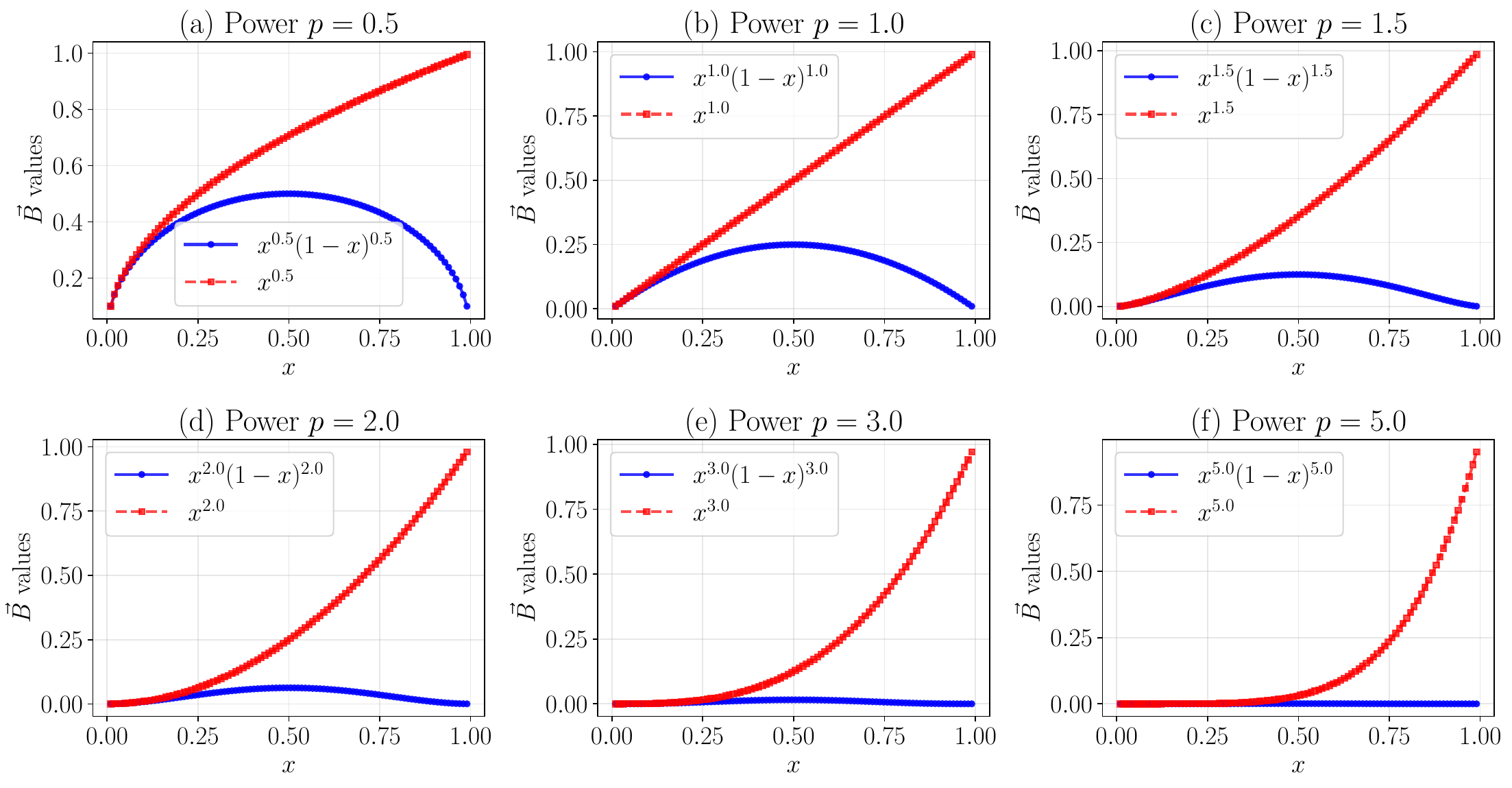}
    \caption{Distributions of power function family boundary functions for varying exponent $p$.}
    \label{fig:kt_power_distributions}
\end{figure}

The influence of the boundary function $B$ on the spectral properties of the trial function NTK matrix $\bm{K_t}$ is empirically investigated in this section.
To ensure the generality of the analysis, a standard fully connected neural network architecture with 2 hidden layers of 500 neurons and $\tanh$ activation, initialized using Kaiming uniform initialization, is employed.
The analysis is focused on the Power function family $B(x) = x^p(1-x)^p$ (symmetric) and $B(x) = x^p$ (asymmetric) with $p \in \{0.5, 1.0, 1.5, 2.0, 2.5, 3.0, 4.0, 5.0\}$, as shown in Fig.~\ref{fig:kt_power_distributions}.

\begin{figure}[!htb]
    \centering
    \includegraphics[width=0.95\textwidth]{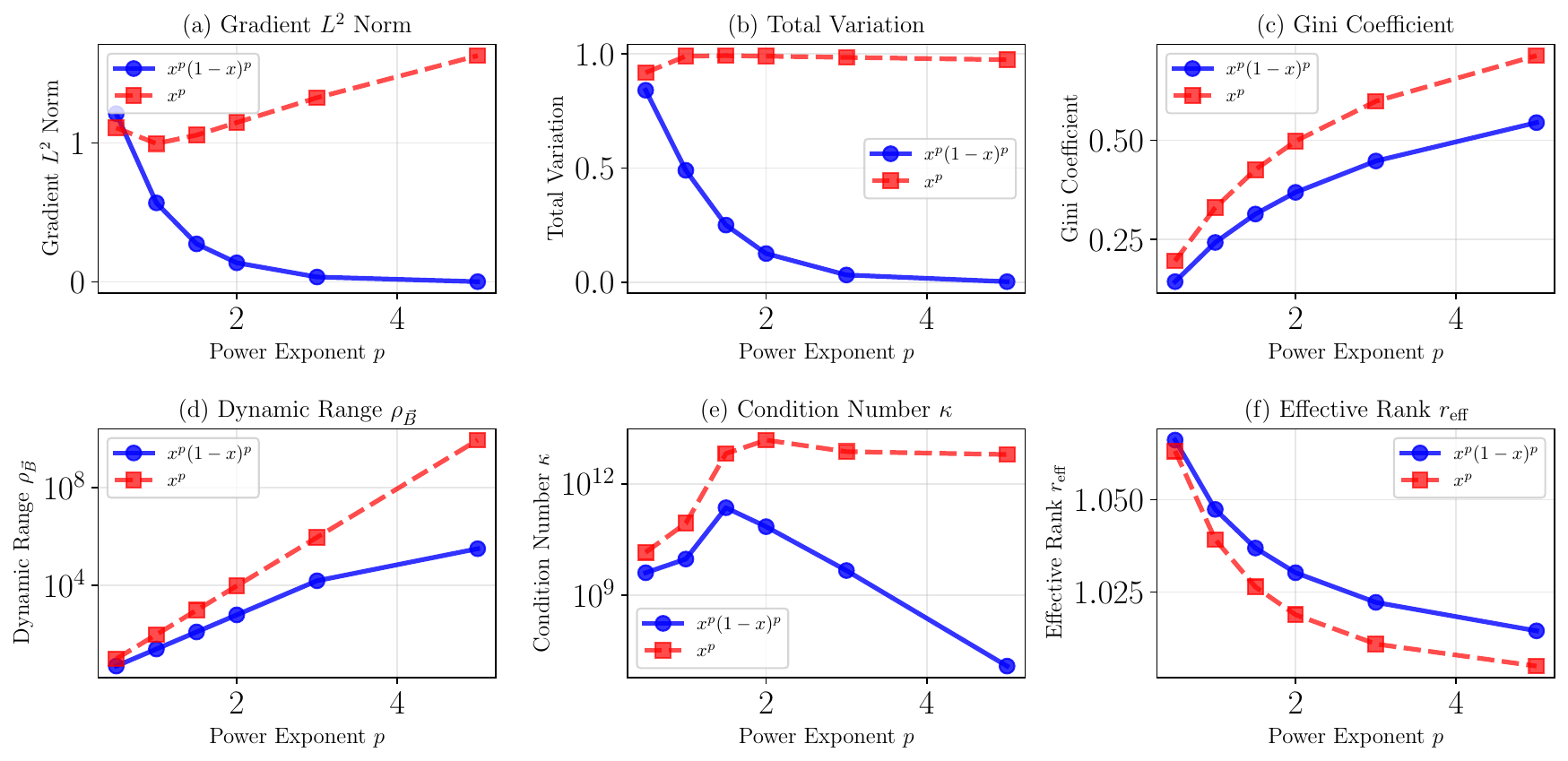}
    \caption{Analysis of power function family characteristics and spectral properties.}
    \label{fig:kt_power_metrics}
\end{figure}

Four characteristic features are computed for each boundary function $B$ to quantify its geometric and distributional properties:
\begin{itemize}
    \item Gradient $L^2$ norm: $\|\nabla B\|_{L^2} = \sqrt{\sum_{i=1}^{N_r} (\partial_x B(x_i))^2}$, measuring the overall rate of variation.
    \item Total variation: $TV(B) = \sum_{i=1}^{N_r-1} |B(x_{i+1}) - B(x_i)|$, quantifying cumulative changes.
    \item Gini coefficient\cite{Dorfman_1979_formulaGinicoefficient}:  $G = \frac{\sum_{i,j} |B_i - B_j|}{2N_r \sum_i B_i}$, quantifying the inequality of diagonal elements.
    \item Dynamic range: $\rho_B = B_{\max}/B_{\min}$.
\end{itemize}
The relationship between these characteristics and the spectral properties (Condition Number $\kappa$, Effective Rank $r_{\text{eff}}$) with respect to the exponent $p$ is summarized in Fig.~\ref{fig:kt_power_metrics}.
Distinct behaviors are observed for the Gradient $L^2$ norm and Total Variation between the symmetric and asymmetric families. Specifically, a monotonic decrease is shown by the symmetric functions, whereas a non-monotonic trend is exhibited by the asymmetric ones.
In contrast, a monotonic increase with $p$ is displayed by both the Gini coefficient and Dynamic Range $\rho_B$, reflecting the increasingly non-uniform value distributions.
Regarding spectral properties, a complex non-monotonic behavior is exhibited by the condition number $\kappa$, suggesting it is influenced by the interplay of multiple factors.
Conversely, a systematic decrease is observed for the effective rank $r_{\text{eff}}$, indicating that higher-power boundary functions lead to more rank-deficient kernels.

\begin{figure}[!htb]
    \centering
    \includegraphics[width=0.95\textwidth]{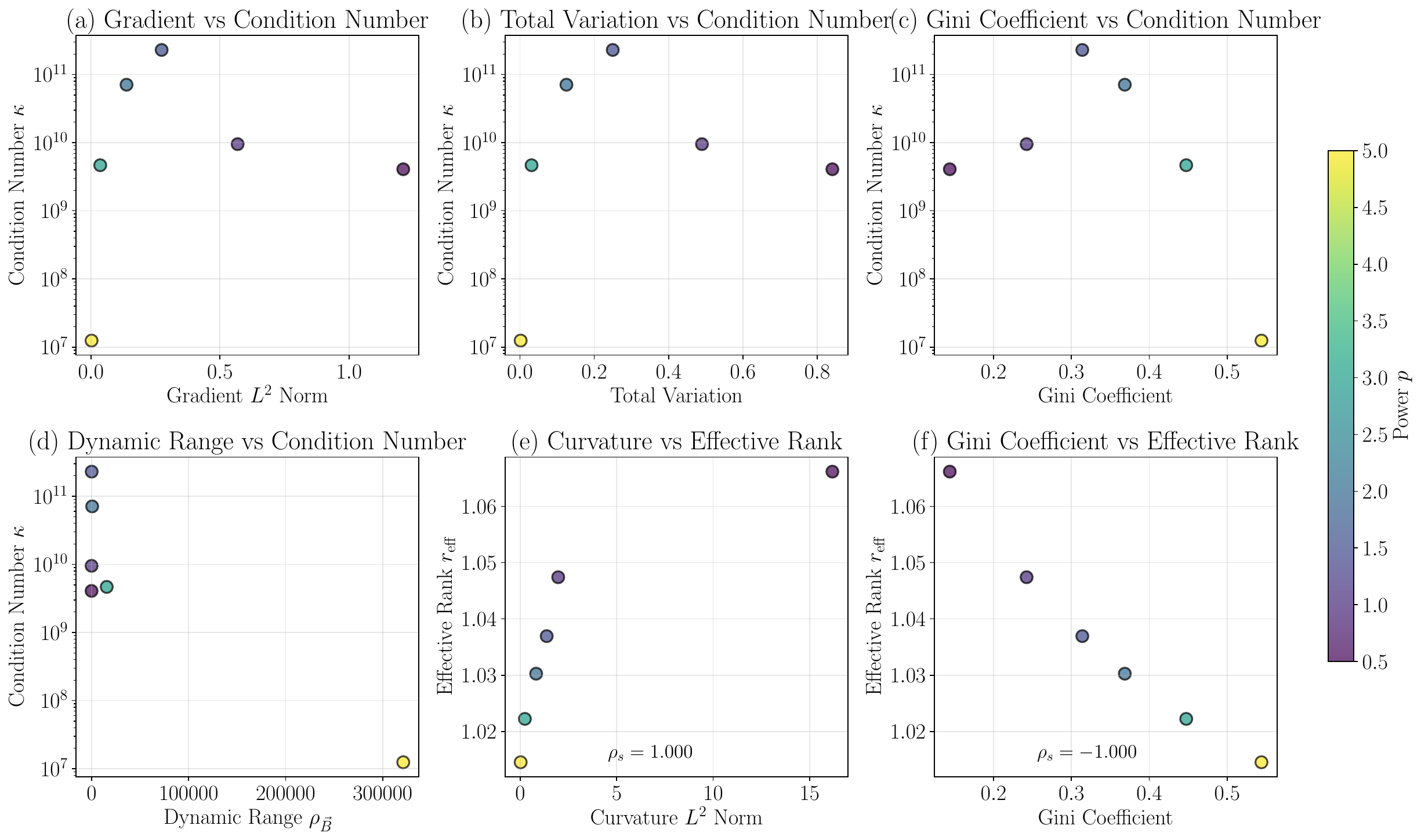}
    \caption{Correlation analysis between boundary function characteristics and spectral properties.}
    \label{fig:kt_correlation}
\end{figure}

The correlations between boundary function characteristics and spectral properties are quantified in Fig.~\ref{fig:kt_correlation}.
For the condition number $\kappa$, no strong correlation with any single feature is observed ($\rho_s \approx 0$).
This is attributed to the intrinsic inverse relationship between the rate-of-change metrics (Gradient $L^2$, TV) and the distribution uniformity metric (Gini coefficient), which cancels out simple linear dependencies.
In contrast, strong correlations are exhibited by the effective rank $r_{\text{eff}}$, particularly with the curvature $L^2$ norm ($\rho_s = 1.0$) and Gini coefficient ($\rho_s \approx -1.0$).
These findings suggest that the rank of the NTK matrix is better preserved by boundary functions with larger second derivatives and more uniform value distributions.
Consequently, the effective rank $r_{\text{eff}}$ is identified as a more reliable predictor of spectral quality than the condition number for HC-PINN trial functions.

\subsection{Analysis of impact of boundary function \texorpdfstring{$B$}{B} on \texorpdfstring{$\bm{K_r}$}{Kr}'s properties}
\label{sec_ntk_res_spectrum}

While Section~\ref{sec_ntk_trial_spectrum} analyzed the trial function NTK $\bm{K}_t$, the residual NTK $\bm{K}_r$ governs the actual training dynamics of HC-PINN, since the loss function involves PDE residuals. 
This section investigates how the boundary function $B$, and particularly its derivatives, influence $\bm{K}_r$ and its spectral properties.

\subsubsection{Boundary function $B$ and its derivatives}

A symmetric power function family is selected as boundary functions:
\begin{equation}
    B(x) = x^\alpha(1-x)^\alpha, \quad \alpha \in \{0.5, 1.0, 1.5, 2.0, 3.0, 5.0\}.
\end{equation}
which corresponds to BCs $u(0) = u(1) = 0$.
This family exhibits diverse characteristics: for $\alpha < 1$, derivatives have boundary singularities; $\alpha = 1$ gives a popular diffusion boundary function; and larger $\alpha$ values create increasingly concentrated functions near the domain center with sharper second derivatives.
The analytical expressions for derivatives are:
\begin{align}
    B'(x) &= \alpha x^{\alpha-1}(1-x)^{\alpha-1}(1-2x), \\
    B''(x) &= \alpha(\alpha-1)x^{\alpha-2}(1-x)^\alpha - 2\alpha^2 x^{\alpha-1}(1-x)^{\alpha-1} + \alpha(\alpha-1)x^\alpha(1-x)^{\alpha-2}.
\end{align}

\begin{figure}[!htb]
    \centering
    \includegraphics[width=0.95\textwidth]{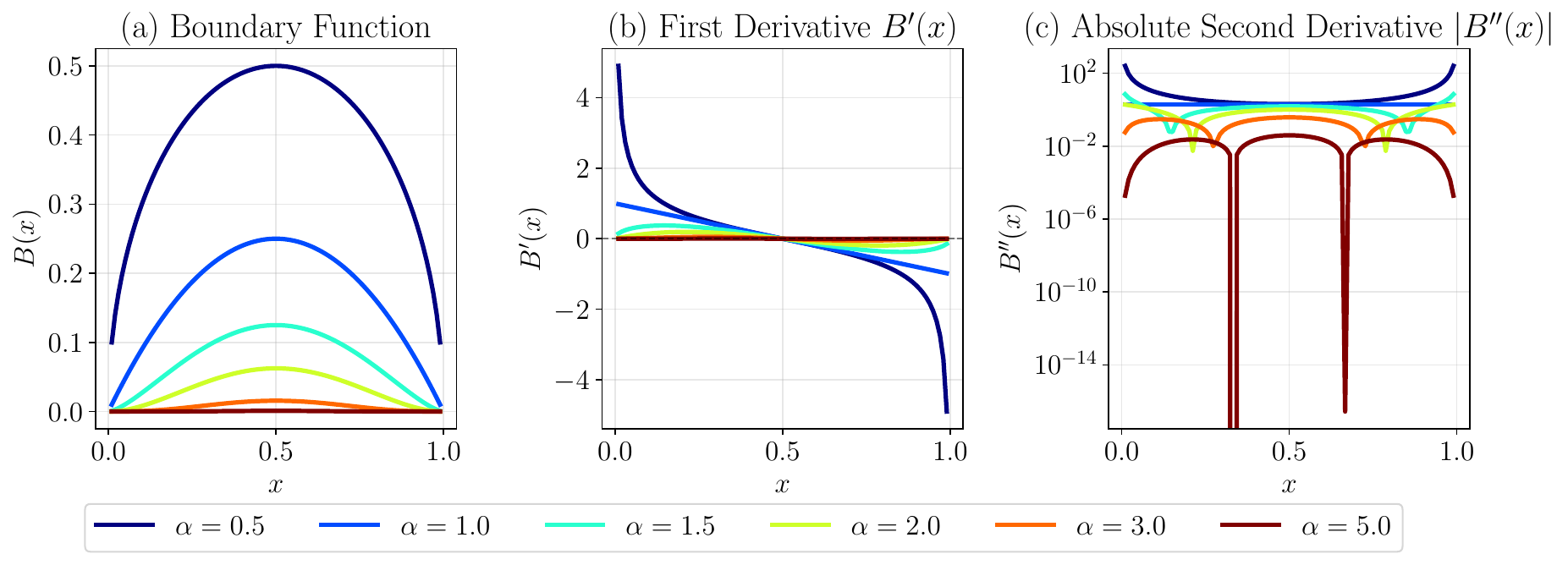}
    \caption{Boundary function $B(x) = x^\alpha(1-x)^\alpha$ and its derivatives for different power exponents $\alpha$. (a) Boundary function $B(x)$, symmetric about $x=0.5$. (b) First derivative $B'(x)$, vanishing at the symmetry point. (c) Absolute second derivative $|B''(x)|$ in log scale, showing dramatic growth for $\alpha=0.5$ near boundaries.}
    \label{fig:kr_boundary_functions}
\end{figure}

Fig.~\ref{fig:kr_boundary_functions} visualizes the boundary function family and its first two derivatives.
Several key observations emerge from Fig.~\ref{fig:kr_boundary_functions}: 
As $\alpha$ increases, $B(x)$ becomes more concentrated near the center (Fig.~\ref{fig:kr_boundary_functions}(a)), with maximum value $B(0.5) = 0.5^{2\alpha}$ decaying exponentially. 
For $\alpha=0.5$ (Fig.~\ref{fig:kr_boundary_functions}(c)), the second derivative exhibits singularities near boundaries, reaching $\max|B''| \approx 250$, while for $\alpha \geq 1$ the second derivatives remain bounded and decrease with increasing $\alpha$ ($\max|B''| \approx 2$ for $\alpha=1$ to $\max|B''| \approx 0.04$ for $\alpha=5$). 
This wide range of second derivative magnitudes (spanning five orders of magnitude) provides an ideal testbed for studying the impact of $B''$ on $\bm{K}_r$'s conditioning.

\subsubsection{Spectral properties of \texorpdfstring{$\bm{K_r}$}{Kr} for different boundary functions $B$}

\begin{figure}[!htb]
    \centering
    \includegraphics[width=0.95\textwidth]{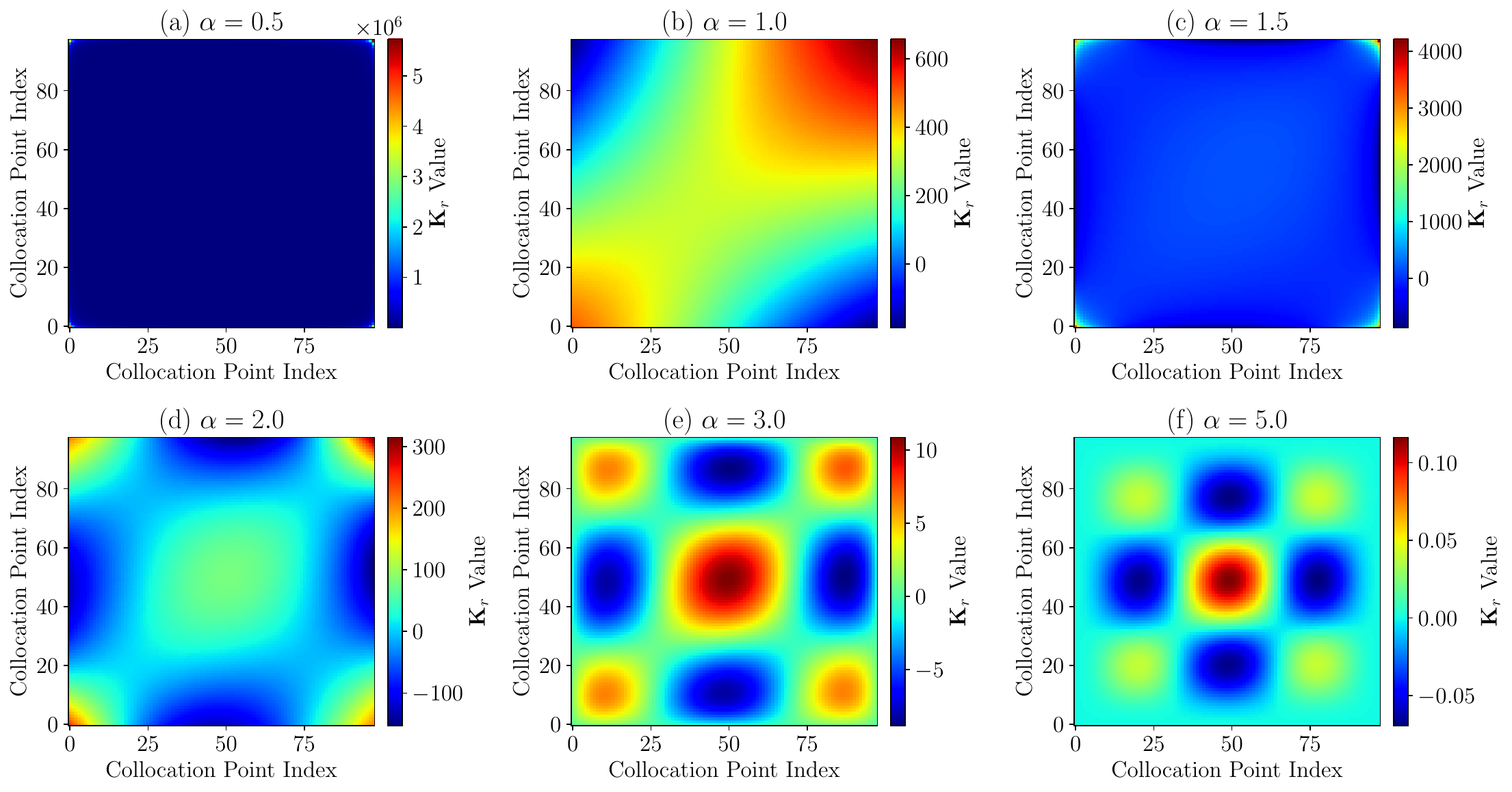}
    \caption{Heatmaps of residual NTK matrix $\bm{K}_r$ for different boundary functions. Colors represent matrix values.}
    \label{fig:kr_heatmaps}
\end{figure}

The PDE considered is the 1D Poisson equation, $u'' = f$, where $f = (\pi^2) \sin(\pi x)$.
The residual NTK $\bm{K}_r$ exhibits fundamentally different structure compared to $\bm{K}_t$ due to the involvement of second derivatives in the PDE residual $\mathcal{R}(x) = \tilde{u}''(x) - f(x)$. Fig.~\ref{fig:kr_heatmaps} visualizes the matrix structure for all tested $\alpha$ values.
Dramatic structural changes are evident: For $\alpha=0.5$ (Fig.~\ref{fig:kr_heatmaps}(a)), the matrix exhibits strong positive values, reflecting the singular behavior of $B''$ near boundaries.
As $\alpha$ increases, the overall magnitude drops precipitously, spanning eight orders of magnitude.
The off-diagonal interactions progressively weaken for larger $\alpha$, indicating more localized coupling between collocation points.

\begin{figure}[!htb]
    \centering
    \includegraphics[width=0.7\textwidth]{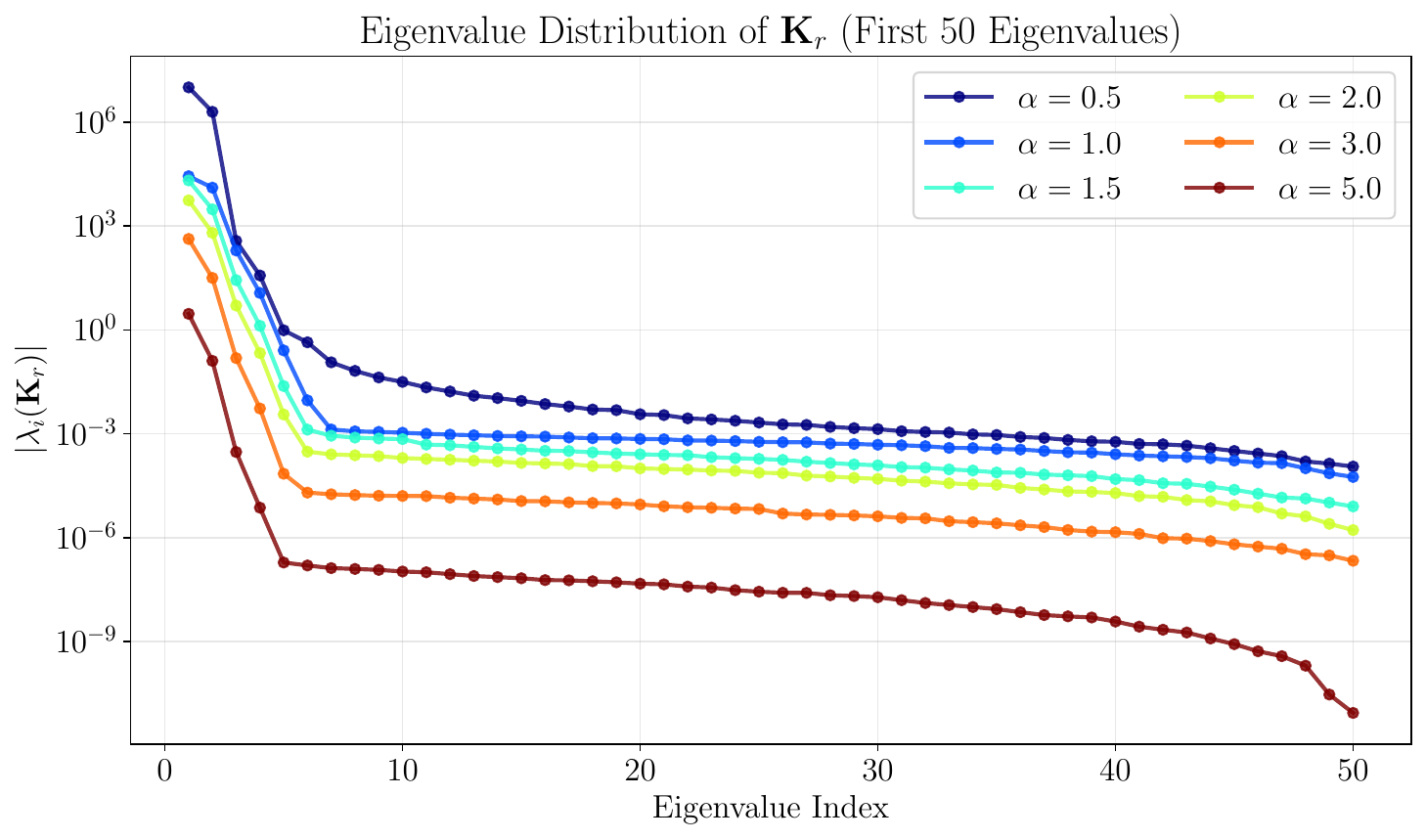}
    \caption{Eigenvalue distributions of $\bm{K}_r$ for different boundary functions (first 50 eigenvalues shown)}
    \label{fig:kr_eigenvalues}
\end{figure}

Fig.~\ref{fig:kr_eigenvalues} presents the eigenvalue spectra of $\bm{K}_r$ for the first \num{50} eigenvalues.
All eigenvalue distributions exhibit steep exponential decay.
For $\alpha=0.5$, the spectrum spans an exceptionally wide range, yielding an extreme condition number $\kappa \approx 3.19 \times 10^{12}$.
Larger $\alpha$ values produce more moderate spectra.

\begin{figure}[!htb]
    \centering
    \includegraphics[width=0.95\textwidth]{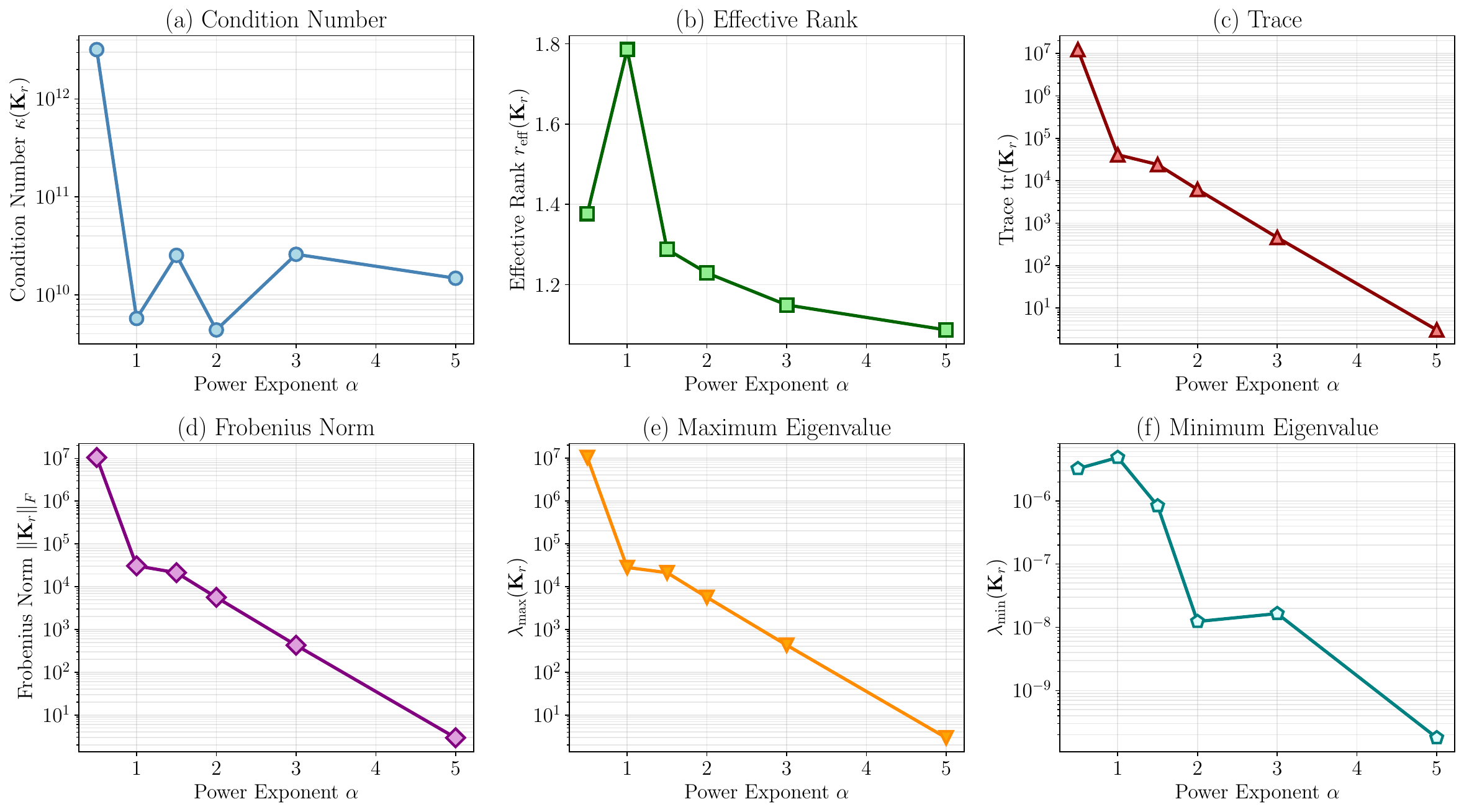}
    \caption{Spectral properties of $\bm{K}_r$ versus power exponent $\alpha$. (a) Condition number exhibits non-monotonic behavior. (b) Effective rank decreases monotonically. (c-d) Trace and Frobenius norm decrease dramatically. (e-f) Maximum and minimum eigenvalues show different trends, driving condition number variations.}
    \label{fig:kr_spectral_metrics}
\end{figure}

To systematically quantify the impact of $\alpha$ on $\bm{K}_r$'s spectral properties, six key metrics are tracked: condition number $\kappa$, effective rank $r_{\mathrm{eff}}$, trace, Frobenius norm $\|\bm{K}\|_F = \sqrt{\sum_{i,j} K_{ij}^2}$ \cite{Bhatia_2013_Matrixanalysis}, maximum eigenvalue $\lambda_{\max}$, and minimum eigenvalue $\lambda_{\min}$. Fig.~\ref{fig:kr_spectral_metrics} illustrates how these metrics evolve with $\alpha$.
The condition number (Fig.~\ref{fig:kr_spectral_metrics}(a)) exhibits non-monotonic behavior, indicating that conditioning is governed by the complex interplay between $\lambda_{\max}$ and $\lambda_{\min}$.
The effective rank (Fig.~\ref{fig:kr_spectral_metrics}(b)) exhibits non-monotonic behavior with a peak at $\alpha=1$, suggesting that moderate boundary function concentration provides the most balanced eigenvalue distribution.
The trace and Frobenius norm (Fig.~\ref{fig:kr_spectral_metrics}(c-d)) both decrease dramatically, directly reflecting the shrinking magnitude of $B''$ as $\alpha$ increases.

\subsubsection{Correlation between second derivative and spectral properties}
\label{sec_ntk_res_correlation}

Given the involvement of $B''$ in the residual definition, it is hypothesized that the second derivative's magnitude and variation directly impact $\bm{K}_r$'s conditioning and effective rank. To test this, three characteristic features of $B''$ are computed:
\begin{itemize}
    \item Maximum absolute value: $\max_x |B''(x)|$, measuring peak curvature
    \item $L^2$ norm: $\|B''\|_{L^2} = \sqrt{\sum_{i=1}^{N_r} (B''(x_i))^2}$, measuring overall magnitude
    \item Total variation: $\mathrm{TV}(B'') = \sum_{i=1}^{N_r-1} |B''(x_{i+1}) - B''(x_i)|$, measuring roughness
\end{itemize}

\begin{figure}[!htb]
    \centering
    \includegraphics[width=0.95\textwidth]{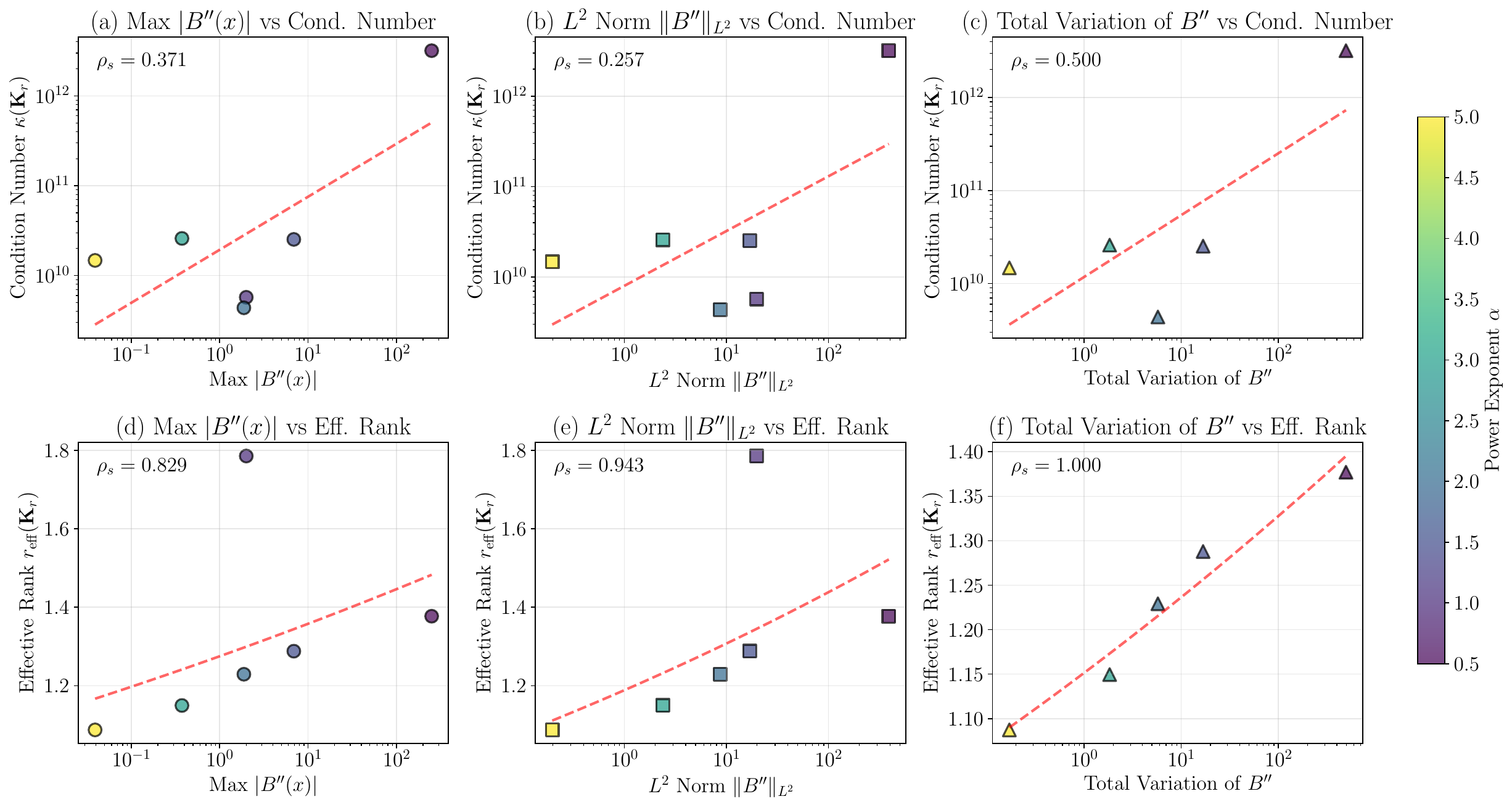}
    \caption{Correlation analysis between $B''$ characteristics and spectral properties of $\bm{K}_r$. Top row: relationships with condition number $\kappa$. Bottom row: relationships with effective rank $r_{\mathrm{eff}}$. Points colored by $\alpha$, with red dashed lines showing fits. Spearman correlation coefficients $\rho_s$ shown in each panel. Note that Total Variation plots exclude $\alpha=1$ (where TV=0).}
    \label{fig:kr_correlation}
\end{figure}

Fig.~\ref{fig:kr_correlation} presents scatter plots analyzing correlations between $B''$ features and two key spectral properties: condition number $\kappa(\bm{K}_r)$ and effective rank $r_{\mathrm{eff}}(\bm{K}_r)$.
For condition number (top row of Fig.~\ref{fig:kr_correlation}): All three $B''$ features show weak positive correlations with $\kappa$, but none are statistically significant.
This weak correlation confirms that conditioning exhibits complex non-monotonic behavior not simply predictable from $B''$ magnitude alone.
For effective rank (bottom row of Fig.~\ref{fig:kr_correlation}): Strong negative correlations emerge between all $B''$ features and $r_{\mathrm{eff}}$ (Spearman $\rho_s \approx -0.9$ to $-1.0$), indicating highly predictable relationships: larger/rougher second derivatives lead to more rank-deficient matrices.
This finding implies that boundary functions with smoother second derivatives preserve better matrix rank structure.

\subsubsection{Summary of \texorpdfstring{$\bm{K_r}$}{Kr} analysis}

The comprehensive analysis of residual NTK $\bm{K}_r$ reveals fundamental differences from trial function NTK $\bm{K}_t$:
\begin{enumerate}
    \item Matrix structure and magnitude: The residual NTK exhibits dramatically different magnitudes across boundary functions, spanning eight orders of magnitude. This extreme sensitivity arises because $\bm{K}_r$ involves $B''$ through the trial function's second derivative in the residual operator.
    \item Spectral property degradation: As $\alpha$ increases and $B(x)$ becomes more concentrated, the effective rank $r_{\mathrm{eff}}$ decreases monotonically, indicating progressively more rank-deficient matrices. The condition number exhibits non-monotonic behavior but remains in the challenging range $10^9$-$10^{12}$.
    \item Second derivative as critical factor: The correlation analysis reveals that $B''$ characteristics are strong predictors of effective rank ($\rho_s \approx -0.9$ to $-1.0$) but weak predictors of condition number.
    \item Practical implications: For HC-PINN practitioners, these findings emphasize the critical importance of boundary function selection: Functions with moderate $\alpha$ (around \num{1.0}-\num{1.5}) offer a reasonable compromise. Extreme choices should be avoided unless problem-specific considerations override conditioning concerns.
\end{enumerate}

\subsection{Training Dynamics Analysis on Simple Problems}
\label{sec_simple_train}

\subsubsection{Training Dynamics with SGD and Adam + L-BFGS Optimizers}
\label{sec_optimizers}

As discussed in Section~\ref{subsec_ntk_limitations}, NTK analysis relies on assumptions like infinitesimal learning rates (lazy training). To evaluate the practical efficiency of this theoretical regime versus modern practices, NTK-theory-based SGD is compared against the standard Adam + L-BFGS strategy for HC-PINN training.

The 1D Poisson equation $u''(x) = \pi^2 \sin(\pi x)$ with $u(0)=u(1)=0$ is solved using $B(x) = x(1-x)$. Two strategies are tested:
\begin{enumerate}
    \item NTK-theory SGD: Vanilla SGD with $\eta = 10^{-5}$ for \num{20000} epochs, following NTK literature \cite{WangWhenWhyPINNs2022}.
    \item Adam + L-BFGS: Adam ($\eta = 10^{-3}$, \num{2000} epochs) followed by L-BFGS ($500$ epochs), a standard approach for PINNs \cite{XiePhysicsspecializedNeuralNetwork2024}.
\end{enumerate}

\begin{figure}[!htb]
    \centering
    \includegraphics[width=\textwidth]{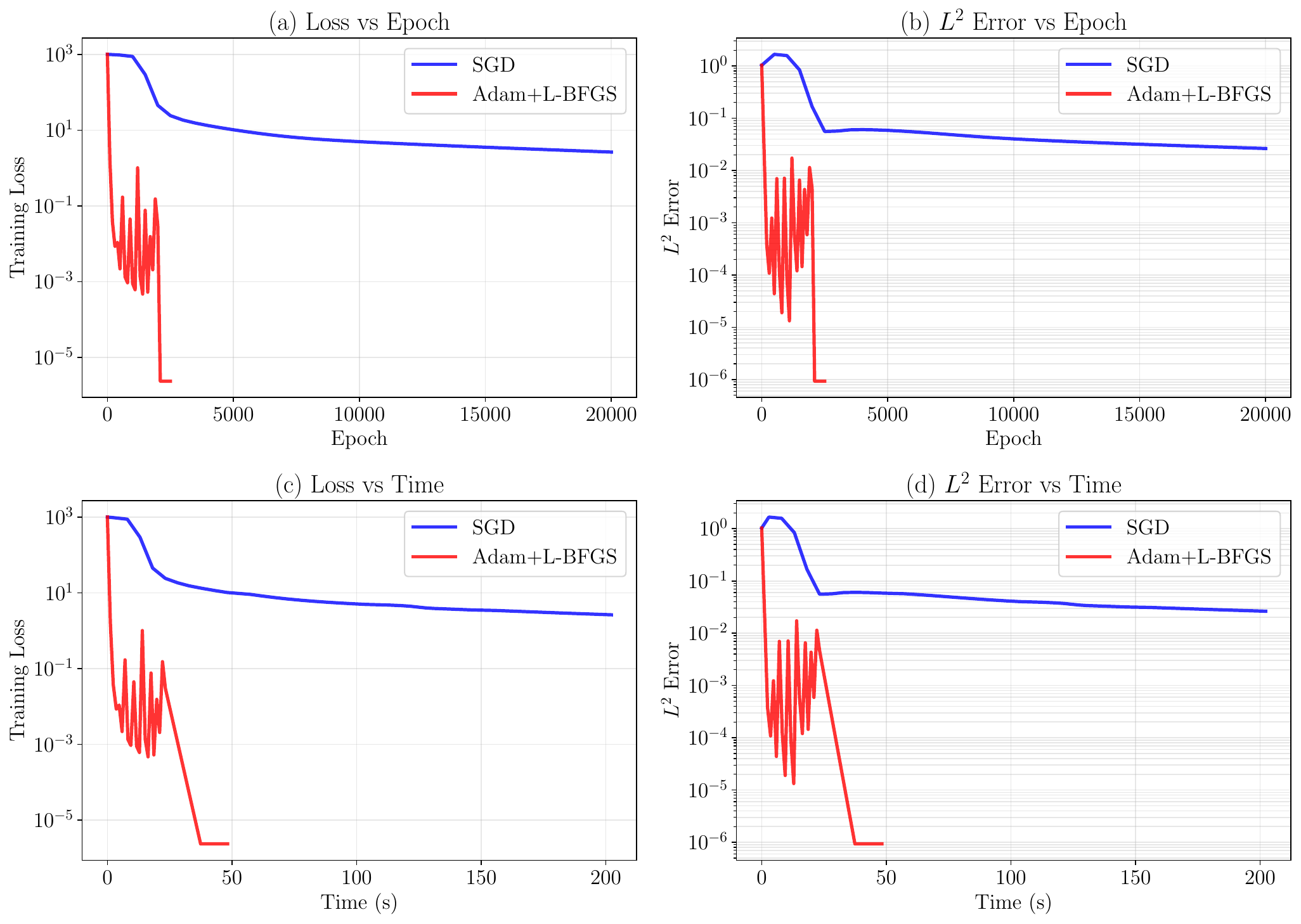}
    \caption{Training dynamics comparison. (a, c) Loss vs epoch/time. (b, d) $L^2$ error vs epoch/time.}
    \label{fig:optimizer_training_comparison}
\end{figure}

The performance is compared in Fig.~\ref{fig:optimizer_training_comparison} and Table~\ref{tab:optimizer_performance}. SGD is observed to be extremely slow, achieving a final loss of only $2.64$ after \num{20000} epochs ($202$s). In contrast, rapid convergence is achieved by Adam + L-BFGS, reaching a loss of $2.36 \times 10^{-6}$ in just $48$s ($2500$ epochs). The combined optimizer is found to be $4.2\times$ faster and achieves six orders of magnitude better accuracy, demonstrating that the theoretical small learning rate is impractical for training.

\begin{table}[!htb]
    \centering
    \caption{Efficiency comparison: SGD vs. Adam + L-BFGS}
    \label{tab:optimizer_performance}
    \begin{tabular}{lccc}
        \toprule
        Metric & SGD & Adam + L-BFGS & Ratio \\
        \midrule
        Total Epochs & \num{20000} & \num{2500} & $8.0\times$ \\
        Time (s) & \num{202.24} & \num{48.22} & $4.2\times$ \\
        Final Loss & $2.64$ & $2.36 \times 10^{-6}$ & $\num{1.1e6}\times$ \\
        Final $L^2$ Error & $2.63 \times 10^{-2}$ & $9.33 \times 10^{-7}$ & $\num{2.8e4}\times$ \\
        \bottomrule
    \end{tabular}
\end{table}

\begin{figure}[p]
    \centering
    \includegraphics[width=0.9\textwidth]{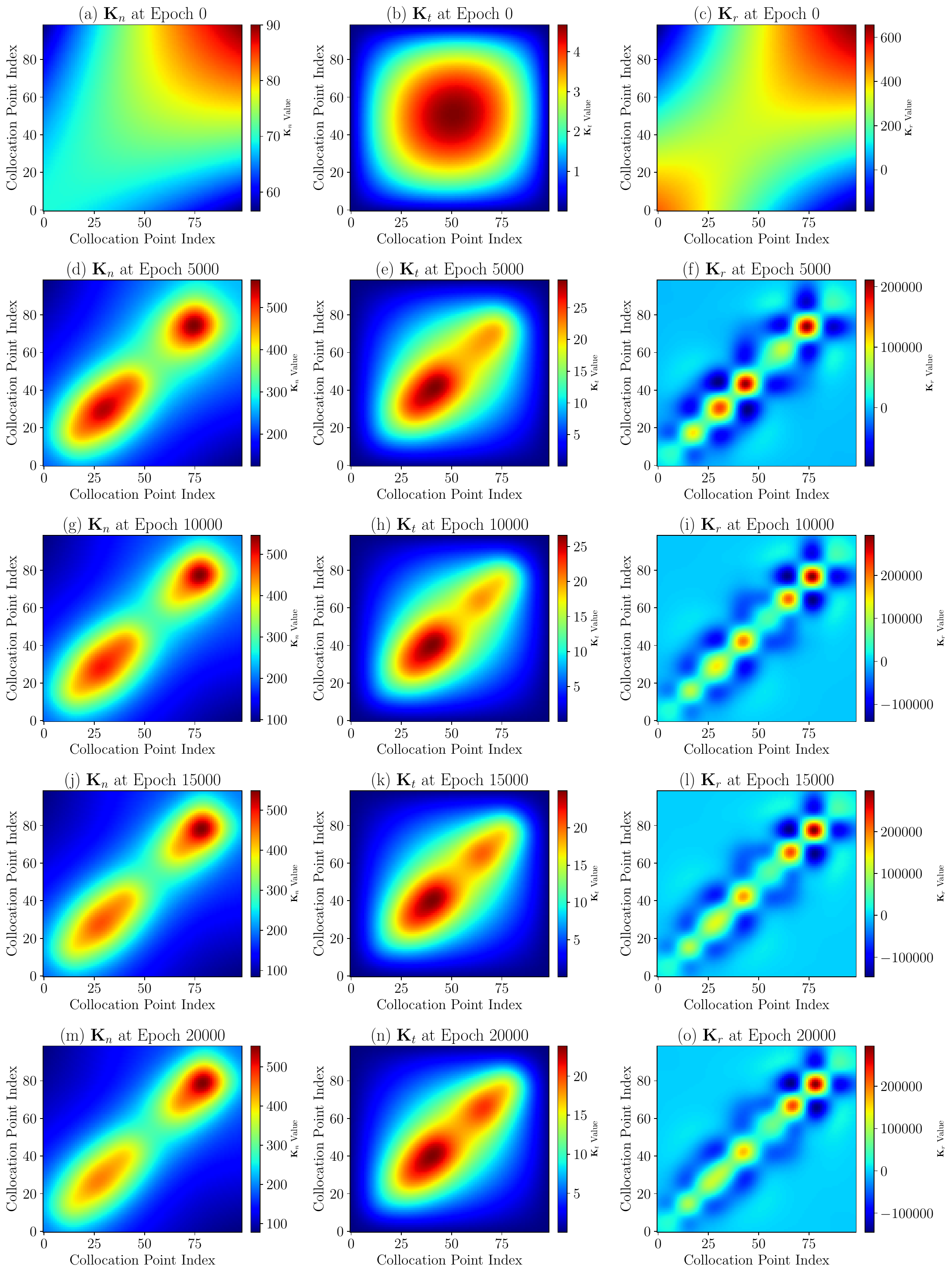}
    \caption{NTK matrix evolution during SGD training.}
    \label{fig:ntk_evolution_sgd}
\end{figure}

To understand the mechanism, the evolution of NTK matrices ($\bm{K}_n, \bm{K}_t, \bm{K}_r$) is tracked.
It is shown in Fig.~\ref{fig:ntk_evolution_sgd} that under SGD, the matrices evolve initially (first \num{5000} epochs) but then stabilize, broadly consistent with the lazy training regime. A progressive diagonalization is observed, particularly in $\bm{K}_r$.
However, this regime is inefficient.

\begin{figure}[!htb]
    \centering
    \includegraphics[width=\textwidth]{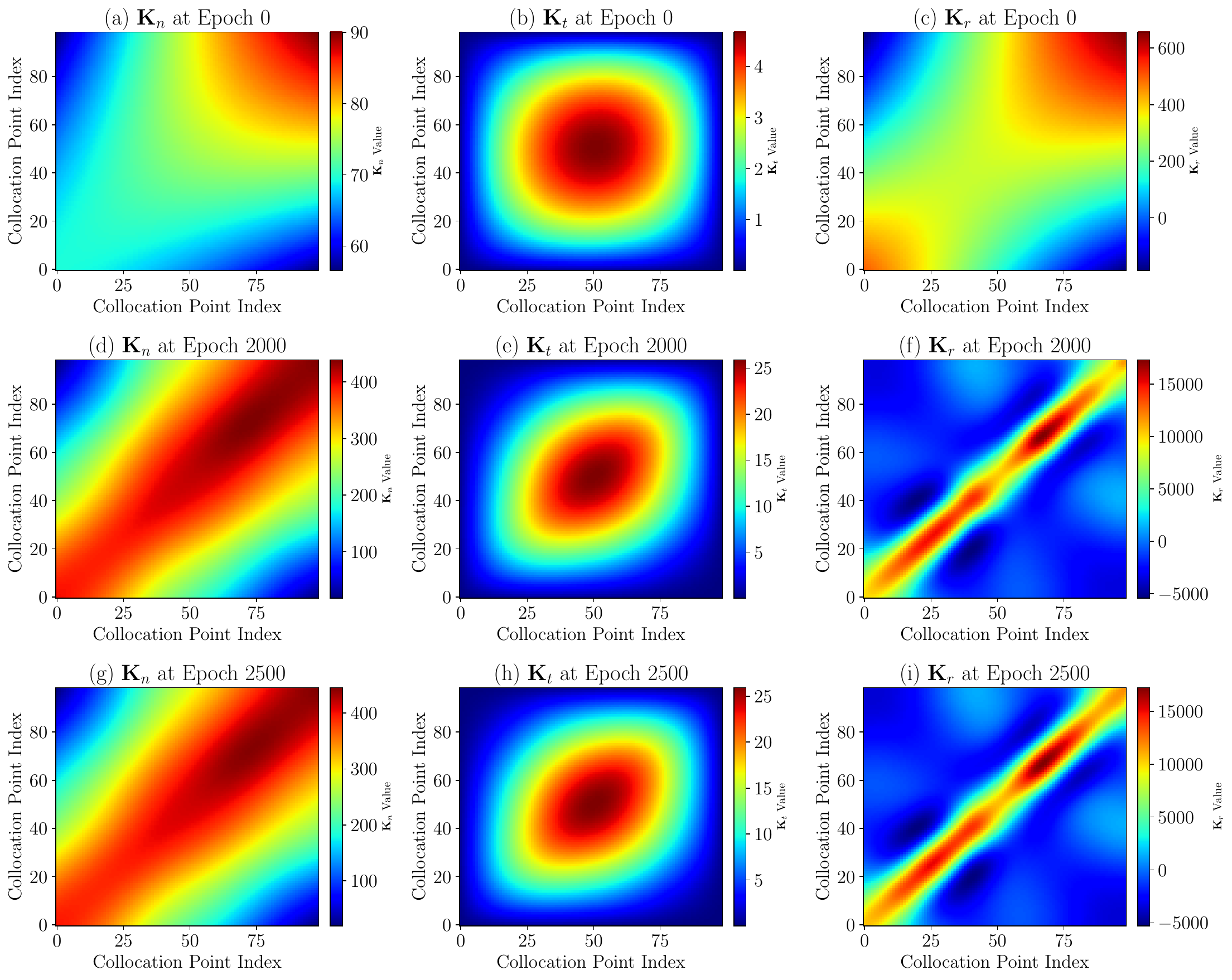}
    \caption{NTK matrix evolution during Adam + L-BFGS training.}
    \label{fig:ntk_evolution_adam_bfgs}
\end{figure}

Under Adam + L-BFGS (Fig.~\ref{fig:ntk_evolution_adam_bfgs}), dramatic structural changes are observed in the NTK matrices, indicating a departure from lazy training into the feature learning regime. The residual NTK $\bm{K}_r$ becomes highly diagonal, and its effective rank increases significantly (Table~\ref{tab:spectral_properties}), allowing the adaptation of the network representation to the problem.

\begin{table}[!htb]
    \centering
    \caption{Spectral properties of $\bm{K}_r$ at initialization and end of training}
    \label{tab:spectral_properties}
    \begin{tabular}{lcccc}
        \toprule
        & \multicolumn{2}{c}{SGD} & \multicolumn{2}{c}{Adam + L-BFGS} \\
        \cmidrule(lr){2-3} \cmidrule(lr){4-5}
        Property & Init  & Final & Init  & Final  \\
        \midrule
        $\kappa$ & $2.73 \times 10^9$ & $1.78 \times 10^{10}$ & $2.73 \times 10^9$ & $4.46 \times 10^9$ \\
        $r_{\mathrm{eff}}$ & $1.79$ & $4.76$ & $1.79$ & $6.86$ \\
        $\|\bm{K}_r\|_F$ & $3.10 \times 10^4$ & $3.99 \times 10^6$ & $3.10 \times 10^4$ & $5.03 \times 10^5$ \\
        \bottomrule
    \end{tabular}
\end{table}

In conclusion, while a lazy training regime is assumed by NTK theory which is computationally inefficient, feature learning is leveraged by practical training with Adam + L-BFGS for superior performance. Crucially, as shown in Section~\ref{sec_B_influence_training}, the \textit{initial} spectral properties of $\bm{K}_r$ remain powerful predictors of training difficulty even when using aggressive optimizers. Thus, NTK analysis serves as a valuable diagnostic tool for problem conditioning, while Adam + L-BFGS is recommended for the actual training.

\subsubsection{One Dimensional Case}
\label{sec_simple_B_case_2}
\label{sec_B_influence_training}

The diffusion equation is chosen due to its fundamental importance in modeling transport phenomena and its ubiquity as a building block for more complex partial differential equations.
The 1D diffusion equation is first investigated:
\begin{equation}
   -\frac{\partial^2 u}{\partial x^2} = f(x), \quad x \in [0, 1],
\end{equation}
with homogeneous Dirichlet BCs $u(0) = u(1) = 0$. The analytical solution is given by $u(x) = \sin(\pi x)\cos(2\pi x)$.
The trial function is defined as $\tilde{u}(x, \theta) = B(x) N(x, \theta)$.
Five parametric families of boundary functions are examined:
\begin{enumerate}
    \item Power: $B(x) = x^\alpha(1-x)^\alpha$, with $\alpha \in \{0.5, 1.0, 1.5, 2.0, 2.5\}$.
    \item Trigonometric: $B(x) = \sin^\alpha(\pi x)$, with $\alpha \in \{1, 2, 3, 4, 5\}$.
    \item Rational: $B(x) = \frac{x(1-x)}{1 + \alpha x(1-x)}$, with $\alpha \in \{0, 2, 5, 10, 20\}$.
    \item Exponential: $B(x) = x(1-x)e^{-\alpha x(1-x)}$, with $\alpha \in \{0, 2, 5, 10, 20\}$.
    \item Hyperbolic tangent: $B(x) = \tanh(\alpha x)\tanh(\alpha(1-x))$, with $\alpha \in \{1, 3, 5, 7, 10\}$.
\end{enumerate}

Twenty-five experiments are designed. Four experiments (Power $\alpha = 2.0$, Trigonometric $\alpha = 3, 5$, Tanh $\alpha = 10.0$) failed due to numerical instability. The analysis is conducted on the remaining 21 experiments.

\begin{figure}[!htb]
    \centering
    \includegraphics[width=0.95\textwidth]{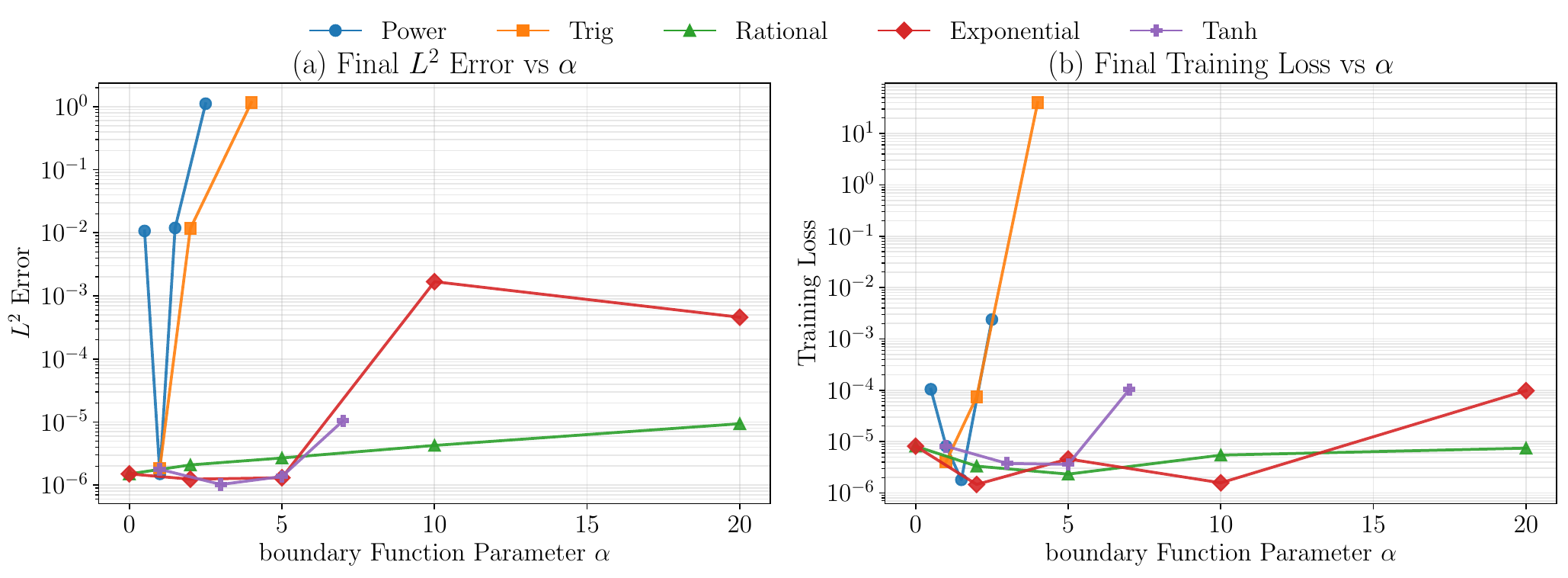}
    \caption{Training performance across different boundary function families. (a) Final $L^2$ errors vs parameter $\alpha$. (b) Final training loss vs parameter $\alpha$.}
    \label{fig:boundary_family_comparison}
\end{figure}

Fig.~\ref{fig:boundary_family_comparison} shows the final $L^2$ errors and training losses.
Significant performance differences are observed. Rational, Tanh, and Exponential families consistently achieve errors below $10^{-3}$, whereas Power and Trigonometric families show high parameter dependence.
Tanh achieves the best mean performance ($L^2 = 3.66 \times 10^{-6}$), followed by Rational and Exponential. This represents a significant improvement over the Trigonometric family.
Rational and Tanh families exhibit robustness to parameter variations, while Power and Trigonometric families display extreme sensitivity.
It is noted that certain Power parameters (e.g., $\alpha = 1.0$) yield high accuracy.
Training loss patterns in Fig.~\ref{fig:boundary_family_comparison}(b) are consistent with these observations.

Table~\ref{tab:boundary_family_summary} summarizes the training performance statistics. It is confirmed that appropriate boundary functions can significantly reduce errors.

\begin{table}[!htb]
    \centering
    \caption{Training performance statistics across boundary function families}
    \label{tab:boundary_family_summary}
    \begin{tabular}{lcccc}
        \toprule
        Family & $L^2$ Error Mean & $L^2$ Error Std & Loss Mean & Loss Std \\
        \midrule
        Power & $2.83 \times 10^{-1}$ & $5.52 \times 10^{-1}$ & $6.25 \times 10^{-4}$ & $1.17 \times 10^{-3}$ \\
        Trigonometric & $3.92 \times 10^{-1}$ & $6.69 \times 10^{-1}$ & $1.34 \times 10^{1}$ & $2.33 \times 10^{1}$ \\
        Rational & $\mathbf{3.99 \times 10^{-6}}$ & $3.20 \times 10^{-6}$ & $5.34 \times 10^{-6}$ & $2.52 \times 10^{-6}$ \\
        Exponential & $4.28 \times 10^{-4}$ & $7.27 \times 10^{-4}$ & $2.28 \times 10^{-5}$ & $4.22 \times 10^{-5}$ \\
        \textbf{Tanh} & $\mathbf{3.66 \times 10^{-6}}$ & $4.55 \times 10^{-6}$ & $2.96 \times 10^{-5}$ & $4.89 \times 10^{-5}$ \\
        \bottomrule
    \end{tabular}
    \\[0.5em]
    \footnotesize{Note: Power family excludes $\alpha=2.0$, Trigonometric excludes $n=3,5$, and Tanh excludes $\lambda=10.0$ due to NTK eigenvalue computation non-convergence.}
\end{table}

The correlation between NTK spectral properties and final training errors is analyzed. Four spectral metrics of the residual NTK matrix $\bm{K}_r$ at initialization are examined: convergence rate $c$, condition number $\kappa$, effective rank $r_{\text{eff}}$, and Frobenius norm $\|\bm{K}_r\|_F$.

\begin{figure}[!htb]
    \centering
    \includegraphics[width=\textwidth]{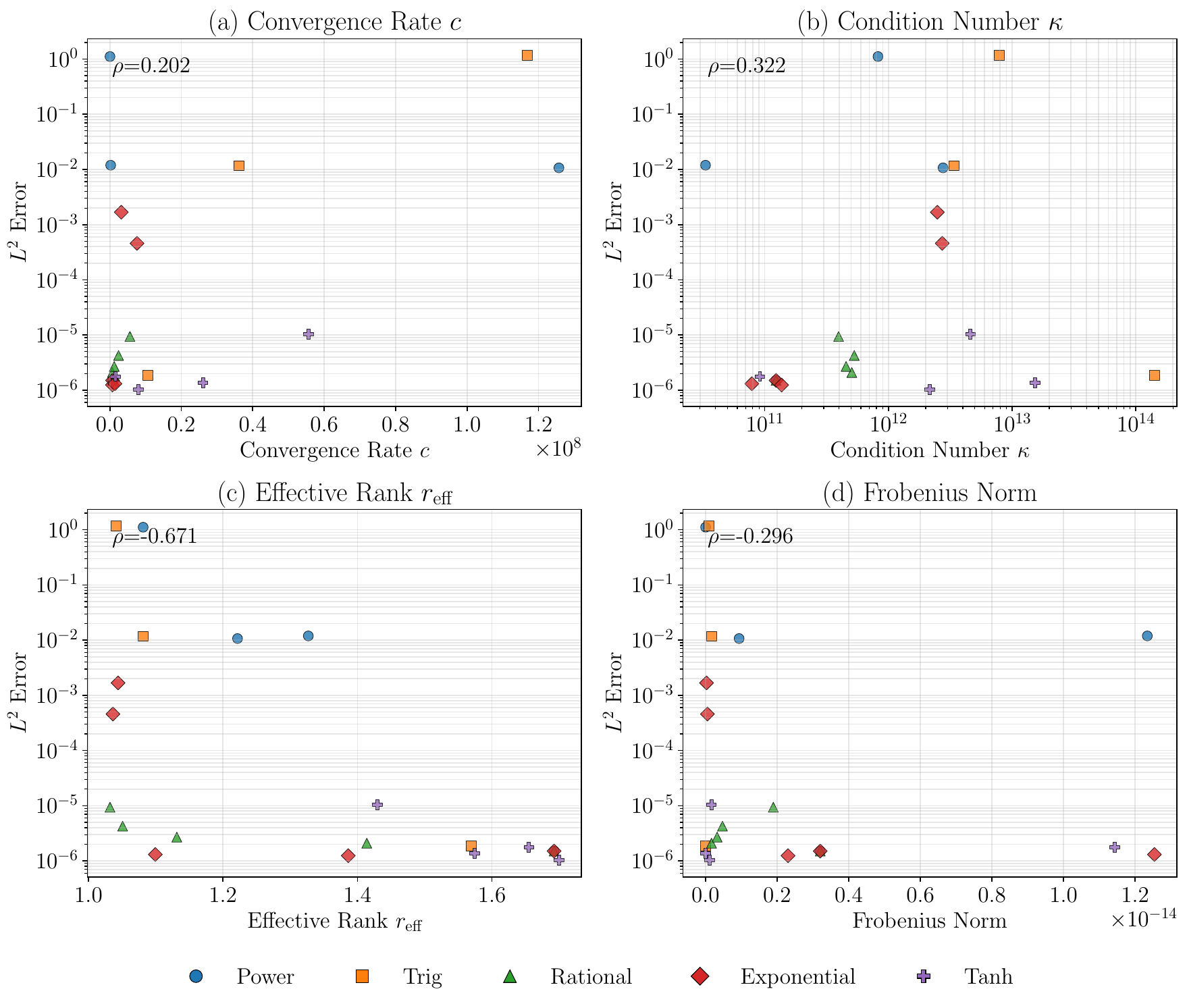}
    \caption{Correlation between $\bm{K}_r$ spectral properties and final $L^2$ errors. (a) Convergence rate $c$. (b) Condition number $\kappa$. (c) Effective rank $r_{\text{eff}}$. (d) Frobenius norm $\|\cdot\|_F$.}
    \label{fig:kr_spectral_correlation}
\end{figure}

Fig.~\ref{fig:kr_spectral_correlation} shows the correlation analysis.
Weak overall correlations are observed. Effective rank $r_{\text{eff}}$ shows the strongest correlation ($\rho = -0.671$), while other metrics exhibit weak correlations.
However, a clustering pattern is revealed in Fig.~\ref{fig:kr_spectral_correlation}(c): high-performing families occupy the high-$r_{\text{eff}}$, low-error region.
This suggests that effective rank is a useful discriminator.
The limitation of aggregate correlation analysis motivates a stratified analysis within each family.

\begin{figure}[!htb]
    \centering
    \includegraphics[width=0.95\textwidth]{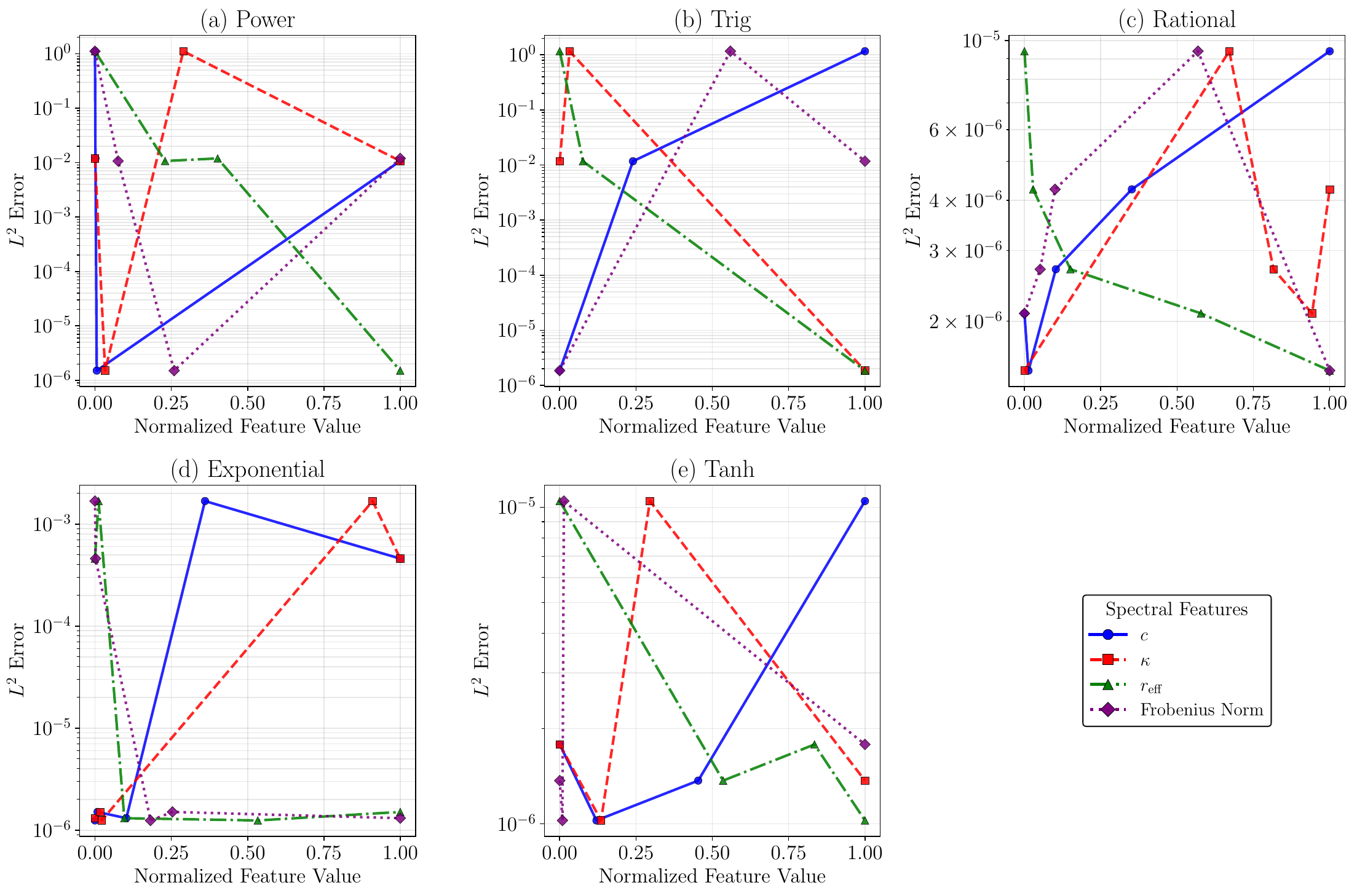}
    \caption{Spectral feature sensitivity analysis within boundary function families. Each panel shows error vs normalized spectral features for one family: (a) Power, (b) Trigonometric, (c) Rational, (d) Exponential, (e) Tanh.}
    \label{fig:spectral_feature_sensitivity}
\end{figure}

Fig.~\ref{fig:spectral_feature_sensitivity} presents the spectral feature sensitivity analysis.
Family-specific patterns are revealed.
The Power family (Fig.~\ref{fig:spectral_feature_sensitivity}(a)) shows strong correlations, with convergence rate $c$ being the most significant. Optimal parameter $\alpha = 1.0$ is associated with well-conditioned $\bm{K}_r$.
The Trigonometric family (Fig.~\ref{fig:spectral_feature_sensitivity}(b)) exhibits severe spectral degradation. Increasing $\alpha$ worsens effective rank and condition number.
The Rational family (Fig.~\ref{fig:spectral_feature_sensitivity}(c)) demonstrates spectral stability with weak correlations.
The Exponential family (Fig.~\ref{fig:spectral_feature_sensitivity}(d)) shows strong negative correlation for $c$, indicating that larger $\alpha$ values improve spectral properties.
The Tanh family (Fig.~\ref{fig:spectral_feature_sensitivity}(e)) shows weak correlations, reflecting its robustness.
It is concluded that spectral feature correlations are family-dependent.

Several key findings are established.
First, performance variations are significant, with smooth families (Rational, Tanh) outperforming oscillatory ones.
Second, stratified within-family analysis reveals strong patterns, with effective rank $r_{\text{eff}}$ being a reliable predictor.
Third, optimal boundary functions favor functional simplicity.
These findings validate NTK spectral analysis as a diagnostic tool.
Smooth boundary functions are recommended for robust performance.

\subsubsection{Two Dimensional Case}
\label{sec:simple_B_case_3}

The analysis is extended to the 2D linear diffusion equation on a unit square:
\begin{equation}
   -\nabla^2 u + u = f(x,y), \quad (x,y) \in [0, 1]^2,
\end{equation}
with homogeneous Dirichlet BCs. The analytical solution is $u(x,y) = \sin(\pi x)\sin(\pi y)$.
The trial function is $\tilde{u}(x, y, \theta) = B(x,y)  N(x, y, \vec \theta)$.
Three parametric families of boundary functions are investigated:
\begin{enumerate}
    \item Power: $B(x,y) = [x(1-x)y(1-y)]^\alpha$, with $\alpha \in \{0.5, 1.0, 1.5, 2.0, 2.5\}$
    \item MixedPower: $B(x,y) = [x(1-x)]^\alpha [y(1-y)]^\beta$, with $(\alpha, \beta) \in \{(0.5, 0.5), (1.0, 1.0), \\(1.5, 1.5), (2.0, 2.0), (1.0, 2.0), (2.0, 1.0)\}$
    \item Hyperbolic tangent: $B(x,y) = \tanh(\alpha x)\tanh(\alpha(1-x))\tanh(\alpha y)\tanh(\alpha(1-y))$, with $\alpha \in \{1, 3, 5, 7, 10\}$
\end{enumerate}
The MixedPower family introduces dimensional anisotropy. Sixteen experiments are designed.

\begin{figure}[!htb]
    \centering
    \includegraphics[width=0.95\textwidth]{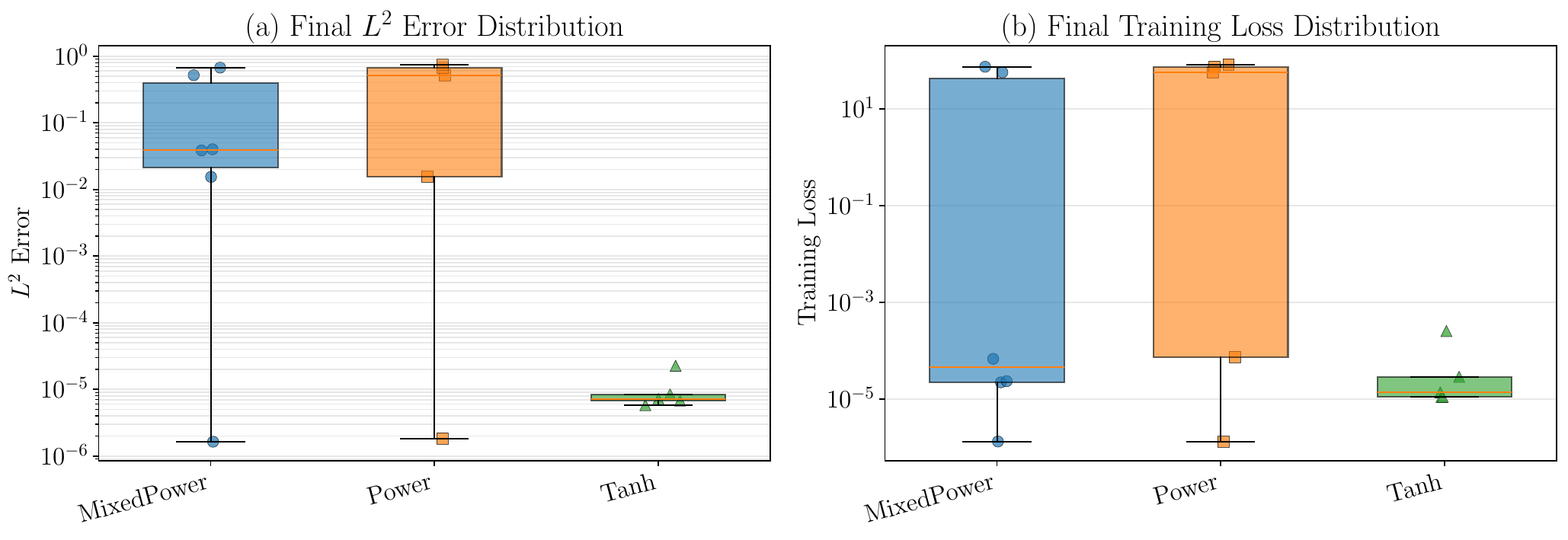}
    \caption{Training performance across different 2D boundary function families. (a) Final $L^2$ errors. (b) Final training loss.}
    \label{fig:boundary_family_comparison_2d}
\end{figure}

Fig.~\ref{fig:boundary_family_comparison_2d} shows the final $L^2$ errors and training losses.
Dramatic performance differences are observed, exceeding those in the 1D case. The Tanh family achieves the best mean performance ($L^2 = 1.02 \times 10^{-5}$), while the Power family performs worst ($L^2 = 3.92 \times 10^{-1}$).
This amplified sensitivity reflects the increased complexity of 2D boundary function construction.
The Tanh family demonstrates exceptional robustness with low error variation.
In contrast, the Power family exhibits extreme parameter sensitivity.
The MixedPower family shows intermediate behavior, but symmetric low-order exponents $(\alpha, \beta) = (1.0, 1.0)$ yield excellent results.

Table~\ref{tab:boundary_family_summary_2d} summarizes the training performance statistics.

\begin{table}[!htb]
    \centering
    \caption{Training performance statistics across 2D boundary function families}
    \label{tab:boundary_family_summary_2d}
    \begin{tabular}{lcccc}
        \toprule
        Family & $L^2$ Error Mean & $L^2$ Error Std & Loss Mean & Loss Std \\
        \midrule
        \multicolumn{5}{c}{\textit{Parametric Families}} \\
        \midrule
        Power & $3.92 \times 10^{-1}$ & $3.61 \times 10^{-1}$ & $4.32 \times 10^{1}$ & $4.06 \times 10^{1}$ \\
        MixedPower & $2.15 \times 10^{-1}$ & $3.00 \times 10^{-1}$ & $2.21 \times 10^{1}$ & $3.47 \times 10^{1}$ \\
        \textbf{Tanh} & $\mathbf{1.02 \times 10^{-5}}$ & $7.03 \times 10^{-6}$ & $6.37 \times 10^{-5}$ & $1.07 \times 10^{-4}$ \\
        \bottomrule
    \end{tabular}
\end{table}

The correlation between NTK spectral properties and final training errors is analyzed. Four spectral metrics of the residual NTK matrix $\bm{K}_r$ are examined: convergence rate $c$, condition number $\kappa$, effective rank $r_{\text{eff}}$, and Frobenius norm $\|\bm{K}_r\|_F$.

\begin{figure}[!htb]
    \centering
    \includegraphics[width=\textwidth]{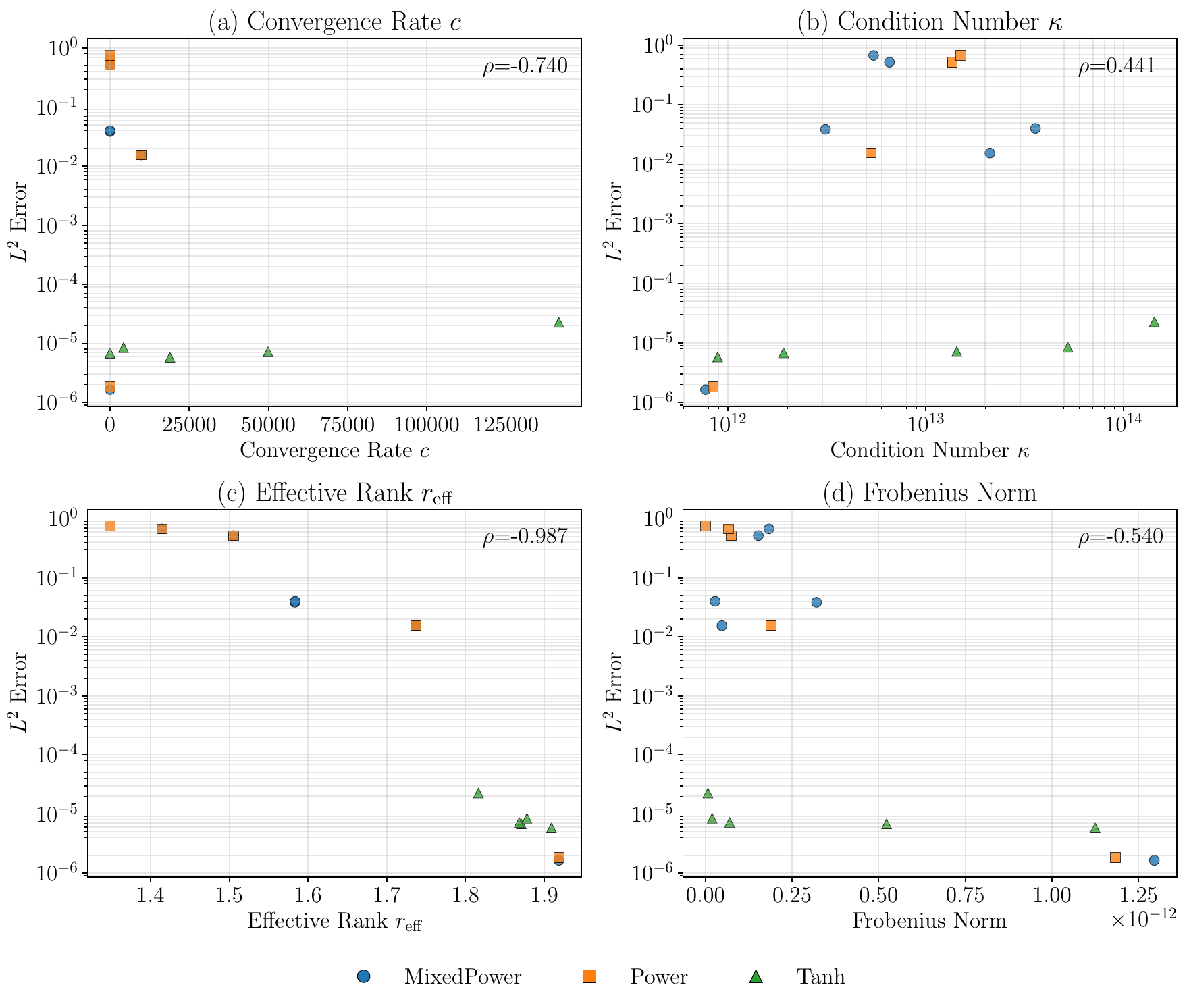}
    \caption{Correlation between $\bm{K}_r$ spectral properties and final $L^2$ errors. (a) Convergence rate $c$. (b) Condition number $\kappa$. (c) Effective rank $r_{\text{eff}}$. (d) Frobenius norm $\|\cdot\|_F$.}
    \label{fig:kr_spectral_correlation_2d}
\end{figure}

Fig.~\ref{fig:kr_spectral_correlation_2d} presents the correlation analysis.
Strong correlations are observed. Effective rank $r_{\text{eff}}$ demonstrates exceptional predictive power ($\rho = -0.987$), stronger than in the 1D case.
This suggests that dimensional coupling amplifies the importance of a well-conditioned $\bm{K}_r$.
Convergence rate $c$ shows strong negative correlation, while Frobenius norm $\|\bm{K}_r\|_F$ exhibits moderate correlation.
Condition number $\kappa$ displays only weak positive correlation.
Fig.~\ref{fig:kr_spectral_correlation_2d}(c) reveals clustering patterns: Tanh occupies the high-$r_{\text{eff}}$, low-error region, while poor Power and MixedPower parameters cluster in the low-$r_{\text{eff}}$, high-error region.

Spectral feature correlations are examined within each boundary function family.

\begin{figure}[!htb]
    \centering
    \includegraphics[width=0.95\textwidth]{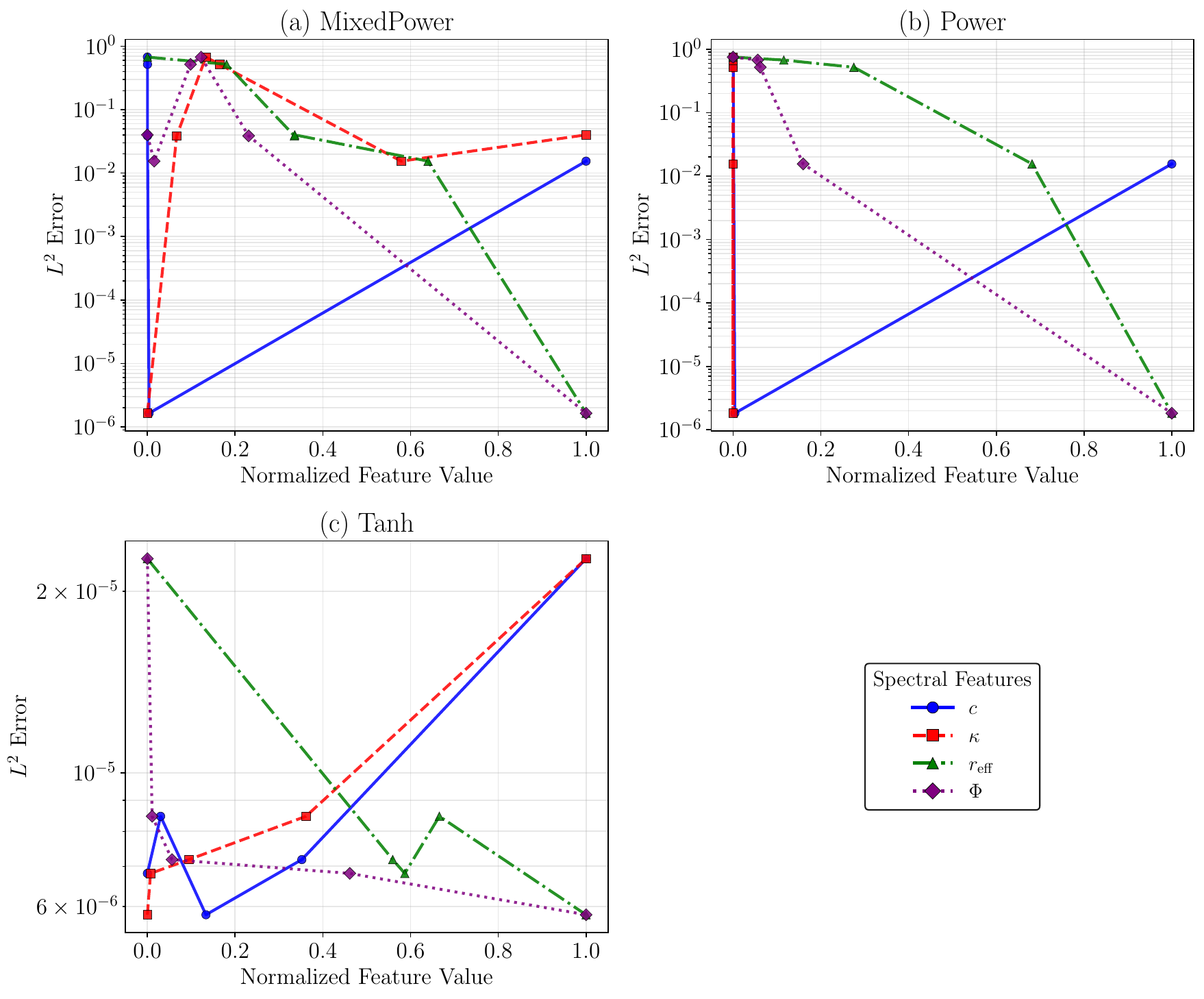}
    \caption{Spectral feature sensitivity analysis within 2D boundary function families. Each panel shows error vs normalized spectral features for one family: (a) MixedPower, (b) Power, (c) Tanh.}
    \label{fig:spectral_feature_sensitivity_2d}
\end{figure}

Fig.~\ref{fig:spectral_feature_sensitivity_2d} presents the spectral feature sensitivity analysis.
The MixedPower family (Fig.~\ref{fig:spectral_feature_sensitivity_2d}(a)) exhibits complex relationships, but effective rank $r_{\text{eff}}$ shows a consistent downward trend.
The Power family (Fig.~\ref{fig:spectral_feature_sensitivity_2d}(b)) demonstrates dramatic sensitivity, with strong correlations for all metrics.
The Tanh family (Fig.~\ref{fig:spectral_feature_sensitivity_2d}(c)) shows flat relationships, reflecting its robustness.
It is observed that effective rank $r_{\text{eff}}$ becomes increasingly critical in higher dimensions.

The MixedPower family allows investigation of dimensional anisotropy.
Symmetric exponents $(\alpha, \beta) = (1.0, 1.0)$ achieve the best performance, while asymmetric high-order combinations produce the worst results.
Moderate asymmetry yields intermediate performance.
For the isotropic square domain, symmetric formulations consistently outperform asymmetric ones.

Several key findings are established.
First, spectral-performance relationships generalize to 2D, with effective rank $r_{\text{eff}}$ being a strong predictor.
Second, optimal boundary function choice is dimension-dependent; Tanh provides robustness, while simple first-order polynomials are effective for rectangular domains.
Third, dimensional anisotropy should be reserved for specific problems.
Fourth, proper boundary function selection is critical in higher dimensions.
Tanh is recommended for robust performance, and effective rank $r_{\text{eff}}$ should be verified.

\subsubsection{Three Dimensional Case}
\label{sec:simple_B_case_4}

The analysis is extended to the 3D linear diffusion equation on a unit cube:
\begin{equation}
   -D\nabla^2 u + a u = f(x,y,z), \quad (x,y,z) \in [0, 1]^3,
\end{equation}
with homogeneous Dirichlet BCs. The analytical solution is $u(x,y,z) = \sin(\pi x)\sin(\pi y) \\\sin(\pi z)$.
The trial function is $\tilde{u}(x, y, z, \theta) = B(x,y,z)  N(x, y, z, \vec \theta)$.
Two parametric families of boundary functions $B$ are investigated:
\begin{enumerate}
    \item MixedPower (Symmetric): $B(x,y,z) = [x(1-x)]^\alpha [y(1-y)]^\alpha [z(1-z)]^\alpha$, with $\alpha \in \{0.5, 1.0, 1.5, 2.0\}$
    \item MixedPower (Asymmetric): $B(x,y,z) = [x(1-x)]^\alpha [y(1-y)]^\beta [z(1-z)]^\gamma$, with various $(\alpha, \beta, \gamma)$ combinations.
    \item Hyperbolic tangent: $B(x,y,z) = \prod_{i \in \{x,y,z\}} \tanh(\alpha i)\tanh(\alpha(1-i))$, with $\alpha \in \{1, 3, 5, 7, 10\}$
\end{enumerate}
The MixedPower asymmetric family introduces three-dimensional anisotropy. Thirteen experiments are conducted.

\begin{figure}[!htb]
    \centering
    \includegraphics[width=0.95\textwidth]{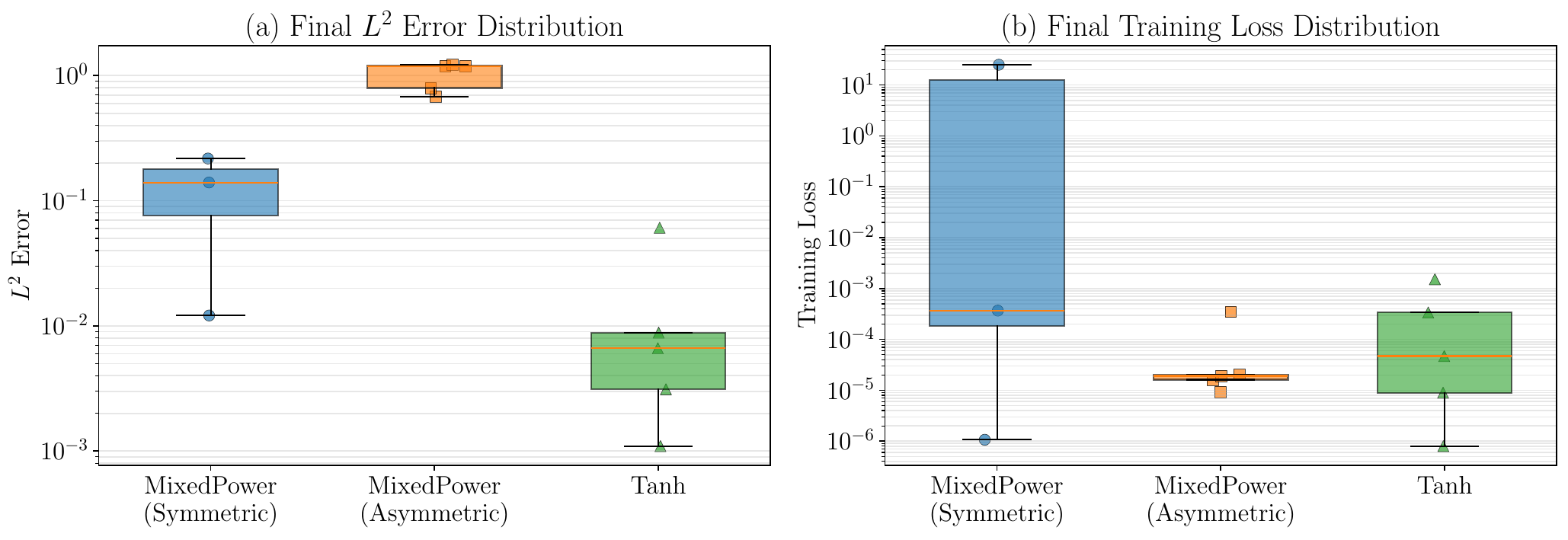}
    \caption{Training performance across different 3D boundary function $B$ families. (a) Final $L^2$ errors. (b) Final training loss.}
    \label{fig:boundary_family_comparison_3d}
\end{figure}

Fig.~\ref{fig:boundary_family_comparison_3d} shows the final $L^2$ errors and training losses.
Dramatic performance differences are observed. The Tanh family achieves the best mean performance ($L^2 = 1.61 \times 10^{-2}$), while the MixedPower (Asymmetric) family performs worst ($L^2 = 1.02$).
The Tanh family demonstrates exceptional robustness.
The MixedPower symmetric family shows moderate robustness, with optimal performance at $\alpha = 1.0$.
In contrast, the MixedPower asymmetric family exhibits poor performance across all configurations, indicating that asymmetric formulations are ineffective in this context.
This suggests that introducing directional anisotropy disrupts spectral properties more severely in 3D than in 2D.

\begin{table}[!htb]
    \centering
    \caption{Training performance statistics across 3D boundary function families}
    \label{tab:boundary_family_summary_3d}
    \begin{tabular}{lcccc}
        \toprule
        Family & $L^2$ Error Mean & $L^2$ Error Std & Loss Mean & Loss Std \\
        \midrule
        \multicolumn{5}{c}{\textit{Parametric Families}} \\
        \midrule
        MixedPower (Sym.) & $1.23 \times 10^{-1}$ & $1.04 \times 10^{-1}$ & $8.36$ & $1.45 \times 10^{1}$ \\
        MixedPower (Asym.) & $1.02$ & $2.59 \times 10^{-1}$ & $8.29 \times 10^{-5}$ & $1.49 \times 10^{-4}$ \\
        \textbf{Tanh} & $\mathbf{1.61 \times 10^{-2}}$ & $2.52 \times 10^{-2}$ & $3.84 \times 10^{-4}$ & $6.53 \times 10^{-4}$ \\
        \bottomrule
    \end{tabular}
\end{table}

Table~\ref{tab:boundary_family_summary_3d} summarizes the training performance statistics.
It is observed that the MixedPower asymmetric family achieves low training loss but poor generalization, indicating that training loss alone is an unreliable indicator.
Well-performing families maintain a consistent correlation between training loss and error.

The correlation between NTK spectral properties and final training errors is analyzed. Four spectral metrics of the residual NTK matrix $\bm{K}_r$ are examined: convergence rate $c$, condition number $\kappa$, effective rank $r_{\text{eff}}$, and Frobenius norm $\|\bm{K}_r\|_F$.

\begin{figure}[!htb]
    \centering
    \includegraphics[width=\textwidth]{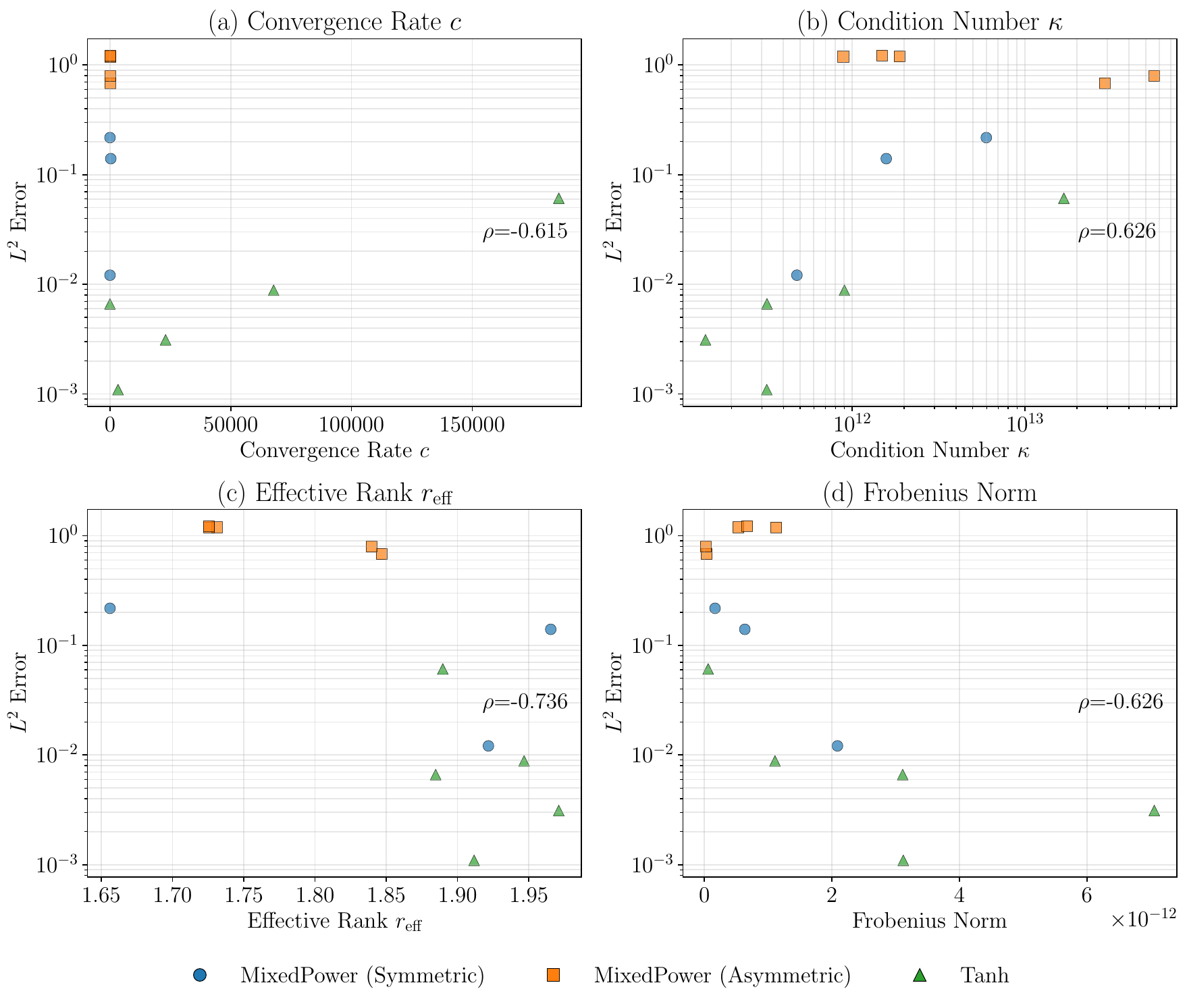}
    \caption{Correlation between $\bm{K}_r$ spectral properties and final $L^2$ errors. (a) Convergence rate $c$. (b) Condition number $\kappa$. (c) Effective rank $r_{\text{eff}}$. (d) Frobenius norm $\|\cdot\|_F$.}
    \label{fig:kr_spectral_correlation_3d}
\end{figure}

Fig.~\ref{fig:kr_spectral_correlation_3d} presents the correlation analysis.
Consistent with 2D patterns, effective rank $r_{\text{eff}}$ is the strongest predictor ($\rho = -0.736$).
Condition number $\kappa$ shows strong positive correlation, while Frobenius norm $\|\bm{K}_r\|_F$ and convergence rate $c$ exhibit moderate negative correlations.
This confirms that effective rank $r_{\text{eff}}$ remains the primary diagnostic across dimensions.
Fig.~\ref{fig:kr_spectral_correlation_3d}(c) reveals clustering patterns: Tanh occupies the high-$r_{\text{eff}}$, low-error region, while MixedPower asymmetric configurations cluster in the low-$r_{\text{eff}}$, high-error region.

Several critical findings are established.
First, effective rank $r_{\text{eff}}$ maintains its dominance as the primary performance predictor in 3D.
Second, a dimensional asymmetry effect is observed: 3D universally penalizes asymmetric formulations.
Third, training loss decouples from solution accuracy in poorly conditioned regimes.
Fourth, optimal boundary function choices exhibit dimensional consistency, with Tanh and first-order symmetric polynomials performing well.
For 3D problems, Tanh is recommended for optimal accuracy and robustness. Initialization-time spectral analysis should be performed to verify $r_{\text{eff}}$. Asymmetric formulations should be avoided unless necessary. Training loss should not be the sole validation metric.

\clearpage

\clearpage
\section{Conclusion}
\label{sec:conclusion}

This work systematically establishes a NTK theoretical framework for HC-PINNs and validates its predictions through comprehensive numerical experiments.
The resulting framework provides both theoretical insights into HC-PINN training mechanisms and practical guidance for boundary function design, though the investigations also reveal fundamental limitations when applying idealized NTK theory to modern optimization algorithms.

The core theoretical contribution lies in the systematic derivation and analysis of three interconnected NTK matrices, $\bm{K}_n$, $\bm{K}_t$, and $\bm{K}_r$, within the HC-PINN framework.
It is revealed that the boundary function $B$ fundamentally influences training dynamics through its spatial modulation of neural network gradients.
For $\bm{K}_t$, the relationship $\bm{K}_t = \bm{B} \bm{K}_n \bm{B}$ is derived. Spectral analysis indicates that the effective rank $r_{\text{eff}}$ is a reliable predictor, showing strong correlations with boundary function curvature and Gini coefficient.
For $\bm{K}_r$, the general form for linear differential operators is derived. Numerical investigations on second-order linear differential equations demonstrate that the second derivative $B''$ exhibits strong negative correlations with effective rank.
It is observed that residual NTK magnitudes span eight orders of magnitude across boundary function choices, confirming extreme sensitivity to boundary function selection.
The training dynamics framework establishes that residual evolution follows $\frac{d\vec{R}}{dt} = -\frac{2\eta}{N_r} \bm{K}_r \vec{R}$ in the NTK regime.
However, systematic comparison reveals that practical Adam + L-BFGS optimization significantly outperforms theoretically-prescribed SGD, explicitly abandoning the lazy training regime through active feature learning.
Despite these departures from idealized assumptions, the initial spectral properties of $\bm{K}_r$ remain powerful predictors of final training performance across 1D, 2D, and 3D benchmark cases.
Effective rank $r_{\text{eff}}$ serves as the primary diagnostic, showing increasingly strong correlations in higher dimensions. 
The investigations yield practical guidelines for HC-PINN implementation: smooth boundary function families (hyperbolic tangent with $\alpha \in [3,5]$, rational functions) should be prioritized for robust performance across diverse problems, with first-order symmetric polynomials offering computational efficiency for rectangular domains; initialization-time computation of $\bm{K}_r$ spectral properties, particularly effective rank $r_{\text{eff}} \geq 1.9$ for 2D/3D problems, provides reliable training difficulty prediction before expensive optimization; asymmetric boundary formulations should be avoided unless problem geometry explicitly requires directional anisotropy, as they universally degrade performance in 3D through eigenvalue concentration. 
This work demonstrates that NTK theory serves as a valuable diagnostic tool that successfully identifies problem characteristics correlating with training difficulty, though the framework's limitations---including violations of lazy training assumptions by modern optimizers and positive correlations between larger convergence rates and higher errors that contradict classical NTK predictions---define its proper scope as providing comparative analysis tools rather than quantitative predictions. 

Future work should extend the framework to complex geometries, time-dependent and nonlinear problems, and develop theoretical refinements that bridge the gap between linearized NTK analysis and practical feature learning regimes.

\clearpage

\appendix
\section{Invariance of initial \texorpdfstring{$\bm{K_n}$}{Kn}}
\label{sec_invariance_K_n}

This section analyzes the invariance of the initial NTK matrix $\bm{K_n}$ under various conditions.

To quantify the similarity between NTK matrices $\bm{K}$, this study employs both global metrics of the NTK matrix and element-wise comparisons.

The global metrics include Centered Kernel Alignment (CKA) \cite{CortesAlgorithmsLearningKernels2012} and common spectral analysis indicators such as condition number, effective rank, trace, and Frobenius norm.

Element-wise comparisons are conducted using the mean, standard deviation, and coefficient of variation (CV) of the NTK matrix elements.

CKA is a standard metric for measuring the similarity between two kernel matrices and is defined as:
$$
    \text{CKA}(\bm{K}_1, \bm{K}_2) = \frac{\text{tr}(\bm{K}_1^c \bm{K}_2^c)}{\sqrt{\text{tr}(\bm{K}_1^c \bm{K}_1^c) \cdot \text{tr}(\bm{K}_2^c \bm{K}_2^c)}}
$$
where $\text{tr}(\cdot)$ represents the trace operation, and $\bm{K}^c$ denotes the centered version of the kernel matrix $\bm{K}$, as defined by:
$$
    \bm{K}^c = H \bm{K} H, \quad H = I - \frac{1}{n}\mathbf{1}\mathbf{1}^T
$$
where $I$ is the identity matrix, $\mathbf{1}$ is a vector of ones, and $n$ is the number of data points.

The CKA value ranges from 0 to 1, where a value of 1 indicates perfect similarity between the two kernel matrices after centering, while a value of 0 indicates no similarity.

This metric is insensitive to scaling of the kernel matrices and focuses solely on structural similarity.

The spectral analysis indicators used in this study include the condition number ($\kappa$), effective rank ($r_{\text{eff}}$), and trace (tr) as defined in Section~\ref{sec:K_t_spectral}, along with the metric: Frobenius norm ($\|\bm{K}\|_F$), which are defined as follows:
$$
    \|\bm{K}\|_F = \sqrt{\sum_{i,j} \bm{K}_{ij}^2} = \sqrt{\text{tr}(\bm{K}^T \bm{K})}
$$
where $\bm{K}_{ij}$ are the elements of the NTK matrix. The Frobenius norm measures the overall size of the matrix in terms of its elements.

For element-wise comparisons of the NTK matrices, given $M$ NTK matrices $\{\bm{K}^{(1)}, \bm{K}^{(2)}, \ldots, \bm{K}^{(M)}\}$ obtained from different experiments, the following statistics are computed for each matrix element position $(i,j)$:
\begin{enumerate}
    \item \textbf{Element-wise Mean.} The mean across $M$ matrices at position $(i,j)$ is defined as:
          $$
              \mu_{ij} = \frac{1}{M} \sum_{m=1}^{M} \bm{K}^{(m)}_{ij}
          $$
    \item \textbf{Element-wise Standard Deviation.} The standard deviation across $M$ matrices at position $(i,j)$ is defined as:
          $$
              \sigma_{ij} = \sqrt{\frac{1}{M} \sum_{m=1}^{M} (\bm{K}^{(m)}_{ij} - \mu_{ij})^2}
          $$
    \item \textbf{Element-wise CV.} The coefficient of variation across $M$ matrices at position $(i,j)$ is defined as:
          $$
              \text{CV}_{ij} = \frac{\sigma_{ij}}{\mu_{ij}}
          $$
\end{enumerate}
These element-wise statistics quantify the variability of each NTK matrix element across different situations.
Lower CV values indicate smaller variations of the element at that position across different experiments, demonstrating better stability of that element.

\subsection{Invariance under Different Random Seeds}
\label{sec:invariance_seed}

This subsection investigates the invariance of the initial NTK matrix $\bm{K_n}$ under different random seed initializations for wide neural networks.

The experiments are conducted using a fully connected neural network with the following configuration:
\begin{itemize}
    \item Network architecture: \num{2} hidden layers with \num{500} neurons per layer (wide network regime)
    \item Activation function: Hyperbolic tangent ($\tanh$)
    \item Input dimension: \num{1}
    \item Output dimension: \num{1}
    \item Collocation points: \num{100} uniformly distributed points in $[0, 1]$
    \item Random seeds: \num{1000} different initializations
          
\end{itemize}

The neural network weights are initialized using the Kaiming uniform initialization method \cite{HeDeepResidualLearning2016}, which is the default initialization scheme in PyTorch's \texttt{nn.Linear} module.

PyTorch is a widely used deep learning framework known for its dynamic computation graph and ease of use \cite{ImambiPyTorch2021}.
Specifically, the weights are initialized from a uniform distribution $\mathcal{U}(-\sqrt{k_{input}}, \sqrt{k_{input}})$, where $k_{input}$ is the inverse of the number of input units, and biases are initialized from $\mathcal{U}(-\sqrt{k_{output}}, \sqrt{k_{output}})$ where $k_{output}$ is the inverse of the number of output units.
This initialization ensures that the variance of activations is preserved across layers in the forward pass.

\begin{figure}[p]  
    \centering
    \includegraphics[width=0.95\textwidth]{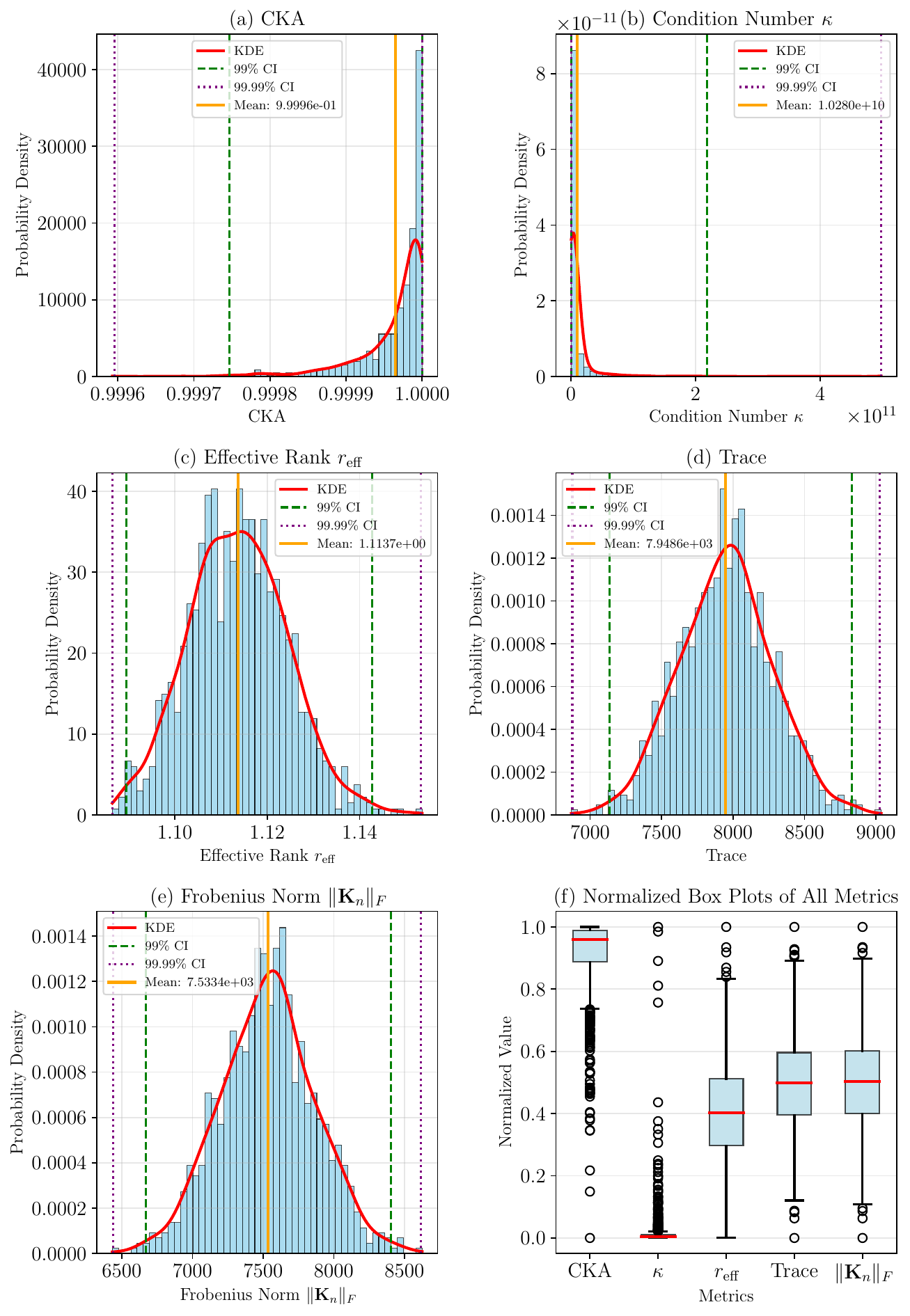}
    \caption{Global metrics across different random seed initializations.}
    \label{fig:seed_global_metrics}
\end{figure}

Fig.~\ref{fig:seed_global_metrics} presents the distribution of global metrics across \num{1000} different random seed initializations.
The CKA values (Fig.~\ref{fig:seed_global_metrics}(a)) show exceptional consistency with a mean of \num{0.999965} and standard deviation of only \num{4.9e-5}, with all values falling within the $99\%$ confidence interval of $[\num{0.999747}, \num{1.000000}]$, indicating nearly perfect structural similarity across all initializations.
In contrast, the condition numbers (Fig.~\ref{fig:seed_global_metrics}(b)) exhibit significant variation, with a mean of \num{1.03e10} and standard deviation of \num{3.62e10}, and the $99\%$ confidence interval spanning $[\num{6.89e8}, \num{2.18e11}]$, reflecting the inherent numerical sensitivity of this metric to small perturbations in eigenvalues.
The effective rank (Fig.~\ref{fig:seed_global_metrics}(c)) remains highly stable with a mean of \num{1.114} and standard deviation of \num{0.011}, with the narrow $99\%$ confidence interval of $[\num{1.089}, \num{1.143}]$ suggesting consistent eigenvalue distribution patterns across different initializations.
The trace values (Fig.~\ref{fig:seed_global_metrics}(d)) concentrate around a mean of \num{7.95e3} with a standard deviation of \num{3.26e2}, showing relative stability with a coefficient of variation of approximately $4.1\%$, while the Frobenius norm (Fig.~\ref{fig:seed_global_metrics}(e)) demonstrates similar consistency with a mean of \num{7.53e3} and standard deviation of \num{3.33e2}, yielding a coefficient of variation around $4.4\%$.

For better visual comparison, we normalize each metric by its mean value, allowing for direct assessment of relative variability across different metrics.
The normalized box plots (Fig.~\ref{fig:seed_global_metrics}(f)) reveal that CKA and effective rank exhibit the tightest distributions, while the condition number shows the greatest variability, confirming that structural similarity metrics are more robust indicators of NTK invariance than magnitude-based metrics.

\begin{figure}[!htb]
    \centering
    \includegraphics[width=\textwidth]{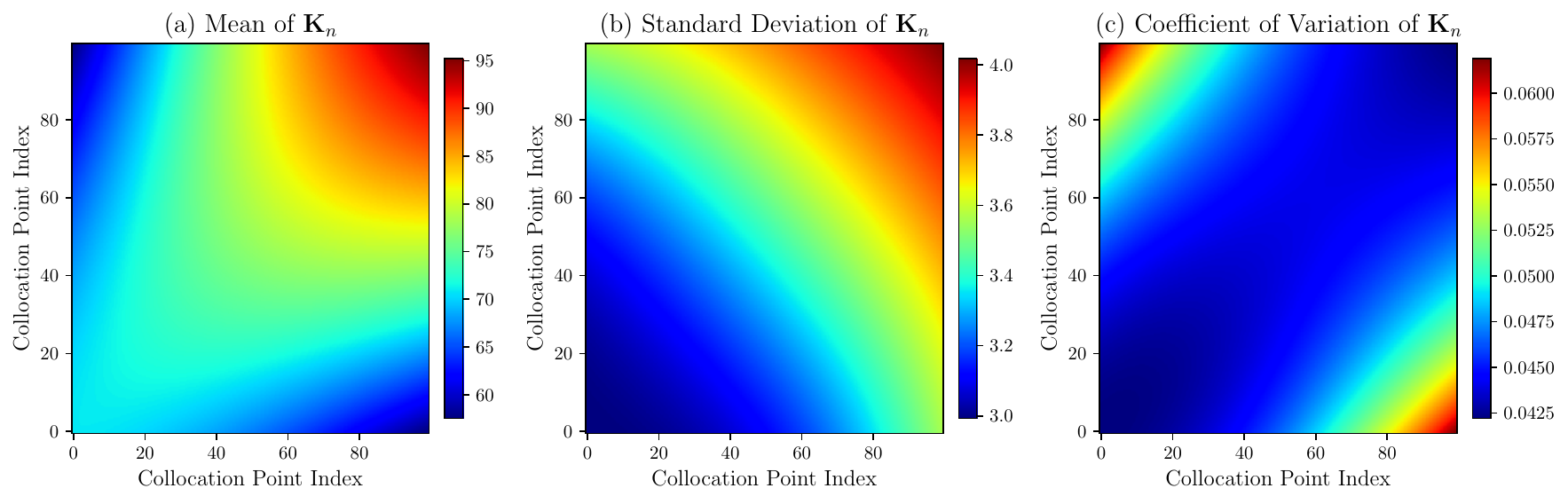}
    \caption{Element-wise statistics of NTK matrices across different random seed initializations.}
    \label{fig:seed_element_wise}
\end{figure}

Fig.~\ref{fig:seed_element_wise} illustrates the element-wise statistics of the NTK matrices across all random seeds.
The mean matrix (Fig.~\ref{fig:seed_element_wise}(a)) shows values ranging from \num{57.5} to \num{95.2}, with an overall mean of \num{75.0}.
The standard deviation matrix (Fig.~\ref{fig:seed_element_wise}(b)) reveals that absolute variations range from \num{2.99} to \num{4.02}, with a mean of \num{3.44}.
Most importantly, the coefficient of variation matrix (Fig.~\ref{fig:seed_element_wise}(c)) demonstrates that relative variations are consistently small, ranging from \num{4.22e-2} to \num{6.19e-2} with a mean of \num{4.60e-2} (approximately $4.6\%$), indicating that each element of the NTK matrix maintains stable relative values across different random initializations.

In conclusion, the numerical experiments demonstrate that the initial NTK matrix $\bm{K_n}$ exhibits strong invariance under different random seed initializations in the wide network regime (width = \num{500}).
The CKA metric consistently exceeds \num{0.9997}, indicating that the NTK's geometric structure remains essentially unchanged regardless of initialization.
The element-wise coefficient of variation analysis confirms that variations are uniformly distributed across the matrix with relative fluctuations limited to approximately $4.6\%$.

These results validate the predictions of NTK theory \cite{ntk} and provide empirical support for subsequent experiments that do not involve initialization invariance: in experiments not focused on initialization invariance, a single random seed can be fixed without the need to recompute the NTK across different initializations; however, for experiments investigating initialization invariance, multiple random initializations are still necessary to assess stability.

\subsection{Variation of Invariance with Network Width}
\label{sec:invariance_width}

This subsection investigates the invariance of the initial NTK matrix $\bm{K_n}$ under different network widths.

The experimental configuration follows \ref{sec:invariance_seed}, with the primary modification being the network width.
The tested widths span a wide range: $\{5, 10, 20, 50, 100, \\ 200, 500, 1000, 2000, 5000\}$, covering from narrow to extremely wide networks. 
For each width, \num{50} different random seed initializations are used to evaluate the stability of the NTK matrix under that specific width.

\begin{figure}[!htb]
    \centering
    \includegraphics[width=\textwidth]{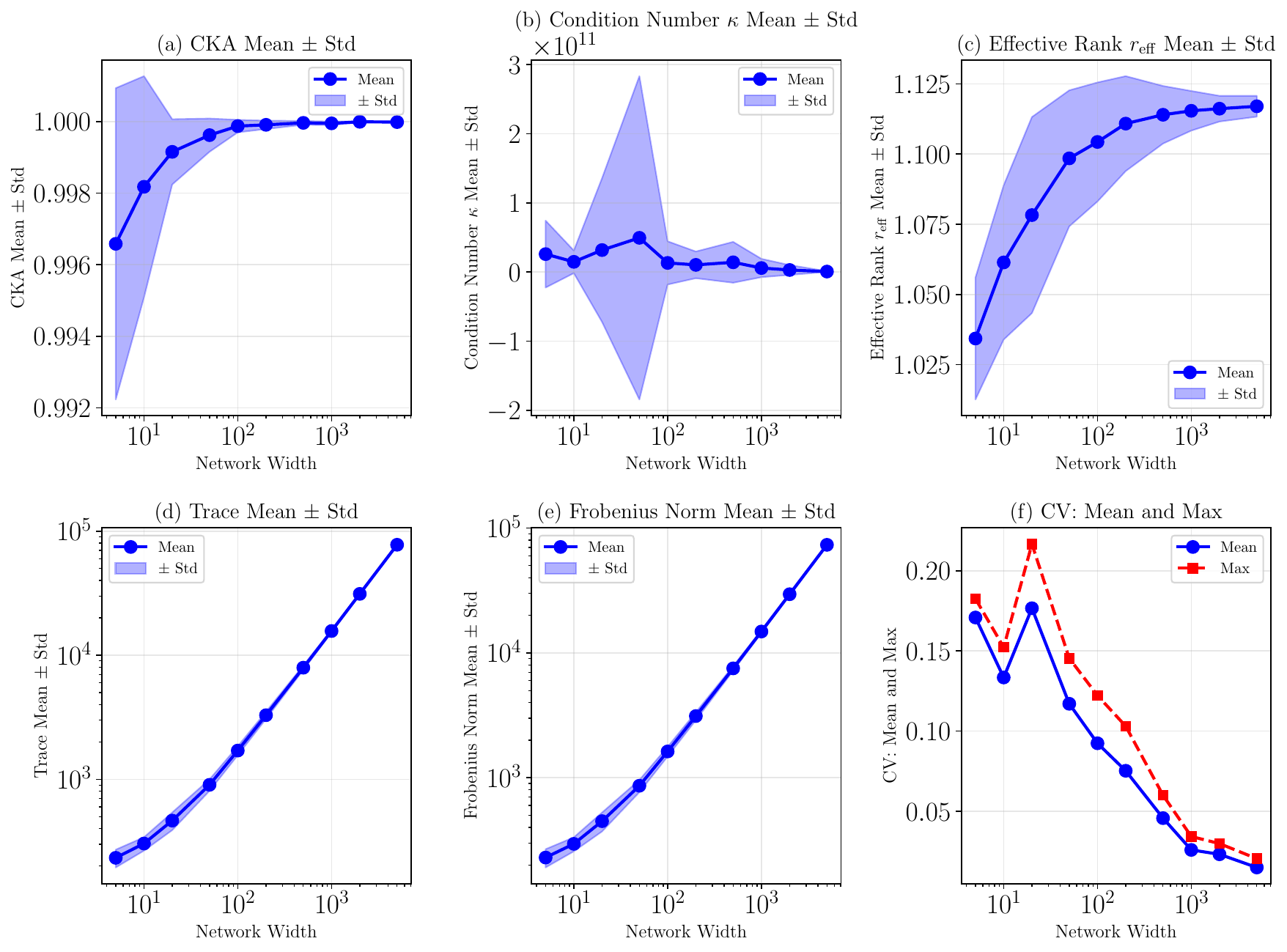}
    \caption{Stability metrics of NTK matrices across different network widths. Each point represents the mean value across \num{50} random seeds, with shaded regions indicating $\pm$ one standard deviation.}
    \label{fig:width_stability_trends}
\end{figure}

Fig.~\ref{fig:width_stability_trends} presents the evolution of stability metrics as network width increases.
The CKA metric (Fig.~\ref{fig:width_stability_trends}(a)) demonstrates a clear convergence trend, rising from \num{0.9966} at width \num{5} to \num{0.9999} at width \num{5000}, with the standard deviation decreasing from \num{4.35e-3} to \num{1.3e-5}, indicating that wider networks produce more consistent NTK structures across different initializations.
The condition number (Fig.~\ref{fig:width_stability_trends}(b)) exhibits substantial variability across widths, ranging from \num{2.59e10} (width \num{5}) to \num{6.20e8} (width \num{5000}), though this metric is known to be sensitive to numerical precision and may not directly reflect functional stability.
The effective rank (Fig.~\ref{fig:width_stability_trends}(c)) shows a gradual increase from \num{1.03} to \num{1.12} as width grows, suggesting that wider networks develop more balanced eigenvalue distributions.
Both the trace (Fig.~\ref{fig:width_stability_trends}(d)) and Frobenius norm (Fig.~\ref{fig:width_stability_trends}(e)) display logarithmic growth with network width, reflecting the scaling behavior predicted by NTK theory.
Most critically, the CV (Fig.~\ref{fig:width_stability_trends}(f)) decreases monotonically with width, from \num{0.171} (mean) and \num{0.182} (maximum) at width \num{5} to \num{0.015} (mean) and \num{0.021} (maximum) at width \num{5000}, demonstrating that element-wise stability improves dramatically as networks become wider.
These results confirm that network widths of \num{500} and above exhibit excellent invariance properties, while narrower networks (width $< 50$) show greater sensitivity to initialization.

\begin{figure}[!htb]
    \centering
    \includegraphics[width=\textwidth]{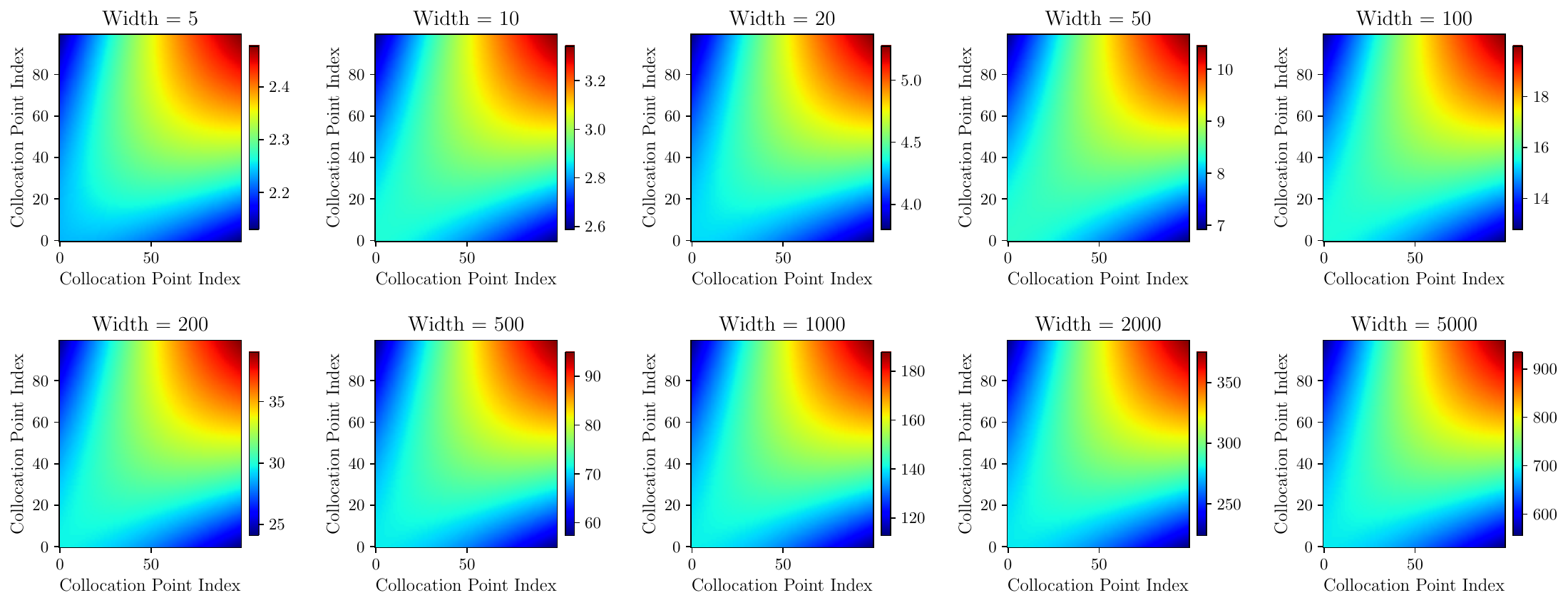}
    \caption{Heatmap comparison of mean NTK matrices across all tested widths. Each subplot shows the element-wise mean of the NTK matrix computed from \num{50} random initializations at the corresponding width.}
    \label{fig:width_heatmap_comparison}
\end{figure}

Fig.~\ref{fig:width_heatmap_comparison} presents a comprehensive visual comparison of the mean NTK matrix structure across all tested widths.
Despite spanning four orders of magnitude in width (from \num{5} to \num{5000}), all NTK matrices exhibit remarkably similar structural patterns.

While the overall structure remains consistent, the magnitude of matrix elements increases systematically with width, as evidenced by the evolving color scale in each subplot.
This observation indicates that the kernel magnitude increases with network width, while the normalized kernel structure remains largely unchanged across widths.
The structural similarity across widths provides strong empirical evidence that the NTK captures intrinsic geometric properties of the neural network function space that are largely independent of the network's capacity.

\begin{figure}[!htb]
    \centering
    \includegraphics[width=0.7\textwidth]{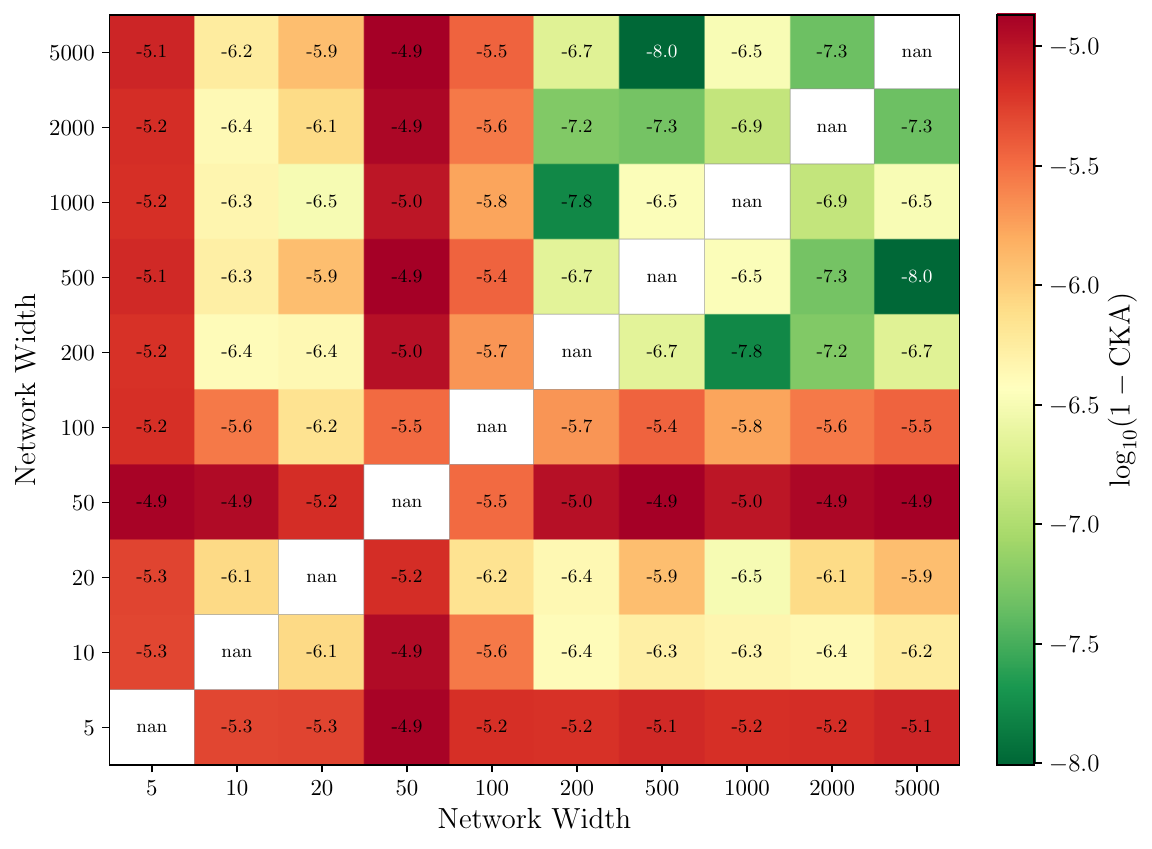}
    \caption{CKA similarity matrix between different network widths, displayed as $\log_{10}(1 - \text{CKA})$. Lower (more negative) values indicate higher similarity. Each cell compares the mean NTK matrices computed from \num{50} random seeds at the corresponding widths.}
    \label{fig:width_cka_similarity}
\end{figure}

To quantitatively assess the structural similarity between NTK matrices at different widths, Fig.~\ref{fig:width_cka_similarity} presents the CKA similarity matrix using a logarithmic scale $\log_{10}(1 - \text{CKA})$ to better visualize the small differences.
The analysis reveals several key findings:
\begin{enumerate}
    \item \textbf{Overall high similarity.} All pairwise CKA values exceed \num{0.999}, with most exceeding \num{0.9999}, indicating extremely high structural similarity across all widths.
    \item \textbf{Increasing similarity with width.} Adjacent width pairs show $\log_{10}(1 - \text{CKA})$ values ranging from \num{-5.3} (widths \num{5} vs \num{10}) to \num{-7.3} (widths \num{2000} vs \num{5000}), demonstrating increasing similarity as networks become wider.
    \item \textbf{Robustness across extreme widths.} Even the extreme comparison between the narrowest (width \num{5}) and widest (width \num{5000}) networks yields a CKA of \num{0.9999925}, corresponding to $\log_{10}(1 - \text{CKA}) = \num{-5.12}$, confirming that the fundamental NTK structure is preserved across the entire width spectrum.
    \item \textbf{Convergence to a common structure.} The progressively darker (more negative) values in the lower-right corner of the matrix indicate that larger widths converge toward a common limiting structure, consistent with the infinite-width NTK theory.
\end{enumerate}

In summary, the analysis of NTK invariance across network widths reveals two fundamental patterns:
\begin{enumerate}
    \item \textbf{Width-dependent stability improvement.} As network width increases, the NTK matrix exhibits progressively better invariance under random initialization, evidenced by CKA values approaching unity, decreasing coefficient of variation, and shrinking standard deviations across stability metrics.
    \item \textbf{Universal structural similarity with magnitude scaling.} Despite substantial differences in network capacity, NTK matrices across all tested widths maintain highly similar geometric structures (CKA $> 0.999$ for all pairs).
\end{enumerate}
This structural invariance coexists with systematic magnitude scaling: matrix element values increase with width while preserving their relative relationships, a phenomenon that can be normalized out when focusing on kernel geometry.
These findings establish practical guidelines for network width selection in applications requiring stable kernel behavior.

\subsection{Variation of Invariance with Network Depth}
\label{sec:invariance_depth}

This subsection investigates the invariance of the initial NTK matrix $\bm{K_n}$ under different network depths.

The experimental configuration follows \ref{sec:invariance_seed}, with the primary modification being the network depth (number of hidden layers).
The tested depths range from shallow to deep networks: $\{1, 2, 3, 4, 5, 6, 7, 8\}$, while the network width is fixed at \num{500} to ensure adequate expressiveness.
Considering the computational cost of deeper networks, \num{20} different random seed initializations are used for each depth to evaluate stability.

\begin{figure}[!htb]
    \centering
    \includegraphics[width=\textwidth]{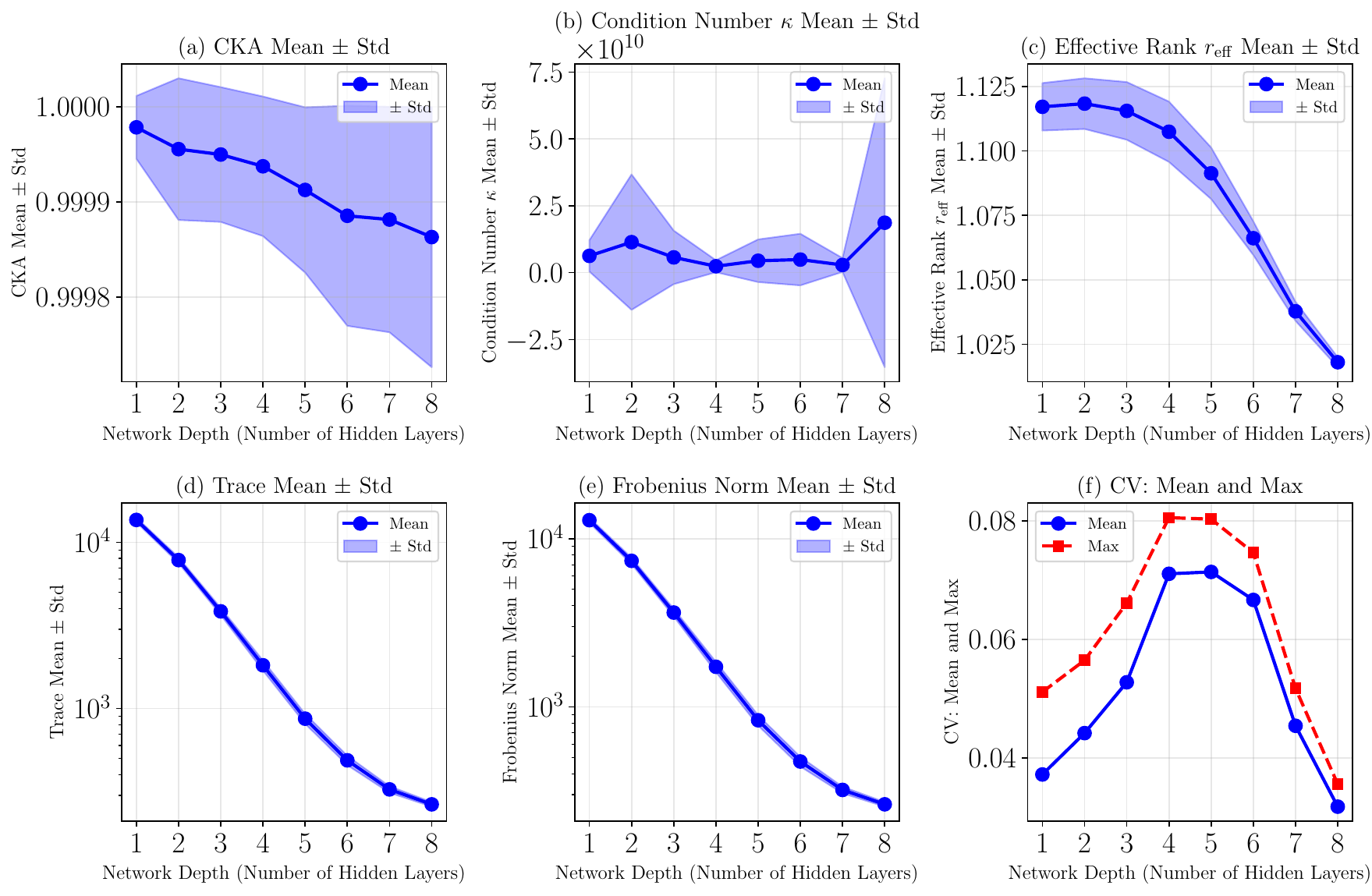}
    \caption{Stability metrics of NTK matrices across different network depths. Each point represents the mean value across \num{20} random seeds, with shaded regions indicating $\pm$ one standard deviation.}
    \label{fig:depth_stability_trends}
\end{figure}

Fig.~\ref{fig:depth_stability_trends} presents the evolution of stability metrics as network depth increases.
The CKA metric (Fig.~\ref{fig:depth_stability_trends}(a)) exhibits a gentle decline from \num{0.999978} at depth \num{1} to \num{0.999863} at depth \num{8}, with standard deviations gradually increasing from \num{3.3e-5} to \num{1.37e-4}.
Despite this slight decrease, all CKA values remain above \num{0.9998}, indicating that NTK structures maintain high consistency across different initializations even in deeper networks.
The condition number (Fig.~\ref{fig:depth_stability_trends}(b)) shows substantial variation across depths, ranging from \num{6.27e9} to \num{1.87e10}, with large standard deviations reflecting the inherent numerical sensitivity of this metric in deeper architectures.
The effective rank (Fig.~\ref{fig:depth_stability_trends}(c)) displays a notable monotonic decrease from \num{1.117} at depth \num{1} to \num{1.018} at depth \num{8}, suggesting that deeper networks produce more concentrated eigenvalue distributions with increasingly dominant leading eigenvalues.
Both the trace (Fig.~\ref{fig:depth_stability_trends}(d)) and Frobenius norm (Fig.~\ref{fig:depth_stability_trends}(e)) exhibit dramatic monotonic decreases with network depth: the trace drops from \num{1.36e4} at depth \num{1} to \num{265} at depth \num{8}, while the Frobenius norm decreases from \num{1.29e4} to \num{263}.
This substantial reduction in both metrics indicates that the overall magnitude of the NTK matrix decreases significantly as networks become deeper, in stark contrast to the width analysis where these metrics increased with network size.
This phenomenon suggests that deeper networks effectively dampen the kernel magnitude through repeated nonlinear transformations.
Interestingly, the CV (Fig.~\ref{fig:depth_stability_trends}(f)) exhibits a non-monotonic pattern: it increases from \num{0.037} (mean) at depth \num{1} to peak at \num{0.071} (mean) around depths \num{4}--\num{5}, then decreases to \num{0.032} (mean) at depth \num{8}.
This non-monotonic behavior suggests that moderate depths (around \num{4}--\num{6} layers) introduce the greatest variability,
while very deep networks (\num{7}--\num{8} layers) paradoxically exhibit improved element-wise stability, possibly through implicit regularization effects.
Although the CKA values decline slightly with increasing depth (from \num{0.999978} at depth \num{1} to \num{0.999863} at depth \num{8}), all values remain exceptionally high ($> 0.9998$), confirming strong structural consistency across random initializations even in deep networks.
While the single hidden layer network achieves the highest initialization stability, it suffers from limited expressiveness and representational capacity for complex functions—a key reason why subsequent numerical experiments in this work consistently employ networks with two hidden layers.
Conversely, very deep networks, despite maintaining structural NTK invariance, face additional practical challenges including gradient vanishing/exploding, increased training difficulty, and degradation in optimization landscape quality, compounded by the observed \num{50}$\times$ magnitude dampening in trace and Frobenius norm.

\begin{figure}[!htb]
    \centering
    \includegraphics[width=\textwidth]{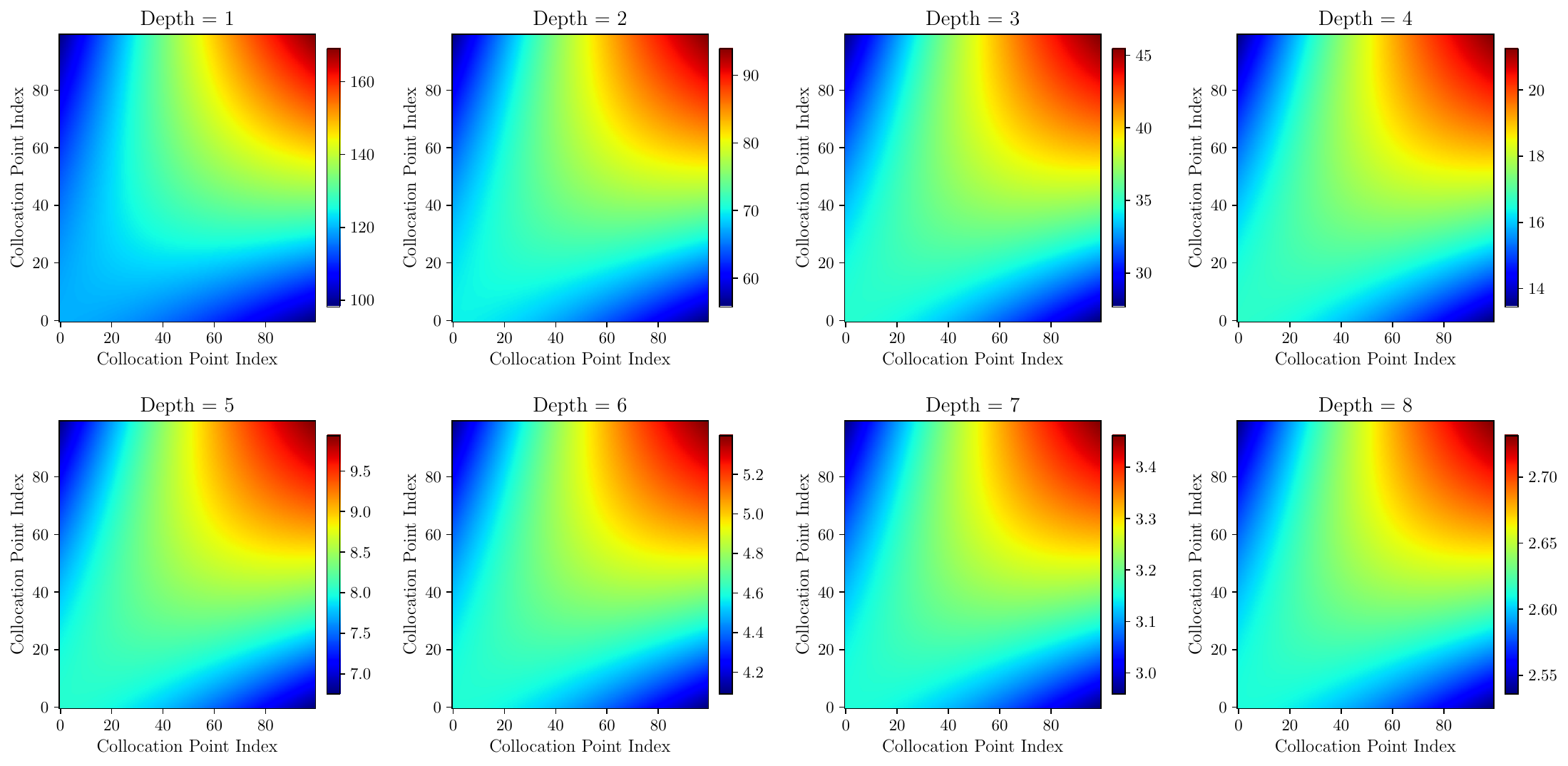}
    \caption{Heatmap comparison of mean NTK matrices across all tested depths. Each subplot shows the element-wise mean of the NTK matrix computed from \num{20} random initializations at the corresponding depth.}
    \label{fig:depth_heatmap_comparison}
\end{figure}

Fig.~\ref{fig:depth_heatmap_comparison} presents a comprehensive visual comparison of the mean NTK matrix structure across all tested depths.
Remarkably, despite the architectural differences introduced by varying the number of hidden layers from \num{1} to \num{8}, all NTK matrices maintain strikingly similar structural patterns.
The consistent geometric structure across depths indicates that the NTK's fundamental characteristics are preserved regardless of network depth.
However, unlike the structural preservation, the magnitude of matrix elements decreases dramatically with depth, as evidenced by the progressively smaller color scales from depth \num{1} to depth \num{8}.
This systematic magnitude reduction, consistent with the trace and Frobenius norm analysis, reveals that depth primarily dampens the overall kernel magnitude through repeated nonlinear transformations while preserving the relative spatial relationships between collocation points.
This observation contrasts sharply with the width analysis in Section~\ref{sec:invariance_width}, where magnitudes increased with network size, highlighting the fundamentally different roles of width and depth in shaping the NTK.

\begin{figure}[!htb]
    \centering
    \includegraphics[width=0.7\textwidth]{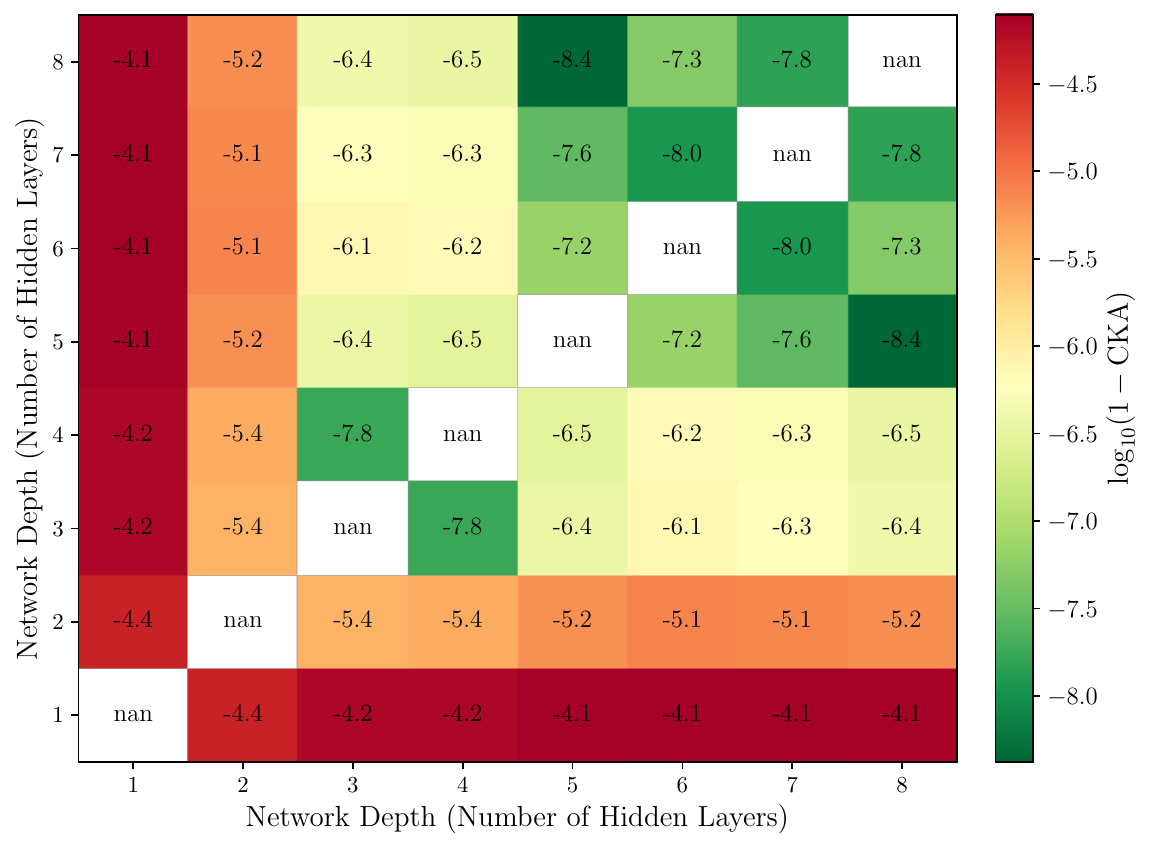}
    \caption{CKA similarity matrix between different network depths, displayed as $\log_{10}(1 - \text{CKA})$. Lower (more negative) values indicate higher similarity. Each cell compares the mean NTK matrices computed from \num{20} random seeds at the corresponding depths.}
    \label{fig:depth_cka_similarity}
\end{figure}

To quantitatively assess the structural similarity between NTK matrices at different depths, Fig.~\ref{fig:depth_cka_similarity} presents the CKA similarity matrix using a logarithmic scale $\log_{10}(1 - \text{CKA})$.
The analysis reveals several important findings:
\begin{enumerate}
    \item \textbf{Exceptional overall similarity.} All pairwise CKA values exceed \num{0.9999}, with most exceeding \num{0.999999}, indicating extremely high structural similarity across all depths—even higher than the width comparison in Section~\ref{sec:invariance_width}.
    \item \textbf{Smooth structural transitions.} Adjacent depth pairs show $\log_{10}(1 - \text{CKA})$ values ranging from \num{-4.42} (depths \num{1} vs \num{2}) to \num{-7.95} (depths \num{6} vs \num{7}), with progressively more negative values as depth increases, demonstrating increasingly similar structures between neighboring depths as networks become deeper.
    \item \textbf{Extreme depth preservation.} Even the comparison between the shallowest (depth \num{1}) and deepest (depth \num{8}) networks yields a CKA of \num{0.99992521}, corresponding to $\log_{10}(1 - \text{CKA}) = \num{-4.13}$, confirming that the fundamental NTK structure remains highly preserved across the entire depth spectrum.
    \item \textbf{Convergence pattern.} The progressively darker (more negative) values in the upper-right region of the matrix indicate that deeper networks exhibit increasingly similar structures to each other, suggesting convergence toward a characteristic deep network NTK geometry.
\end{enumerate}

In summary, the analysis of NTK invariance across network depths reveals distinct patterns that differ from width effects:
\begin{enumerate}
    \item \textbf{Complex stability dynamics.} Unlike width, which shows monotonic improvement in stability, depth exhibits non-monotonic behavior: CKA values gradually decrease with depth while CV shows a peak at intermediate depths (\num{4}--\num{6} layers) before declining in very deep networks.
          This suggests that moderate depths introduce the greatest initialization sensitivity, possibly due to competing effects of gradient propagation and expressiveness.
    \item \textbf{Structural preservation with feature concentration.} All depths maintain remarkably high structural similarity (CKA $> 0.9999$), even higher than across widths, while the monotonically decreasing effective rank reveals that deeper networks concentrate their NTK's spectral energy into fewer dominant modes, indicating a more hierarchical feature representation.
    \item \textbf{Contrasting effects of width and depth.} While increasing width enhances initialization stability and scales kernel magnitudes upward, increasing depth dampens kernel magnitudes dramatically (trace and Frobenius norm decrease by approximately \num{50}$\times$ from depth \num{1} to depth \num{8}) while introducing non-monotonic variability patterns.
          This distinction reflects the fundamentally different roles these architectural parameters play: width governs the kernel approximation quality to its infinite-width limit and amplifies kernel magnitudes, while depth shapes the hierarchical structure of learned features and attenuates kernel magnitudes through repeated nonlinear transformations.
\end{enumerate}
Combining the numerical analysis above with practical experience, networks with depths of \num{2}--\num{3} layers are found to offer a good balance between expressiveness and initialization stability, while deeper architectures require more careful consideration of initialization and gradient flow despite maintaining good structural similarity.

\subsection{Variation of Invariance with Activation Functions}
\label{sec:invariance_activation}

This subsection investigates the invariance of the initial NTK matrix $\bm{K_n}$ under different activation functions.

The experimental configuration follows Section~\ref{sec:invariance_seed}, with the primary modification being the activation function.
Six commonly used activation functions are tested: ReLU, Tanh, Sigmoid, LeakyReLU, ELU, and SELU.

Their mathematical definitions are as follows:
\begin{itemize}
    \item ReLU: $\text{ReLU}(x) = \max(0, x)$
    \item Tanh: $\tanh(x) = \frac{e^x - e^{-x}}{e^x + e^{-x}}$
    \item Sigmoid: $\text{Sigmoid}(x) = \frac{1}{1 + e^{-x}}$
    \item LeakyReLU: $\text{LeakyReLU}(x) = \begin{cases} x, & x \geq 0 \\ 0.01x, & x < 0 \end{cases}$
    \item ELU: $\text{ELU}(x) = \begin{cases} x, & x \geq 0 \\ \alpha(e^x - 1), & x < 0 \end{cases}$, with $\alpha = 1.0$
    \item SELU: $\text{SELU}(x) = \lambda \begin{cases} x, & x \geq 0 \\ \alpha(e^x - 1), & x < 0 \end{cases}$, with $\lambda \approx 1.0507$ and $\alpha \approx 1.6733$
\end{itemize}
The network architecture is fixed at depth \num{2} (hidden layers) and width \num{500} to ensure fair comparison.
For each activation function, \num{20} different random seed initializations are used to evaluate stability.
These activation functions represent two main categories: smooth functions (Tanh, Sigmoid, ELU, SELU) with continuous derivatives, and non-smooth functions (ReLU, LeakyReLU) with discontinuous derivatives at specific points.

\begin{figure}[!htb]
    \centering
    \includegraphics[width=\textwidth]{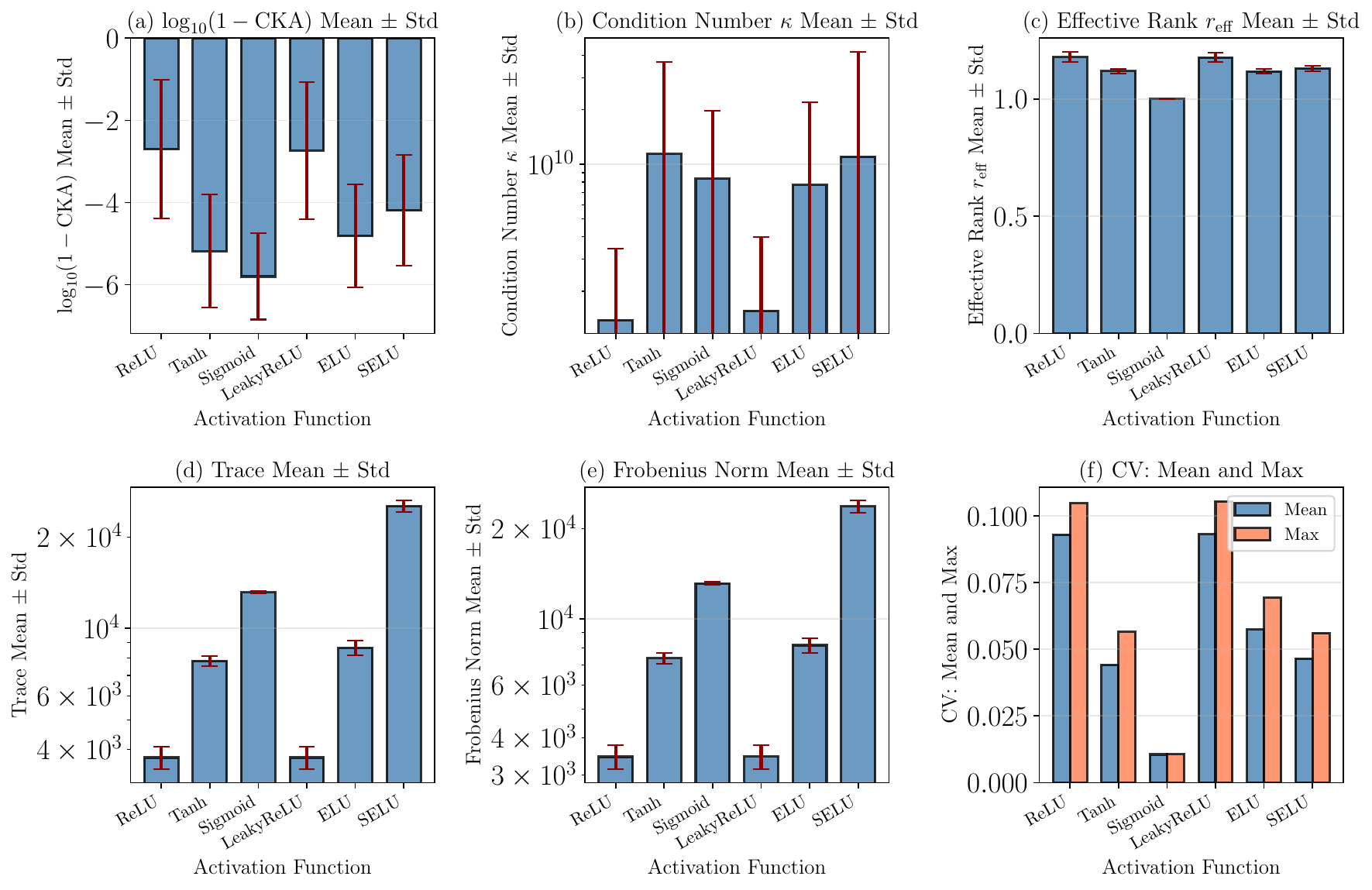}
    \caption{Stability metrics of NTK matrices across different activation functions. Each bar represents the mean value across \num{20} random seeds, with error bars indicating $\pm$ one standard deviation. For comparison, subplot (a) shows $\log_{10}(1 - \text{CKA})$ where more negative values indicate higher stability.}
    \label{fig:activation_stability_comparison}
\end{figure}

Fig.~\ref{fig:activation_stability_comparison} presents the stability metrics across different activation functions.
The $\log_{10}(1 - \text{CKA})$ metric (Fig.~\ref{fig:activation_stability_comparison}(a)) reveals a clear dichotomy between activation function types. Smooth activation functions (Sigmoid: \num{-5.80}, Tanh: \num{-5.19}, ELU: \num{-4.82}, SELU: \num{-4.19}) exhibit substantially more negative values compared to non-smooth functions (ReLU: \num{-2.70}, LeakyReLU: \num{-2.74}), indicating that smooth activations produce NTK matrices with dramatically better initialization invariance.
Specifically, Sigmoid achieves the highest stability with $\log_{10}(1 - \text{CKA}) = \num{-5.80}$, corresponding to CKA $= \num{0.999996}$, while ReLU and LeakyReLU show relatively lower stability with $\log_{10}(1 - \text{CKA}) \approx \num{-2.7}$, corresponding to CKA $\approx \num{0.995}$.
This two-to-three order of magnitude difference in $\log_{10}(1 - \text{CKA})$ demonstrates that activation function smoothness profoundly impacts NTK stability under random initialization.
The condition number (Fig.~\ref{fig:activation_stability_comparison}(b)) shows high variability across all activation functions, ranging from \num{1.39e9} (ReLU) to \num{1.14e10} (Tanh), with large error bars reflecting its sensitivity to eigenvalue perturbations.
The effective rank (Fig.~\ref{fig:activation_stability_comparison}(c)) reveals that Sigmoid produces the most concentrated eigenvalue distribution (\num{1.0004}), while ReLU-type activations yield slightly higher values (\num{1.18}), suggesting broader spectral distributions.
Both trace (Fig.~\ref{fig:activation_stability_comparison}(d)) and Frobenius norm (Fig.~\ref{fig:activation_stability_comparison}(e)) exhibit significant magnitude variations across activation functions, with Sigmoid producing the largest values (\num{3.33e4} for trace, \num{3.16e4} for Frobenius norm) and ReLU yielding smaller values (\num{2.00e3} for trace, \num{1.89e3} for Frobenius norm), indicating that activation function choice substantially affects kernel magnitude.
Most notably, the CV comparison (Fig.~\ref{fig:activation_stability_comparison}(f)) clearly distinguishes the two activation categories: non-smooth functions exhibit mean CV values around \num{0.093} with maxima around \num{0.105}, while smooth functions show dramatically lower values, with Sigmoid achieving mean CV of only \num{0.0106} and maximum of \num{0.0107}.
This represents a factor of 9$\times$ improvement in element-wise stability for the best smooth activation (Sigmoid) compared to non-smooth activations (ReLU/LeakyReLU), confirming that derivative continuity is crucial for initialization-invariant NTK behavior.

\begin{figure}[!htb]
    \centering
    \includegraphics[width=\textwidth]{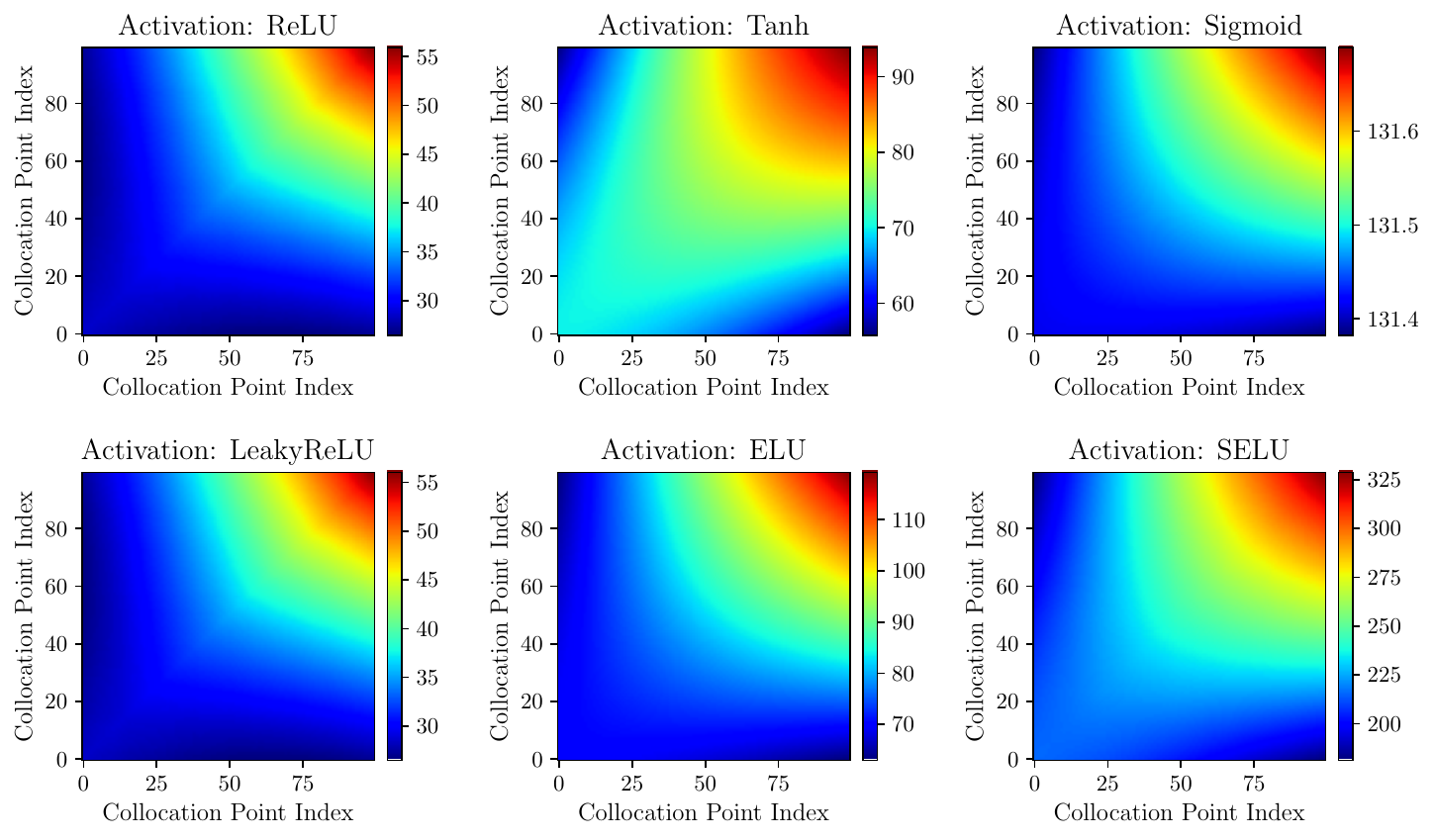}
    \caption{Heatmap comparison of mean NTK matrices across all tested activation functions. Each subplot shows the element-wise mean of the NTK matrix computed from \num{20} random initializations using the corresponding activation function.}
    \label{fig:activation_heatmap_comparison}
\end{figure}

Fig.~\ref{fig:activation_heatmap_comparison} presents a comprehensive visual comparison of the mean NTK matrix structure across all tested activation functions.
Careful examination of the heatmaps reveals notable structural differences between activation function families, beyond simple magnitude scaling.
ReLU and LeakyReLU exhibit nearly identical structures, as expected from their closely related mathematical forms (LeakyReLU is ReLU with a small negative slope of \num{0.01}).

Both show a strong diagonal dominance with sharp transitions near the diagonal, reflecting the piecewise linear nature of these activations.

Sigmoid and ELU also show structural similarity.

Tanh and SELU exhibit distinct NTK distributions compared to the other activations.
The magnitude differences are also substantial: Sigmoid generates the largest kernel magnitudes (color scale reaching $\sim$\num{300}), followed by Tanh ($\sim$\num{80}), ELU ($\sim$\num{65}), SELU ($\sim$\num{40}), and finally ReLU/LeakyReLU with the smallest magnitudes ($\sim$\num{20}).
This 15$\times$ magnitude range from ReLU to Sigmoid, combined with the observable structural variations, demonstrates that activation function choice significantly impacts both the scale and geometric structure of the NTK, rather than merely applying a uniform scaling factor.

\begin{figure}[!htb]
    \centering
    \includegraphics[width=0.8\textwidth]{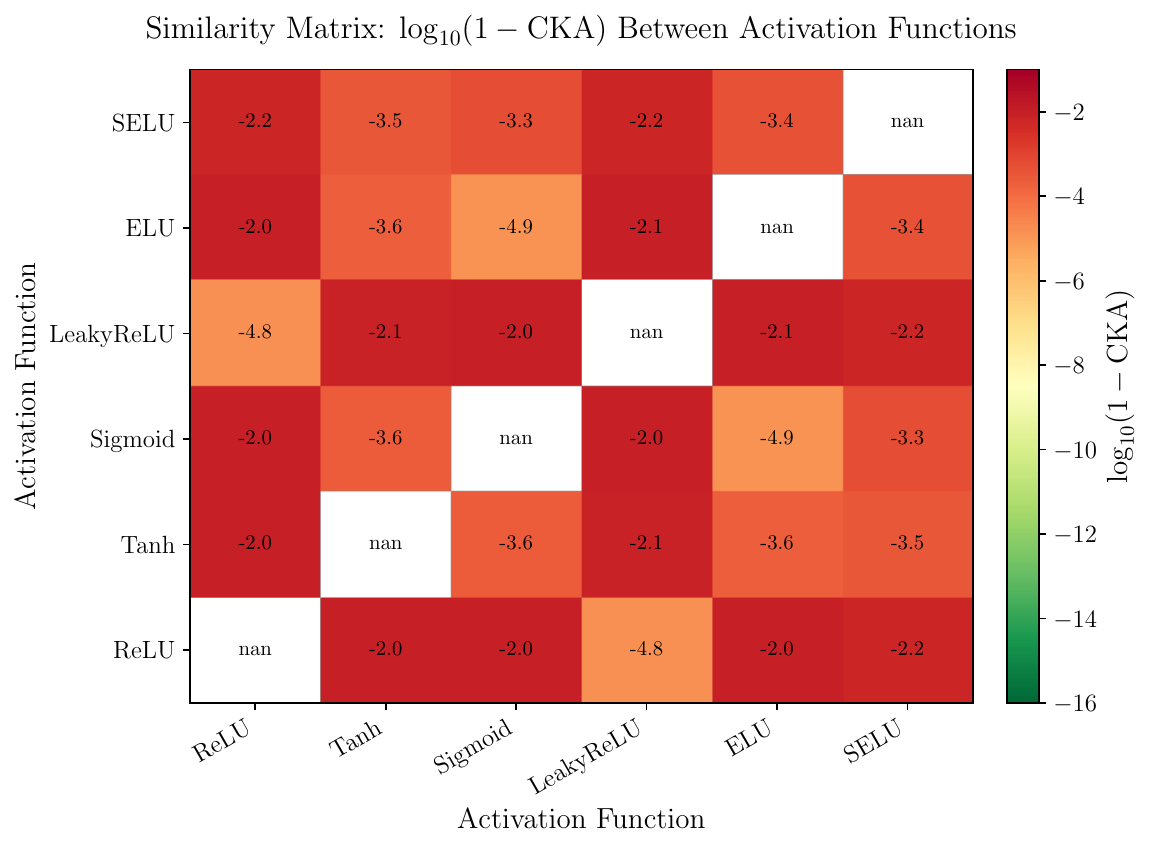}
    \caption{CKA similarity matrix between different activation functions, displayed as $\log_{10}(1 - \text{CKA})$. Lower (more negative) values indicate higher similarity. Each cell compares the mean NTK matrices computed from \num{20} random seeds using the corresponding activation functions.}
    \label{fig:activation_cka_similarity}
\end{figure}

To quantitatively assess the structural similarity between NTK matrices produced by different activation functions, Fig.~\ref{fig:activation_cka_similarity} presents the CKA similarity matrix using a logarithmic scale $\log_{10}(1 - \text{CKA})$.
The analysis reveals several important findings organized by activation function smoothness difference
\begin{enumerate}
    \item \textbf{Within smooth activation group.} The four smooth activation functions (Tanh, Sigmoid, ELU, SELU) exhibit extremely high mutual similarity, with pairwise CKA values exceeding \num{0.9993} and $\log_{10}(1 - \text{CKA})$ ranging from \num{-3.27} to \num{-4.89}.
          The highest similarity is observed between Sigmoid and ELU ($\log_{10}(1 - \text{CKA}) = \num{-4.89}$, CKA $= \num{0.99998707}$), while the lowest within this group is between Sigmoid and SELU ($\log_{10}(1 - \text{CKA}) = \num{-3.27}$, CKA $= \num{0.99946078}$).
          Despite variations in their specific functional forms and derivatives, these smooth activations converge to highly similar NTK structures, suggesting that derivative continuity is more important than the exact nonlinearity shape.
    \item \textbf{Within non-smooth activation group.} ReLU and LeakyReLU demonstrate near-perfect similarity with $\log_{10}(1 - \text{CKA}) = \num{-4.81}$ (CKA $= \num{0.99998462}$), which is expected given that LeakyReLU is a generalization of ReLU with a small negative slope.
          This exceptional similarity validates the consistency of NTK behavior within the ReLU family of activations.
    \item \textbf{Cross-group comparisons.} When comparing smooth versus non-smooth activations, CKA values drop to approximately \num{0.99}--\num{0.991}, with $\log_{10}(1 - \text{CKA})$ ranging from \num{-2.01} to \num{-2.06}.
          For instance, Tanh versus ReLU yields $\log_{10}(1 - \text{CKA}) = \num{-2.04}$ (CKA $= \num{0.99089945}$), and Sigmoid versus LeakyReLU yields $\log_{10}(1 - \text{CKA}) = \num{-2.02}$ (CKA $= \num{0.99051734}$).
          While these CKA values remain high in absolute terms, they represent approximately 1\% structural differences—substantially larger than the within-group variations ($< 0.1\%$).
          This clear separation demonstrates that the smoothness property fundamentally shapes the NTK structure, creating distinct geometric signatures for smooth versus non-smooth activation families.
\end{enumerate}

In summary, the analysis of NTK invariance across activation functions reveals fundamental relationships between activation properties and kernel stability:
\begin{enumerate}
    \item \textbf{Smoothness determines stability.} Activation functions with continuous derivatives (Tanh, Sigmoid, ELU, SELU) produce NTK matrices that are highly insensitive to random initialization, with CKA values exceeding \num{0.9998} and CV values below \num{0.06}.
          In contrast, non-smooth activations (ReLU, Leaky-\\ReLU) exhibit greater initialization sensitivity, with CKA values around \num{0.995} and CV values around \num{0.093}, representing a 9$\times$ reduction in element-wise stability compared to the most stable smooth activation (Sigmoid).
    \item \textbf{Within-group similarity and cross-group distinction.} Activation functions within the same category (smooth or non-smooth) produce highly similar NTK structures (CKA $> 0.9993$), while cross-category comparisons show measurably lower similarity (CKA $\approx 0.99$).
          This suggests that the continuity of derivatives creates a stronger structural imprint on the NTK than the specific functional form of the activation.
    \item \textbf{Practical implications.} For applications requiring stable NTK behavior—such as theoretical analysis based on NTK theory or kernel-based training algorithms—smooth activation functions are strongly preferred, with Sigmoid offering the highest stability.
          When using non-smooth activations like ReLU (common in practice due to their computational efficiency and mitigation of vanishing gradients), ensemble methods or multiple random initializations may be necessary to achieve reliable results.
          The observed 1\% structural difference between smooth and non-smooth activation groups is sufficient to impact certain NTK-dependent applications, though the overall high similarity (CKA $> 0.99$) confirms the NTK's robustness across diverse activation choices.

\end{enumerate}
This finding reinforces the importance of smoothness assumptions in theoretical analysis of neural networks and provides empirical validation for the role of activation function regularity in neural network behavior.

\subsection{Summary of the invariance of initial \texorpdfstring{$\bm{K_n}$}{Kn}}

This section systematically investigated the invariance of the initial NTK matrix $\bm{K_n}$ under four fundamental factors: random initialization seeds, network width, network depth, and activation functions.
\begin{enumerate}
    \item \textbf{Random initialization.} Wide networks exhibit strong NTK invariance across different random seeds, validating the theoretical predictions of NTK theory in the wide network regime.
    \item \textbf{Network width.} NTK invariance improves monotonically with increasing width. Wider networks demonstrate excellent stability, while narrow networks show greater sensitivity to initialization. Structural similarity is preserved across all tested widths, though kernel magnitudes scale upward with width.
    \item \textbf{Network depth.} Unlike width, depth exhibits non-monotonic stability patterns. All depths maintain high structural similarity, with intermediate depths showing the greatest variability. Notably, kernel magnitudes decrease dramatically with depth, contrasting sharply with width effects.
    \item \textbf{Activation functions.} Smooth activations significantly outperform non-smooth ones in initialization stability. Within each category, activation functions produce highly similar NTK structures, while cross-category comparisons reveal measurable structural differences.
\end{enumerate}
These findings establish that the NTK matrix exhibits remarkable structural robustness across diverse architectural choices and initializations. For practical applications requiring reliable NTK behavior, wider networks with moderate depth and smooth activation functions are recommended.

\begin{flushleft}
\bibliographystyle{unsrt}
\bibliography{references}
\end{flushleft}

\end{document}